\documentclass{IEEEtaes}

\usepackage{color,array,amsthm}
\usepackage{graphicx}
\let\labelindent\relax
\usepackage{enumitem}
\usepackage{amssymb}
\usepackage{amsmath}
\usepackage{tikz}

\newcommand{\Rmnum}[1]{\uppercase\expandafter{\romannumeral #1}}  
\usepackage{gensymb}
\usepackage[ruled,linesnumbered]{algorithm2e}
\usepackage{subfloat}
\usepackage{subcaption}
\jvol{XX}
\jnum{XX}
\jmonth{XXXXX}
\paper{1234567}
\pubyear{2022}
\doiinfo{TAES.2022.Doi Number}

\begin{document}

\title{Three-dimensional Trajectory Optimization for Quadrotor Tail-sitter UAVs: Traversing through Given Waypoints}

\author{MINGYUE FAN}
\affil{Zhejiang University, Hangzhou, Zhejiang 310058, P.R. China} 

\author{FANGFANG XIE}
\affil{Zhejiang University, Hangzhou, Zhejiang 310058, P.R. China} 

\author{TINGWEI JI}
\affil{Zhejiang University, Hangzhou, Zhejiang 310058, P.R. China}

\author{YAO ZHENG}
\affil{Zhejiang University, Hangzhou, Zhejiang 310058, P.R. China}

\receiveddate{Manuscript received XXXXX 00, 0000; revised XXXXX 00, 0000; accepted XXXXX 00, 0000.\\
The authors declare that no funds, grants, or other support were received during the preparation of this manuscript.}

\corresp{{\itshape (Corresponding author: Tingwei Ji.)} }

\authoraddress{Authors’ addresses: Mingyue Fan, Fangfang Xie, Tingwei Ji, and Yao Zheng are with the School of Aeronautics and Astronautics, Zhejiang University, Hangzhou, 310058 China. (E-mail: \href{mingyue\_fan@zju.edu.cn}{mingyue\_fan@zju.edu.cn}; \href{fangfang\_xie@zju.edu.cn}{fangfang\_xie@zju.edu.cn}; \href{zjjtw@zju.edu.cn}{zjjtw@zju.edu.cn}; \href{yao.zheng@zju.edu.cn}{yao.zheng@zju.edu.cn})}


\markboth{FAN ET AL.}{TRAJECTORY OPTIMIZATION FOR TAIL-SITTER UAVs: TRAVERSING THROUGH GIVEN WAYPOINTS}
\maketitle

\begin{abstract}
    Given the evolving application scenarios of current fixed-wing unmanned aerial vehicles (UAVs), it is necessary for UAVs to possess agile and rapid 3-dimensional flight capabilities. Typically, the trajectory of a tail-sitter is generated separately for vertical and level flights. This limits the tail-sitter's ability to move in a 3-dimensional airspace and makes it difficult to establish a smooth transition between vertical and level flights. In the present work, a 3-dimensional trajectory optimization method is proposed for quadrotor tail-sitters. Especially, the differential dynamics constraints are eliminated when generating the trajectory of the tail-sitter by utilizing differential flatness method. Additionally, the temporal parameters of the trajectory are generated using the state-of-the-art trajectory generation method called MINCO (minimum control). 
    Subsequently, we convert the speed constraint on the vehicle into a soft constraint by discretizing the trajectory in time. This increases the likelihood that the control input limits are satisfied and the trajectory is feasible.
    Then, we utilize a kind of model predictive control (MPC) method to track trajectories. Even if restricting the tail-sitter's motion to a 2-dimensional horizontal plane, the solutions still outperform those of the L1 Guidance Law and Dubins path.

\end{abstract}

\begin{IEEEkeywords}
Autonomous aerial vehicles, trajectory optimization, vehicle dynamics.
\end{IEEEkeywords}

\section{INTRODUCTION}
\label{sec:introduction}
Vertical Take-off and Landing (VTOL) Unmanned Aerial Vehicles (UAVs) differ from fixed-wing UAVs in that they do not require a runway for takeoff and landing, and they possess the ability to hover in mid-air. This versatility has led to their wide applications across various fields\cite{huang2024pesticide,duan2023differential,prajapati2023aerial,vasconcelos2024uav}. The tail-sitter is a hybrid UAV that combines VTOL's agility with fixed-wing's endurance, allowing longer flights and heavier payloads. 
The quadrotor tail-sitter, with its rotors rigidly attached to the wing, can transition by tilting, offering a simple, efficient design ideal for small, portable UAVs. Traditional UAV navigation focused on Global Positioning System (GPS) and Inertial Measurement Unit (IMU) for high-altitude flights. In this case, generating smooth flight trajectories between two steady flight conditions is particularly critical, especially when they are at varying altitudes \cite{hong2021hierarchical}. In addition, with a growing need for UAVs to fly in regions with limited space, agile 3D maneuverability is essential. Therefore, tail-sitters face challenges in generating smooth, efficient 3-dimensional trajectories due to fast-varying speed and attitude.

Designing a 3-dimensional trajectory that spans a large flight envelope of a tail-sitter requires significant processing resources and time. This is due to the extreme nonlinear aerodynamic forces caused by a wide range of angle of attack (AoA). As a compromise, tail-sitters are typically set at a specific AoA. This allows them to be modeled as a rotary-wing model during slow vertical flight and as a fixed-wing aircraft during level flight \cite{cheng2022transition}. During slow vertical flight, tail-sitters can be modeled as rotary-wing UAVs, leveraging well-researched trajectory planning methods \cite{xing2024probabilistic,toumieh2024high,song2023reaching,kaufmann2023champion}. In level flight, where the kinematics simplify that of a unicycle, guidance algorithms such as the Dubins path or analytical methods can be applied \cite{moon2023time,rao2024curvature,kumar2024robust}.
However, the transition phase between these two flight modes remains a formidable challenge \cite{wang2023adaptive}. Conventionally, the incorporation of aerodynamics during the transition phase involves formulating the trajectory generation as a complex nonlinear optimization problem. This entails addressing various objectives and constraints such as time minimization \cite{mcintosh2024aerodynamic,gupta2023optimal}, energy efficiency \cite{panish2024tiltwing,kaneko2024simultaneous}, safety \cite{cong2024longitudinal}, and altitude-hold \cite{cheng2023corridor}.
Nevertheless, these methods typically restrict the transition phase in one- or two-dimensional Euclidean spaces, hindering the achievement of rapid and agile flight. Furthermore, despite the existence of structurally simple geometric methods for generating flight trajectories in level flight, their effectiveness is occasionally compromised by the failure to fulfill the dynamics of the aircraft, resulting in a mismatch between the planned trajectory and the actual maneuver \cite{hong2021computationally}.

Recently, to overcome these limitations and advance the field, researchers from Massachusetts Institute of Technology (MIT) and The University of Hong Kong (HKU) has proposed innovative trajectory planning methods for tail-sitter UAVs that leverage the property of differential flatness \cite{tal2021global,lu2024trajectory}. Differential flatness, originally introduced by Fliess \cite{fliess1995flatness}, is a crucial characteristic of under-actuated systems. It enables the direct computation of reference states and inputs from the transient information of the flat output trajectory, avoiding the need to integrate the differential equations of the dynamic system. Tal \cite{tal2021global} demonstrated the differential flatness transformation of a tail-sitter UAV based on the $\phi$-theory aerodynamic model \cite{lustosa2019global}. In this demonstration, the vehicle's position and Euler yaw angle were selected as flat-outputs with disregarding the effect of wind conditions. By optimizing the minimum snap trajectory in the flat-output space, they achieved agile and maneuverable flight of the tail-sitter \cite{tal2023aerobatic}. In contrast, Lu's study \cite{lu2024trajectory} adopts a realistic and comprehensive 3D model of the tail-sitter, which offers a more accurate representation of the tail-sitter's dynamics, potentially overcoming the limitations of the $\phi$-theory model and improving the precision and robustness of trajectory planning for tail-sitter UAVs. The use of a full 3D model also provides valuable insights into the aerodynamic characteristics of tail-sitters, facilitating the development of advanced guidance, navigation, and control algorithms for these versatile aircraft. 

In the meanwhile, global trajectory optimization is especially important for tail-sitters. It enhances operational efficiency, safety, performance, flexibility, and cost-effectiveness, enabling more precise and effective mission completion \cite{hong2021free}. Generating high-quality, 3-dimensional trajectories for tail-sitters involves several key algorithmic obstacles. 
First, real-time trajectory calculations must be performed within the limitations of size, weight, and power, requiring efficient utilization of onboard resources. Second, high-quality motions typically require fine discretization of the dynamic process, taking into account the aerodynamics of the tail-sitter. Third, trajectories must not only be smooth geometrically but also dynamically feasible, meeting both acceptable states and inputs. Finally, the nonlinear aerodynamics produced by the fuselage and wings makes it challenging to design a trajectory that avoids undesirable AoA during flight. However, there is rare relevant research work in this aspect. Most extensive works have been conducted for multicopters utilizing differential flatness transformation. For example, Mellinger \cite{mellinger2011minimum} employed fixed-duration splines to characterize the flat trajectories of quadrotors, integrating the square norm of the fourth-order time derivative as a cost function for quadratic programming to ensure the smoothness of the trajectory.
Bry \cite{bry2015aggressive} eliminated equality constraints by leveraging boundary derivative transformations, thereby solving an unconstrained quadratic programming (QP) problem and obtaining a closed-form solution. Nevertheless, the efficiency of solving sparse linear equation systems becomes questionable when the problem scale escalates. Furthermore, the vehicle's speed and other dynamic constraints cannot be explicitly incorporated into the trajectory generation process.
Burke \cite{burke2020generating} achieves a remarkable feat by developing an algorithm that solves the primal-dual variable of quadratic programming problems with linear complexity, showcasing significant efficiency gains particularly when dealing with numerous trajectory stages. This advancement underscores the algorithm's potential to streamline complex trajectory optimization processes.
Meanwhile, Wang \cite{wang2021generating} presents a analytical inverse of the boundary derivative transformation, complete with an analytical gradient of parameters. This analytical approach provides a precise and efficient means of handling scenarios with a single start and end point.

In summary, the current state-of-the-art tail-sitter trajectory planning methods primarily focus on one- or two-dimensional Euclidean spaces, often compromising on critical aspects like time optimization, constraint fidelity, or resorting to heuristic approximations in pursuit of online planning's computational efficiency. Consequently, the quality of these trajectory solutions falls short of practical expectations, which requires further trajectory optimization to ensure a smooth and efficient transition, minimizing energy consumption and maximizing stability. This paper addresses these limitations by making the following contributions:
\begin{enumerate}[leftmargin=1cm,itemsep=0pt,topsep=-5pt,labelsep=0.5em]
\item We propose an optimization-driven, multistage
trajectory generation method tailored for
quadrotor tail-sitters. This method allows the tail-sitter
to start from a hovering state and execute an optimal
trajectory that passes through any given 3-dimensional
intermediate waypoints. This method meticulously considering flight duration and maximum speed limits.
\item We map the rotation matrix to Euclidean space and then use model predictive control (MPC) to track the position, velocity, and attitude of the tail-sitter. This enables it to achieve full envelope flight in 3-dimensional space. Meanwhile, during trajectory tracking, actuator constraints are satisfied.
\item Comparative experiments show that our method significantly outperforms well-established fixed-wing algorithms in terms of efficiency, optimality, and robustness. 
\end{enumerate}
\vspace{5pt}

The structure of this paper is organized as follows. The flight dynamics of the tail-sitter is presented in Section \ref{sec:flight_dynamics}. Section \ref{sec:multi_op} provides a thorough explanation of the trajectory optimization problem and our method. In Section \ref{sec:MPC}, an MPC-based trajectory tracking controller is proposed and its performance is verified on two trajectories. In Section \ref{sec:comparison} comparative experiments are given and discussed. Conclusions are draw in Section \ref{sec:conclusion}.

\section{FLIGHT DYNAMICS}
\label{sec:flight_dynamics}
This section presents the flight dynamics model of the tail-sitter, which serves as the foundation for our trajectory generation method. In Section \hyperref[sec:flight_dynamics1]{\Rmnum{2}-A}, the translational and rotational dynamics of the tail-sitter is introduced. Then in Section \hyperref[sec:flight_dynamics2]{\Rmnum{2}-B}, the analytical model is shown. Most importantly, the corresponding aerodynamics parameters are derived in Section \hyperref[sec:flight_dynamics3]{\Rmnum{2}-C} by conducting the computational fluid dynamics.

\subsection{Vehicle Equations of Motion}
\label{sec:flight_dynamics1}
In this subsection, the tail-sitter that we utilize is firstly presented. Then the corresponding inertial coordinate and body coordinate frames are defined. Finally, the dynamics of the tail-sitter are derived.
\begin{figure}[h]
    \centering
    \includegraphics[width=0.8\linewidth]{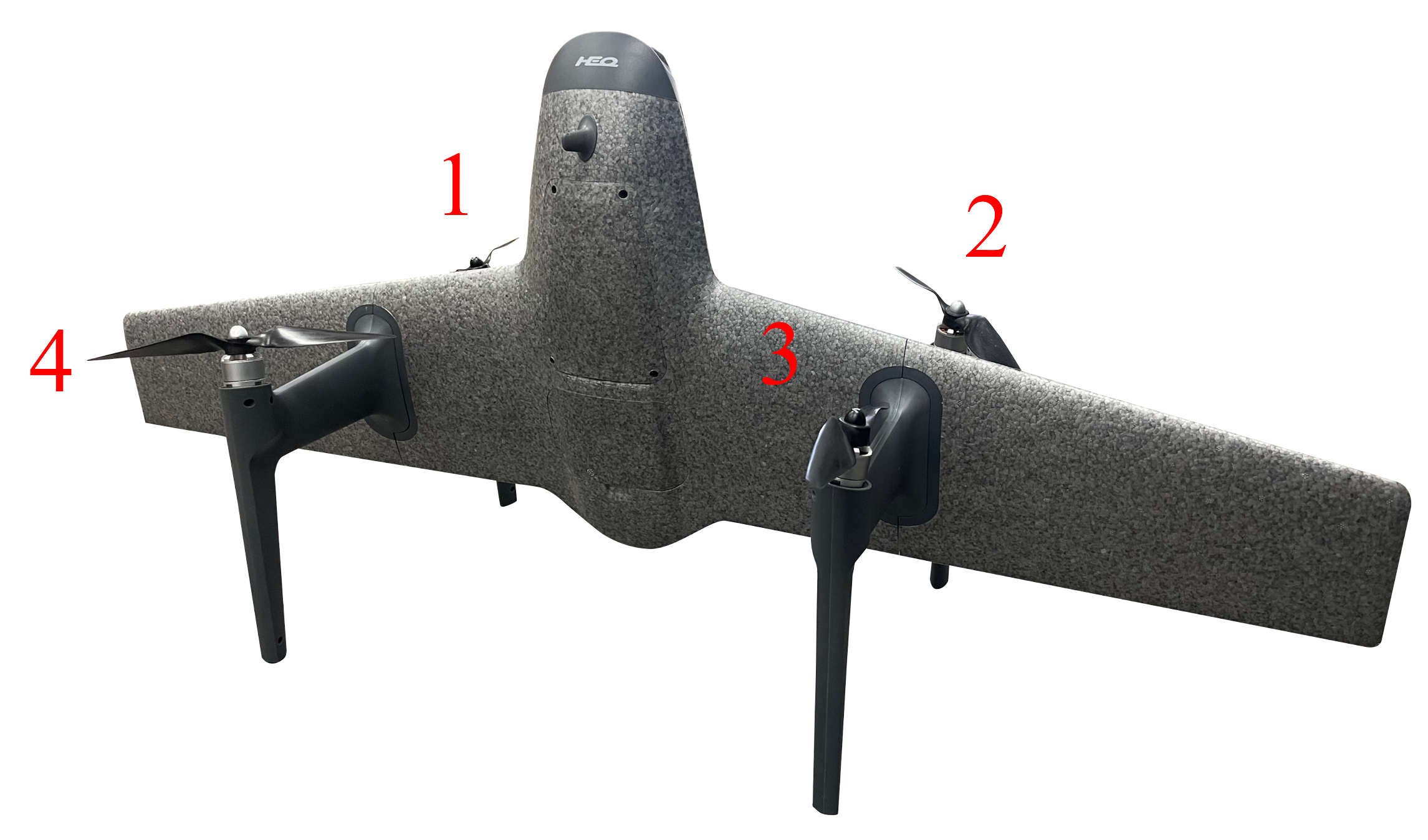}
    \caption{SWAN K1 PRO}
    \label{fig:tailsitter}
\end{figure}

The SWAN K1 PRO, developed by HEQ UAV Technical Company in Shenzhen, China, is the vehicle we utilize. The vehicle's mass is 1.3328 $kg$ and its wingspan measures 1.085 $m$, as shown in Fig.\ref{fig:tailsitter}. It is evident that the vehicle lacks control surfaces, relying solely on the four propellers on the fuselage to generate all torques and forces. Propellers 1 and 3 consistently rotate in a clockwise direction, whereas propellers 2 and 4 consistently rotate in a counterclockwise direction, thereby maintaining balance by counteracting torques.
\begin{figure}[h]
    \centering
    \includegraphics[width=1\linewidth]{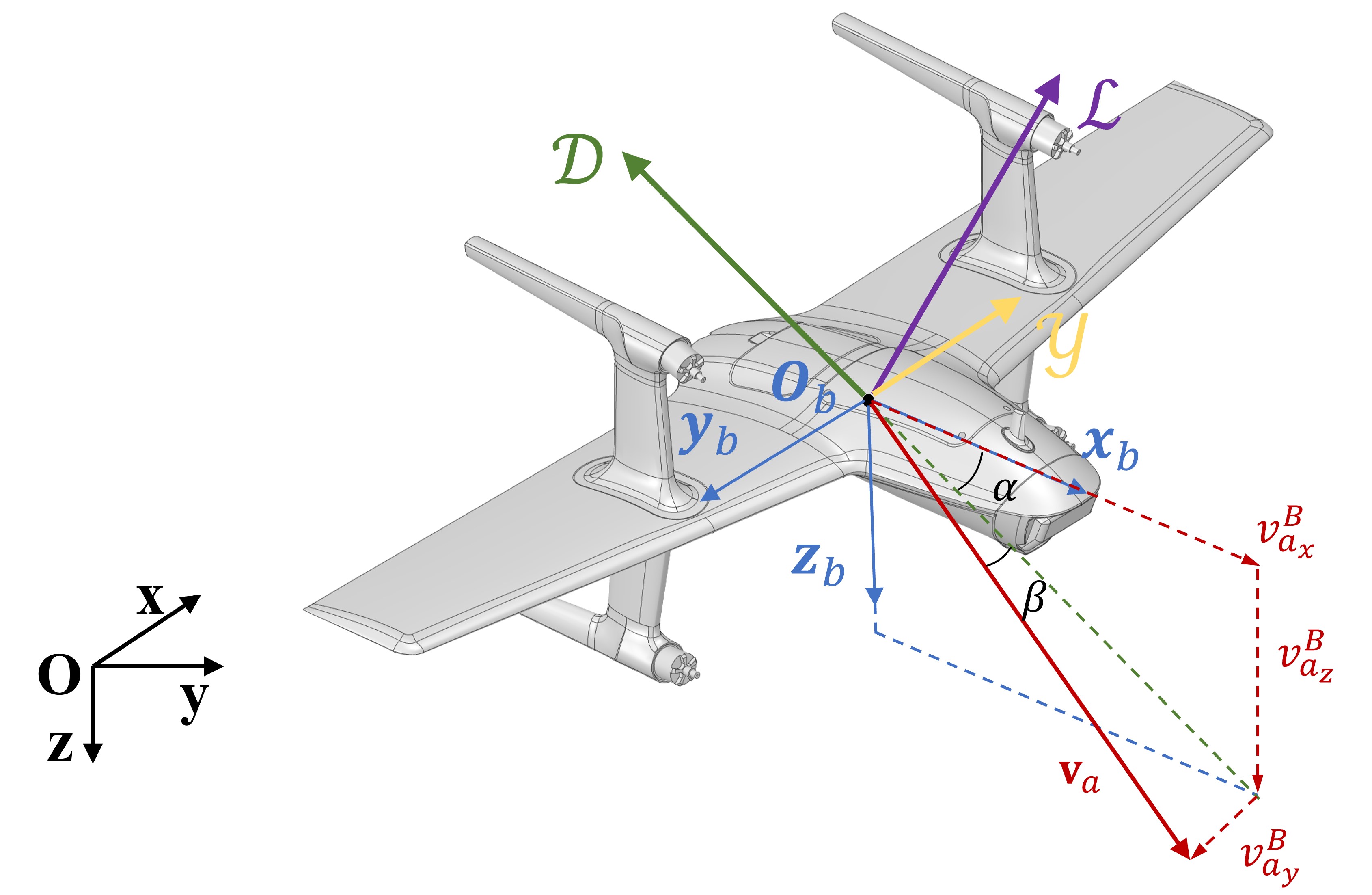}
    \caption{Coordinate frames and aerodynamic forces}
    \label{fig:coordinateframe}
\end{figure}

As shown in Fig.\ref{fig:coordinateframe}, the inertial frame $\left \{ \mathbf{O}\mathbf{x}\mathbf{y}\mathbf{z} \right \}$ is defined as North-East-Down (NED), and correspondingly, the body frame  $\left \{ \mathbf{O}_b\mathbf{x}_b\mathbf{y}_b\mathbf{z}_b \right \}$ is defined as Front-Right-Down (FRD). $\mathbf{O}$ is the origin of the world coordinate frame, while $\mathbf{O}_b$ is the center of gravity of the vehicle.

The states of the tail-sitter are denoted by $\mathbf{x}_{full}=\left \{ \mathbf{p},\mathbf{v},\mathbf{R},\boldsymbol{\omega} \right \}$, where $\mathbf{p}\in\mathbb{R}^3$ represents the vehicle position in the world-fixed reference frame. $\mathbf{v}\in\mathbb{R}^3$ is the vehicle velocity in the world-fixed reference frame. $\mathbf{R}\in SO(3)$ denotes the rotation from the body frame to the inertial frame, $\boldsymbol{\omega}\in\mathbb{R}^3$ is the angular velocity in the body frame. The inputs for the tail-sitter in terms of force and torques are $\mathbf{u}_{full}=\left \{ f,\boldsymbol{\tau} \right \}$, where $f$ and $\boldsymbol{\tau}\in \mathbb{R}^3$ denote the thrust and control moment vector produced by 4 propellers, respectively. Hence, the quadrotor tail-sitter translational and rotational dynamics are given by:
\begin{subequations}
    \label{equ:tr_r_tail-sitter}
    \begin{align}
        \dot{\mathbf{p}} =& \mathbf{v}\\
         \label{co_R}
        \dot{\mathbf{v}} =& \mathbf{g}+\frac{1}{m}(f\mathbf{R}\mathbf{e}_1+\mathbf{R}\mathbf{f}_a)\\
        \dot{\mathbf{R}} =& \mathbf{R}\left \lfloor \boldsymbol{\omega}  \right \rfloor\\
        \label{eq:app_omega}
        \mathbf{J}\dot{\boldsymbol{\omega}} =& \boldsymbol{\tau}+\mathbf{M}_a -\boldsymbol{\omega}\times\mathbf{J}\boldsymbol{\omega}
    \end{align}
\end{subequations}
where $\mathbf{g}=(0,0,9.8)^T$ is the gravitational acceleration, and $m$ is the vehicle mass. $\mathbf{e}_1=(1,0,0)^T$, $\mathbf{e}_2=(0,1,0)^T$, $\mathbf{e}_3=(0,0,1)^T$, are unit vectors. $\mathbf{f}_a\in\mathbb{R}^3$ and $\mathbf{M}_a\in\mathbb{R}^3$ are the aerodynamic force and moment in the body frame, which will be introduced in Section \hyperref[sec:flight_dynamics2]{\Rmnum{2}-B}. $\mathbf{J}\in \mathbb{R}^{3\times3}$ is the inertia tensor matrix of the vehicle about the body frame with the center of gravity as the origin. The notation $\left \lfloor \cdot \right \rfloor $ converts a 3-D vector into a antisymmetric matrix.

It should be noted that the direction of thrust is assumed to align with the $\mathbf{x}_b$. In situation where propellers are installed at a fixed angle, it only requires a simple transformation using a constant matrix.

\subsection{Anaylitical Aerodynamic model}\label{sec:flight_dynamics2}
The aerodynamic forces and moments of an aircraft can be described as a set of functions that depend on the AoA $\alpha$ and the sideslip angle $\beta$.
In the present work, the aerodynamic force $\mathbf{f}_a$ is modeled in the body frame as follows:
\begin{align}
    \label{eq:faLDY}
    & \mathbf{f}_a=\begin{bmatrix}
                        -\cos \alpha & 0 & \sin \alpha \\
                        0 & 1 & 0 \\
                        -\sin \alpha & 0 & -\cos \alpha
                        \end{bmatrix}\begin{bmatrix}
                                                    \mathcal{D} \\
                                                    \mathcal{Y} \\
                                                    \mathcal{L}
                                                \end{bmatrix}
\end{align}
where $\mathcal{L}$, $\mathcal{D}$, $\mathcal{Y}$ are respectively the lift, drag, and side force produced by fuselage and wings. $L$, $M$, and $N$ are, respectively, the rolling, pitching, and yawing moment along the body axis $\mathbf{x}_b$, $\mathbf{y}_b$, $\mathbf{z}_b$. Hence, the aerodynamic moment vector $\mathbf{M}_a$ can be modeled as: 
\begin{align}
    \label{eq:Ma}
    & \mathbf{M}_a=(L, M, N)^T
\end{align}

Referring to Etkin and Reid \cite{etkin1995dynamics}, the aerodynamic forces and moments are further parameterized as:
\begin{equation}
    \label{eq:etkin}
    \begin{split}
        \mathcal{L}=\frac{1}{2} \rho V^{2} S C_{L},\; \mathcal{D}=&\frac{1}{2} \rho V^{2} S C_{D},\; \mathcal{Y}=\frac{1}{2} \rho V^{2} S C_{Y} \\
        L=\frac{1}{2} \rho V^{2} S b C_{l},\; M=&\frac{1}{2} \rho V^{2} S \bar{c} C_{m},\; N=\frac{1}{2} \rho V^{2} S b C_{n}
    \end{split}
\end{equation}
where $V=\left\|\mathbf{v}_a\right\|$ is the magnitude of the air speed, $\rho$ is the density of the air, $S$ is the reference area of the wing, $\bar{c}$ is the mean aerodynamic chord, $b$ is the span of airplane. Note that $\mathbf{v}_a = \mathbf{v}-\mathbf{w}$, where $\mathbf{w}$ is wind velocity. $C_L$, $C_D$, and $C_Y$ are the lift coefficients, drag coefficients, and side force coefficients of the vehicle, respectively, while $C_l$, $C_m$, and $C_n$ are the rolling, pitching, and yawing moment coefficients, respectively. Each set of non-dimensional numbers is exclusively linked to the aircraft's shape and attitude.

For readability, the total aerodynamic force $\mathbf{f}_a$ in (\ref{eq:faLDY}) can be rewritten as:
\begin{equation}
    \label{eq:fa}
    \mathbf{f}_{a}=\frac{1}{2} \rho V^{2} S \mathbf{C}(\alpha, \beta)
\end{equation}
where
\begin{subequations}
    \begin{align}
        \mathbf{C}(\alpha, \beta) =&\left[\mathbf{C}_{x}(\alpha, \beta) \quad  \mathbf{C}_{y}(\alpha, \beta)\quad \mathbf{C}_{z}(\alpha, \beta)\right]^T \\
        \mathbf{C}_{x}(\alpha, \beta) =&-C_{D}(\alpha, \beta) \cos \alpha+C_{L}(\alpha, \beta) \sin \alpha \\
        \mathbf{C}_{y}(\alpha, \beta) =& C_{Y}(\alpha, \beta) \\
        \label{eq:app_fa}
        \mathbf{C}_{z}(\alpha, \beta) =&-C_{D}(\alpha, \beta) \sin \alpha-C_{L}(\alpha, \beta) \cos \alpha
    \end{align}
\end{subequations}

The AoA $\alpha$, and the sideslip angle $\beta$ are calculated as follows:
\begin{align}
    \mathbf{v}_a^B=&\mathbf{R}^T\mathbf{v}_{a}=\left[\mathbf{v}_{a_{x}}^B \quad \mathbf{v}_{a_{y}}^B \quad \mathbf{v}_{a_{z}}^B\right]^{T}\\
    \alpha=&\arctan\left(\frac{\mathbf{v}_{a_{z}}^{B}}{\mathbf{v}_{a_{x}}^{B}}\right)\\ \beta=&\arcsin\left(\frac{\mathbf{v}_{a_{y}}^{B}}{V}\right)
\end{align}

Due to the airframe is symmetric to the $\mathbf{x}_b\mathbf{O}_b\mathbf{z}_b$ plane, which indicates:
\begin{subequations}
    \begin{align}
        &C_{L}(\alpha, \beta)=C_{L}(\alpha,-\beta), \forall \alpha, \beta \\
        &C_{D}(\alpha, \beta)=C_{D}(\alpha,-\beta), \forall \alpha, \beta \\
        &C_{Y}(\alpha, \beta)=-C_{Y}(\alpha,-\beta), \forall \alpha, \beta
    \end{align}
\end{subequations}

\subsection{Data-driven Aerodynamic model}\label{sec:flight_dynamics3}
\begin{figure}[h!]
    \centering
    \includegraphics[width=1\linewidth]{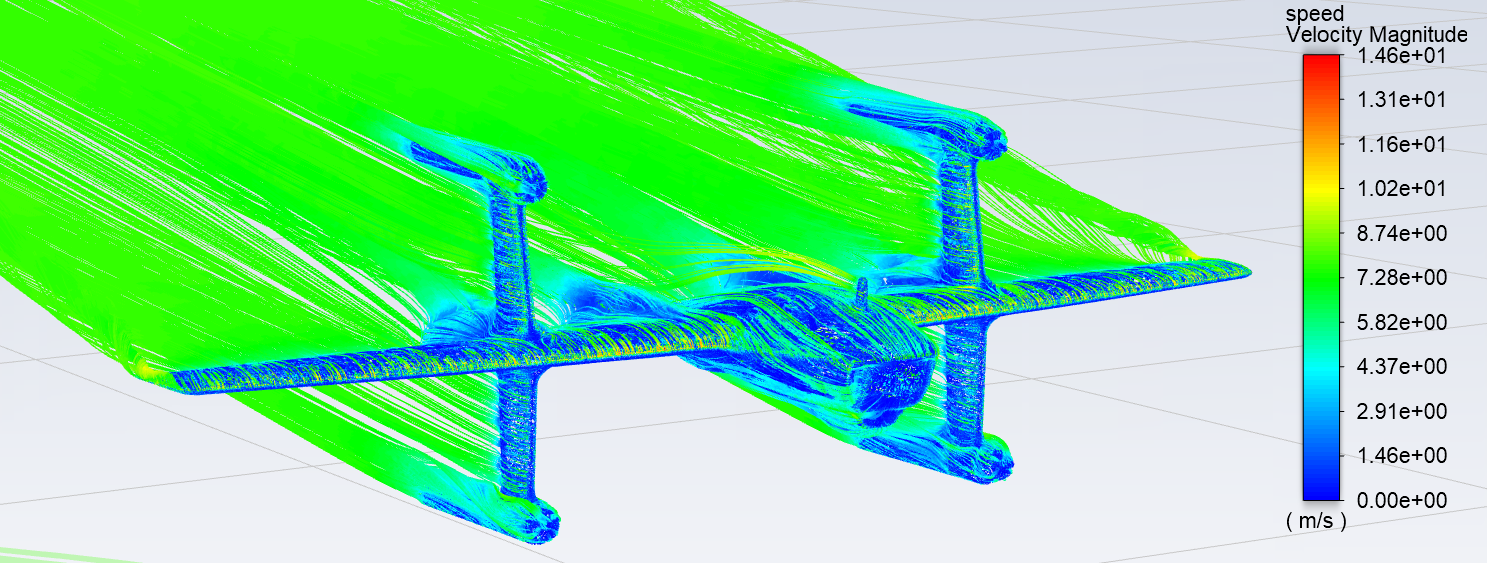}
    \caption{The flow pattern (velocity distribution) of the tail-sitter calculated by CFD simulation. The AoA is 10$\degree$ and the airspeed is 8 $m/s$.}
    \label{fig:Fluent}
\end{figure}

To derive the aerodynamics parameters defined in the previous Section \hyperref[sec:flight_dynamics2]{\Rmnum{2}-B}, the computational fluid dynamics (CFD) are conducted on various AoA and sideslip angles. 
Specifically, the geometric model of the utilized vehicle (without propellers) is obtained by scanning and reconstruction in three dimensions. Then numerical simulations are conducted to predict the the aerodynamic performance of the vehicle.
Aerodynamics analysis is mostly based on CFD simulation or wind tunnel experimental test \cite{bie2021design}. 
Lyu \cite{lyu2018simulation} conducted a full-scale vehicle (without propellers) wind tunnel test to parameterize the aerodynamic coefficient. In his experiment, the wind speed ranged from 2.9 $m/s$ to 18.9 $m/s$. The results indicated that different speeds have a negligible effect on the value of the coefficients when the tail-sitter is working under the full operating speed.
Hence, in our CFD simulation, we only focus on the scenario when the wind speed is 8 $m/s$. To simulate the turbulent flow around the vehicle accurately, the commonly used Spalart-Allmaras model is used. The corresponding turbulence model parameter are set as follow: the y-plus value was set to $1$, the boundary layer was set to 5 layers, the boundary layer growth rate was set to 1.15. Moreover, the pressure-velocity coupling algorithm SIMPLEC was used to solve the Navier-Stokes equations. In the verification of grid independence, a number of grids of 4.57 million will be enough for the CFD simulations. Fig.\ref{fig:Fluent} presents the flow pattern around the vehicle, with the airspeed of 8 $m/s$, AoA of 10 degrees, and sideslip angle of 0 degrees.

\begin{figure*}[h!]
    \centering
    \includegraphics[width=1\textwidth]{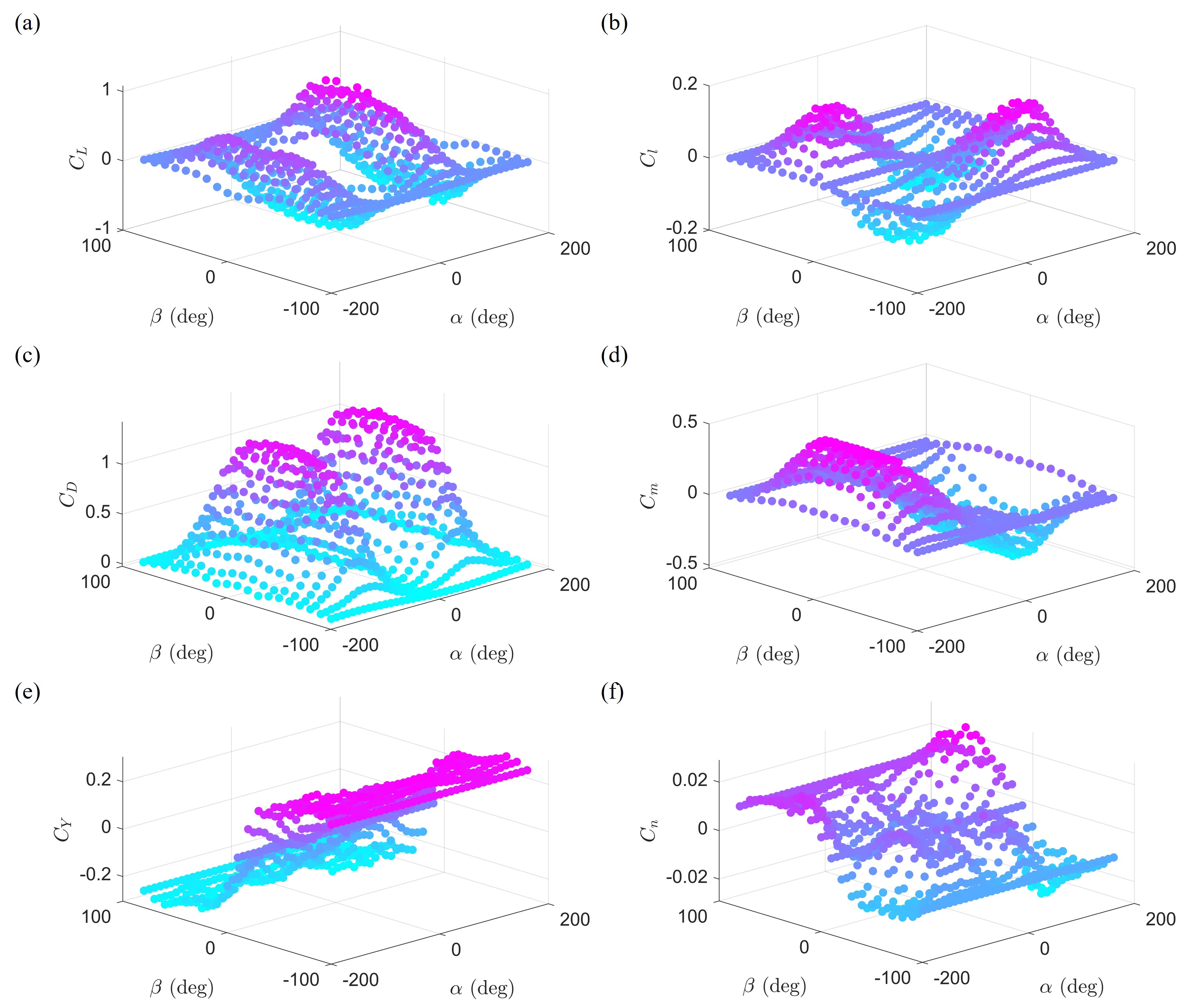}
    \caption{Full envelope aerodynamic coefficients of SWAN K1 PRO tail-sitter prototype, identified by CFD simulation: (a) lift coefficient, (b) rolling moment coefficient, (c) drag coefficient, (d) pitching moment coefficient, (e) side force coefficient, (f) yawing moment coefficient.}
    \label{fig:coefficient}
\end{figure*}

We had collected data at intervals of 10 degrees of AoA and sideslip angles. The final outcomes can be further seen in Fig.\ref{fig:coefficient}. We conduct spline interpolation on the results obtained from CFD, enabling the calculation of aerodynamic coefficients at any angle of attack and sideslip angle. This approach is effective and essential for future realistic flight and simulations \cite{lyu2018simulation}.

It should be noted that we make the assumption that the aerodynamic forces are not affected by the propeller slipstream. This assumption hold true for quadrotor tail-sitters since the propellers are mounted on racks located on either side of the wing, far away from the fuselage.

\section{MULTISTAGE TRAJECTORY OPTIMIZATION}
\label{sec:multi_op}
This section introduces the multistage trajectory optimization problem for tail-sitters. The expression of the problem is explained in detail in Section \hyperref[sec:multi_op1]{\Rmnum{3}-A}. To solve this problem, differential flatness, MINCO trajectory, and penalty functional are used as seen in Section \hyperref[sec:multi_op2]{\Rmnum{3}-B}, Section \hyperref[sec:multi_op3]{\Rmnum{3}-C}, Section \hyperref[sec:multi_op4]{\Rmnum{3}-D}, respectively.  Section \hyperref[sec:multi_op5]{\Rmnum{3}-E} provides a comprehensive explanation and conclusion of our trajectory optimization method. Section \hyperref[sec:multi_op6]{\Rmnum{3}-F} presents an analysis and discussion of the optimisation algorithm.

Notably, the optimization method described in this section necessitates the employment of sampling-based motion planners, such as RRT* \cite{karaman2011sampling} or its derivatives \cite{li2016asymptotically,gammell2018informed}, to overcome the complexity from environments. Our method aims to generate dynamically feasible trajectories that are homotopic to paths created by sampling-based planners. In addition, our method satisfies systems state-input constraints, which are absent in sampling-based planners. This separation allows for the complexity from the environments and dynamics to be addressed independently.

\subsection{Problem Formulation}
\label{sec:multi_op1}
In the present work, the multistage trajectory planning problem is translated into an optimization problem. Assuming there are $M-1$ intermediate waypoints, the overall trajectory can be divided into $M$ segments. The optimization problem is formulated as follows:
\begin{subequations}
    \label{equ:general_problem}
    \begin{align}
        \label{eq:general_problem1}
        & \min \int_{0}^{\sum_{1}^{M} T_i} \mathbf{p}^{(4)}(t)^T\mathbf{p}^{(4)}(t)dt + \Phi(\sum_{1}^{M} T_i)\\
        \label{eq:general_problem3}
        \mathbf{s.t.} \ &   t\in[0,\sum_{1}^{M} T_i]  \\
        \label{eq:general_problem5}
        & \mathbf{p}^{[3]}(0)=\mathbf{s}_0, \ \mathbf{p}^{[3]}(\sum_{1}^{M} T_i)=\mathbf{s}_f\\
        \label{eq:general_problem6}
        & \mathbf{p}(T_i) = \mathbf{s}_i, \ 1 \leq i < M\\
        \label{eq:general_problem4}
        & \left \|\mathbf{v}(t)\right \|<V_{max}\\
         \label{eq:general_problem7}
        & T_i>0, \ 1 \leq i \leq M\\
        \label{eq:general_problem8}
        & \dot{\mathbf{p}} = \mathbf{v}\\
        \label{eq:general_problem9}
        & \dot{\mathbf{v}} = \mathbf{g}+\frac{1}{m}(f\mathbf{R}\mathbf{e}_1+\mathbf{R}\mathbf{f}_a)\\
        \label{eq:general_problem10}
        & \dot{\mathbf{R}} = \mathbf{R}\left \lfloor \boldsymbol{\omega}  \right \rfloor\\
        \label{eq:general_problem11}
        & \mathbf{J}\dot{\boldsymbol{\omega}} = \boldsymbol{\tau}+\mathbf{M}_a - \boldsymbol{\omega}\times\mathbf{J}\boldsymbol{\omega}
    \end{align}
\end{subequations}
where $T_i$ is used to denote the temporal duration of the $i$-th segment of the entire trajectory, $\mathbf{p}^{(x)}(t)$ means the $x$-th derivative of position $\mathbf{p}$. $\mathbf{s}_0$,$\mathbf{s}_f\in\mathbb{R}^{4\times3}$ are constant boundary conditions. $\mathbf{s}_i\in\mathbb{R}^{4\times3}$ is intermediate condition, $\mathbf{p}^{[x]} \in \mathbb{R}^{(x+1)\times3}$ defined by 
\begin{align}
    \mathbf{p}^{[x]}=(\mathbf{p},\dot{\mathbf{p}}, \ldots,\mathbf{p}^{(x)})^T
\end{align}

Equation (\ref{eq:general_problem1}) represents objective function, which demonstrates that we aim to minimize the snap (i.e., the fourth derivative of position) of the trajectory. In practice, minimizing snap optimization roughly corresponds to reducing the required control moment and thus increasing the likelihood that the control input limits are satisfied and the trajectory is feasible \cite{tal2023aerobatic}. To attain a specific harmony between control effort of the generated trajectory and total time, we incorporate a temporal regularization $\Phi$:
\begin{align}
   \Phi(\sum_{1}^{M} T_i)=\sum_{1}^{M}b_{i}T_{i}
\end{align}
where $b_{M}$ is positive.

Equation (\ref{eq:general_problem4}) represents that the tail-sitter's speed must always be under the maximum speed for safe flight. According to (\ref{eq:etkin}), it can be inferred that the aerodynamic forces and moments generated by the fuselage and wings of the tail-sitter are strongly dependent on the speed. In the absence of any speed constraint on the tail-sitter, the trajectory it executes may result in an undesirable AoA. This, in turn, could render the tail-sitter's inputs unachievable or potentially lead to a direct crash. Equations (\hyperref[eq:general_problem8]{11g-11j}) are synonymous with (\ref{equ:tr_r_tail-sitter}), indicating that our trajectory must satisfy the dynamic constraints of the tail-sitter. 

Concluding the above descriptions, our objective is to generate a trajectory that minimizes snap and considers total duration while also subjecting to the maximum speed constraint from the tail-sitter. Additionally, the trajectory must enable the tail-sitter to traverse through intermediate waypoints (generated by sampling-based motion planner). 

\subsection{Differential Flatness}
\label{sec:multi_op2}
First, the differential flatness are used to eliminate the dynamic constraints (\hyperref[eq:general_problem8]{11g-11j}). Consider the following type of the dynamical system:
\begin{align}
    \label{eq:differential_flatness}
    & \dot{\mathbf{x}}= f(\mathbf{x})+g(\mathbf{x})\mathbf{u}
\end{align}
with state $\mathbf{x} \in \mathbb{R}^{n}$, and input $\mathbf{u} \in \mathbb{R}^{m}$. The map $g$ is assumed to have rank $m$. If there exists a flat-output $\mathbf{y}$ such that $\mathbf{x}$ and $\mathbf{u}$ can be represented by finite order derivatives of $\mathbf{y} \in \mathbb{R}^{m}$:
\begin{align}
    & \dot{\mathbf{x}} = \mathcal{A}(\mathbf{y},\dot{\mathbf{y}}, \ldots,\mathbf{y}^{(k)})\\
    & \dot{\mathbf{u}} = \mathcal{B}(\mathbf{y},\dot{\mathbf{y}}, \ldots,\mathbf{y}^{(j)})
\end{align}
then the system is said to be differentially flat \cite{fliess1995flatness}. $\mathcal{A}$ and $\mathcal{B}$ each represent a set of differential flat transformations, determined by function $f$ and $g$. $k$ and $j$ are both natural numbers.
\begin{figure}[h!]
    \centering
    \includegraphics[width=1\linewidth]{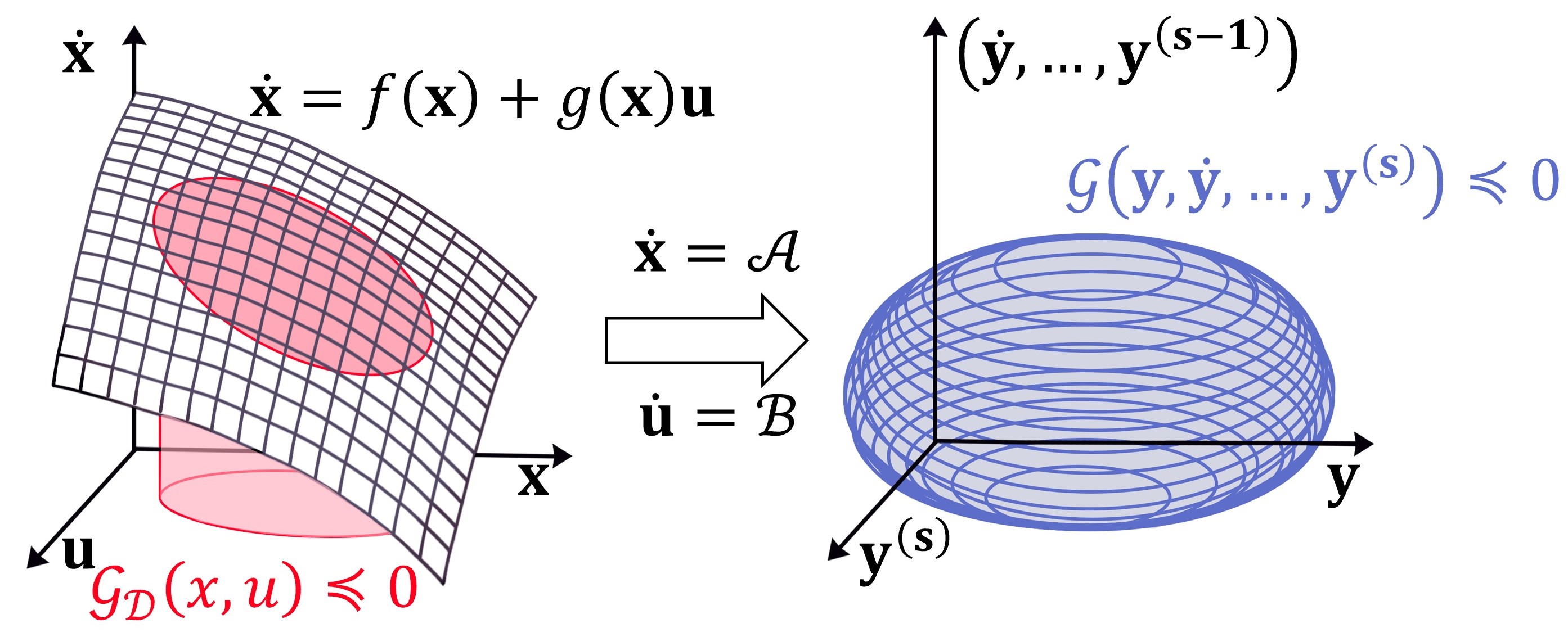}
    \caption{Transform $\mathcal{A}$ and $\mathcal{B}$ of a flat system: Black surface on the left coordinate represents differential constraints from dynamics. Red area on the left coordinate denotes user-defined state-input constraints, such as actuator limits and maximum speed constraint. Blue volume on the right coordinate represents the solution space in the flat output space after eliminating the dynamic constraints.}
    \label{fig:differential_flatness}
\end{figure}

As shown in Fig.\ref{fig:differential_flatness}, the flatness maps $\mathcal{A}$ and $\mathcal{B}$ eliminate the differential constraints (\ref{eq:differential_flatness}). As a side effect, state-input constraints also need to be transformed into the flat-output space, which might brings additional difficulties in trajectory generation. However, if the flat-output is part of the states or inputs, this effect can be relieved. For dynamical systems like those described in (\ref{equ:tr_r_tail-sitter}), and states and inputs defined as in Section \hyperref[sec:flight_dynamics1]{\Rmnum{2}-A}, their differential flatness transformations during coordinated flight have been thoroughly proven by Lu \cite{lu2024trajectory}: the flat-output $\mathbf{y}$ is the position $\mathbf{p}\in \mathbb{R}^3$ of the tail-sitter. Specifically, the coordinated flight refers to a flight where there is no sideslip, as shown in Fig.\ref{fig:coordinateframe}, $\beta=0$ and $v_{a_y}^B=0$. This type of flight consistently places the airspeed encountered by the UAV in the $\mathbf{x}_b\mathbf{O}_b\mathbf{z}_b$ plane of the body frame, but does not restrict the degrees-of-freedom of the tail-sitter in 3-D space. The tail-sitter can achieve any position by utilizing centripetal force generated by rolling for turning. Compared with uncoordinated flight, coordinated flight not only maximizes aerodynamic efficiency but also minimizes potential undesirable aerodynamic torques \cite{stevens2015aircraft}. Therefore, it can greatly reduce the computational load of the trajectory generation algorithm. The differential flatness transformation $\mathcal{A}$ and $\mathcal{B}$ for the tail-sitter during coordinated flight is given in Appendix \hyperref[appendixa]{A}.

Next, trajectory optimization can be performed in the flat-output space, and the continuity order in the corresponding flat-output space can ensure that the dynamical differential constraints are precisely satisfied, which means the problem formulation is reduced to (\hyperref[equ:general_problem]{11a-11f}).

\subsection{Minimum Snap Trajectory Generation}
\label{sec:multi_op3}
The problem formulation using differential flatness  still involves one optimization problem and some constraints. In this subsection, the MINCO method is used to solve the linear quadratic optimization problem (\ref{eq:general_problem1}) with waypoints constraints (\hyperref[eq:general_problem3]{11b-11d}).

The research on linear quadratic optimization problem (\ref{eq:general_problem1}) mostly focuses on single-stage problems under different cost functions \cite{mueller2015computationally,liu2017search}. Wang \cite{wang2022geometrically} proposed optimality conditions for unconstrained trajectory optimization problems. Leveraging these conditions, an unconstrained trajectory optimization problem can be solved in linear complexity of time and space, without evaluating the cost functional (\ref{eq:general_problem1}) explicitly or implicitly. Specifically, consider an $M$-stage trajectory optimization problem in the (\hyperref[eq:general_problem1]{11a-11d}) form, the MINCO method is used to solve this propblem.

Here we define time vector $\mathbf{T}\in\mathbb{R}^M$ as
\begin{align}
    & \mathbf{T}=(T_1,T_2, \ldots,T_M)^T
\end{align}
The trajectory could be characterized by a multistage polynomial with respect to $t$, where a 7th-degree polynomial trajectory with $C^4$ continuity is used to connect two consecutive points $\mathbf{p}_j,\;\mathbf{p}_{j+1}$ at their respective passing time $T_j,\;T_{j+1}\in\mathbf{T}$. The entire trajectory is therefore uniquely determined by all waypoints and respective passing time $\mathbf{T}$, having the endpoint constraint (\ref{eq:general_problem5}) and intermediate waypoint constraint (\ref{eq:general_problem6}) naturally satisfied.
Therefore, the expression for the $i$th piece of a 3-dimensional trajectory is denoted by:
\begin{align}
    & \mathbf{p}_i(t)=\mathbf{c}_i^T\rho(t), t\in[0,T_i]
\end{align}
where $\mathbf{c}_i\in\mathbb{R}^{8\times3}$ are the coefficients of 7th-degree polynomials, $\rho(x)=(1, x, x^2,\ldots, x^7)^T$. To simplify, we utilize relative time for each phase of the trajectory, meaning that the initial time for each phase is set to 0. The states of the vehicle at time $T_i$ is exactly the same as it was at the start of the following segment. Hence, the trajectory could be described by a coefficient matrix $\mathbf{c}\in\mathbb{R}^{8M\times3}$ defined by:
\begin{align}
    & \mathbf{c}=(\mathbf{c}_1, \mathbf{c}_2, \ldots, \mathbf{c}_M)^T
\end{align}

To acquire the matrix $\mathbf{c}$, we establish a linear system:
\begin{align}
    & \mathbf{A}\mathbf{c}=\mathbf{b}
    \label{eq:coeffi}
\end{align}
where $\mathbf{A}\in\mathbb{R}^{8M\times8M}$ and $\mathbf{b}\in\mathbb{R}^{8M\times3}$ are:
\begin{align}
    \mathbf{A}=&\begin{pmatrix}
      &\mathbf{H}_0  &\mathbf{0}  &\mathbf{0}  &\ldots  &\mathbf{0} \\
      &\mathbf{G}_1  &\mathbf{H}_1  &\mathbf{0}  &\ldots  &\mathbf{0} \\
      &\mathbf{0}  &\mathbf{G}_2  &\mathbf{H}_2  &\ldots  &\mathbf{0} \\
      &\vdots  &\vdots  &\vdots  &\ddots &\vdots \\
      &\mathbf{0}  &\mathbf{0}  &\mathbf{0}  &\ldots  &\mathbf{H}_{M-1}\\
      &\mathbf{0}  &\mathbf{0}  &\mathbf{0}  &\ldots  &\mathbf{G}_M
    \end{pmatrix}\\
     \mathbf{b}=&({\mathbf{p}_0^{[3]}}^T, \mathbf{p}_1, \mathbf{0}^{3\times7}, \ldots, \mathbf{p}_{M-1}, \mathbf{0}^{3\times7}, {\mathbf{p}_f^{[3]}}^T)^T
\end{align}

The matrix $\mathbf{A}$, which is nonsingularity, can be regarded as a function of the time vector $\mathbf{T}$.  $\mathbf{H}_0$, $\mathbf{G}_M\in\mathbb{R}^{4\times8}$ are defined as:
\begin{align}
    & \mathbf{H}_0=(\rho(0), \dot{\rho}(0), \ldots, \rho^{(3)}(0))^T\\
    & \mathbf{G}_M=(\rho(T_M), \dot{\rho}(T_M), \ldots, \rho^{(3)}(T_M))^T
\end{align}
and  $\mathbf{H}_i$, $\mathbf{G}_i\in\mathbb{R}^{8\times8}$  are defined as:
\begin{align}
    & \mathbf{H}_i=(\mathbf{0}^{8\times1}, -\rho(0), -\dot{\rho}(0), \ldots, -\rho^{(6)}(0))^T\\
    & \mathbf{G}_i=(\rho(T_i), \rho(T_i), \dot{\rho}(T_M), \ldots, \rho^{(6)}(T_M))^T
\end{align}
whereas the matrix $\mathbf{b}$ can be seen as a function of the spatial parameter $\mathbf{p}$. By solving (\ref{eq:coeffi}), one and only minimum snap trajectory can be obtained.
\begin{figure}[]
    \centering
    \includegraphics[width=1\linewidth]{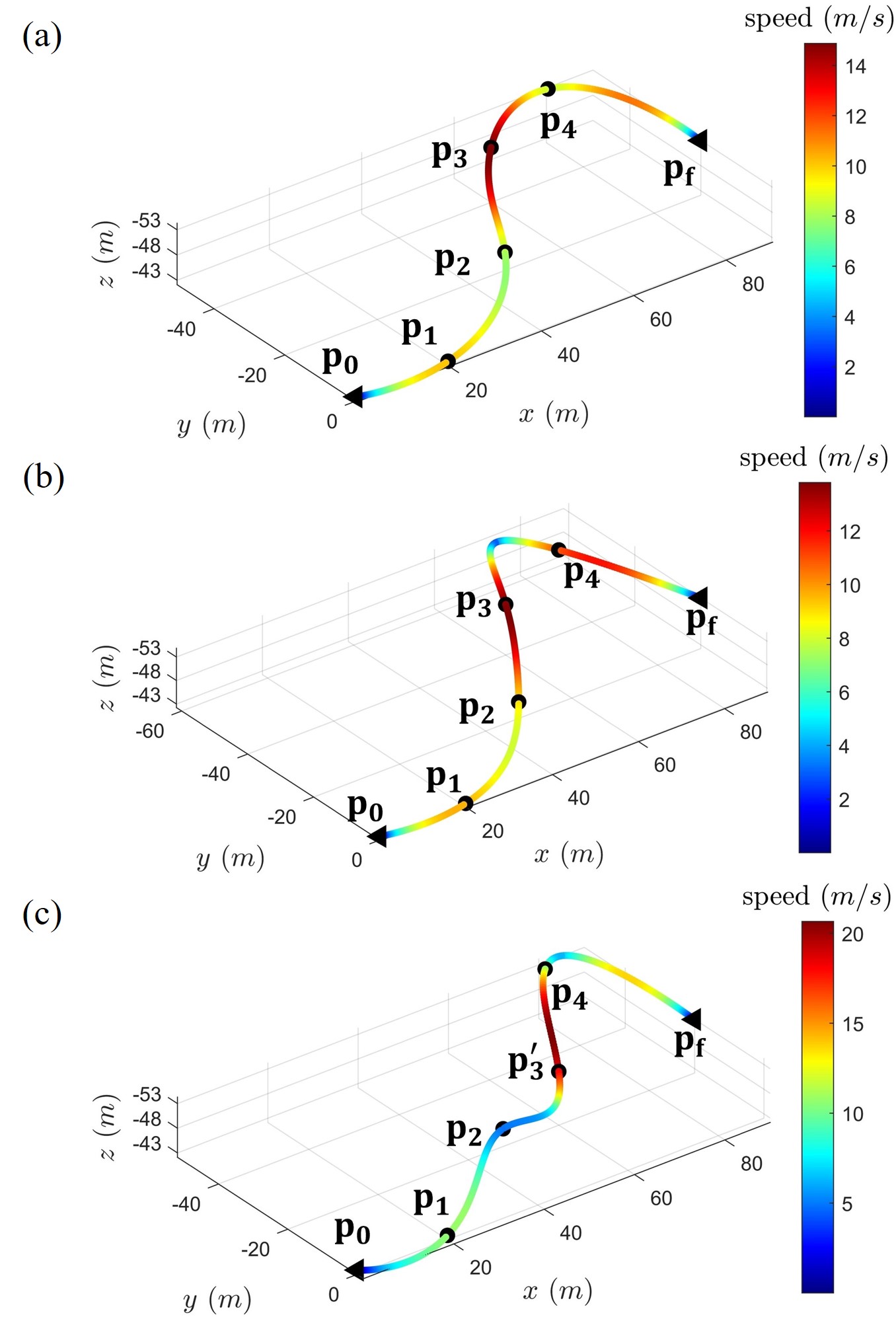}
    \caption{Minimum snap trajectory with 4 intermediate waypoints: (a) and (b) have the same waypoints positions, but the temporal parameter $T_4$ is different. (a) and (c) have the same temporal parameter $\mathbf{T}$, but the waypoint position $\mathbf{p}_3$ is different.}
    \label{fig:minsnap}
\end{figure}

To validate the effectiveness of MINCO method, a four intermediate waypoints trajectory generation problem without any other constraints is tested. Fig.\ref{fig:minsnap} shows the minimum snap trajectories using different temporal and spatial parameters. The black triangles represent the starting and ending locations, while the black dots represent the positions of intermediate waypoints. 
First, Fig.\hyperref[fig:minsnap]{6b} presents the effect of temporlal parameter $T_4$ between waypoints $\mathbf{p}_3$ and $\mathbf{p}_4$ while maintaining the positions constant, in comprison to the trajectory shown in Fig.\hyperref[fig:minsnap]{6a}. Second, the effect of the position of $\mathbf{p}_3$ while maintaining a constant time interval between waypoints is depicted in Fig.\hyperref[fig:minsnap]{6c}. It can be seen that a strong correlation between the vehicle's speed and the spatial parameters $\mathbf{p}$ as well as the temporal parameters $\mathbf{T}$. Any small alteration in any of these factors not only impacts the speed of the vehicle in its entire trajectory, but also results in a modification of the overall shape of the trajectory.

\subsection{Transformation of Maximum Speed Constraint and Temporal Constraint Elimination}
\label{sec:multi_op4}

In this subsection, we will introduce how to transform the maximum speed constraint and eliminate temporal constraint. 

First, the maximum speed constraint (\ref{eq:general_problem4}) necessitates that the inequality is fulfilled at every point along the trajectory. Given that the trajectory position is represented by $M$-stage 7th-degree polynomials, it is necessary to determine the highest value of $M$ distinct 6th-degree polynomials on $M$ separate convex sets in order to meet the maximum speed constraint. This task is quite demanding. 
The penalty functional is a straightforward and effective approach where the constraints have obvious physical interpretations and precision requirements are not stringent. Furthermore, penalty functional don't require a feasible initial guess, which can be challenging to construct. 
To reduce computing burden while maintaining trajectory quality, the trajectory may be discretely parameterized. To guarantee the fulfillment of the continuous-time constraint $V_ {max}$ at a specific resolution, we adopt temporal discretization in the time domain. 
We define:
\begin{align}
    \label{eq:potential}
    & P(||\mathbf{v}||)=\sum_{i=1}^{M}\sum_{j=1}^{N} \max[V_{ij}-V_{max}, 0]^3
\end{align}
where $N$ controls the resolution of trajectory, $V_{ij}$ represents the speed at a sampled point on the trajectory. As $N$ increases, the probability of the trajectory breaching the speed constraint at any given moment decrease. Moreover, the cubic form of the function forms a differentiable strictly convex penalty.

As for the temporal constraint (\ref{eq:general_problem7}), it is evident that $\mathbf{T}\in\mathbb{R}_{>0}^{M}$ is a hard constraint, which restrict the domain of $\mathbf{T}$ to simple manifolds. The optimization on the manifold frequently requires retractions, therefor an explicit diffeomorphisms for $\mathbf{T}$ is given by defining:
\begin{align}
    & \mathbf{T}=e^{\mathbf{q}}
\end{align}
where $\mathbf{q}=(q_1, q_2, \ldots, q_M)$. As a result, we can directly optimize the unconstrained surrogate variables $\mathbf{q}$ in Euclidean space. It is worth noting that $\Phi$ is the sum of linear functions with respect to each piece of trajectory duration $T_i$, and is thus a convex function. The function $f(x)=e^x$ is also a convex function over its domain $x\in R$. Since the sum of convex functions is still convex, setting $\mathbf{T}=e^{\mathbf{q}}$ does not alter the convexity of the time regularization $\Phi$.

\subsection{Optimization}
\label{sec:multi_op5}

Combining all methods above, we can solve a lightweight relaxed optimization via unconstrained linear programming (LP), which is defined as
\begin{align}
    \label{eq:final}
    & h(\mathbf{q})=\min \sum_{i=1}^{M} b_i e^{q_i} + wP(||\mathbf{v}||)
\end{align}
where $w\in\mathbb{R}_{>0}$ is a weight number which should be a large constant. The original problem (\ref{equ:general_problem}), as evident, has been converted from an optimization problem to an unconstrained optimal problem (\ref{eq:final}). The speed of the vehicle is affected by the time interval between two waypoints. Therefore, we select $\mathbf{q}$ as the decision variable for (\ref{eq:final}).

Modifications to any component within the temporal parameter $\mathbf{T}$ will result in modifications to the properties of the overall trajectory, as illustrated in Fig.(\ref{fig:minsnap}). The challenge in multistage trajectory optimization is in finding the gradient of equation (\ref{eq:potential}) with respect to the time vector $\mathbf{T}$. With available gradient, the relaxation (\ref{eq:final}) could be solved by the limited-memory Broyden-Fletcher-Goldfarb-Shanno algorithm \cite{liu1989limited}. The trajectory generation can be achieved by combining all the aforementioned parts, as demonstrated in Algorithm \hyperref[algorithm]{1}.
\phantomsection
\label{algorithm}
\begin{algorithm}
    \caption{Multistage trajectory optimization}
    \textbf{Given: }Initial conditions $\mathbf{s}_0$, final conditions $\mathbf{s}_f$, waypoints $\mathbf{p}_i$ and $V_{max}$.\\
    \textbf{L-BFGS: }\\
    Initialize $\mathbf{T}^0\leftarrow e^{\mathbf{q}^0}$, $\mathbf{g}^0\leftarrow\nabla h(\mathbf{q}^0)$, $\mathbf{B}^0\leftarrow\mathbf{I}$, $k\leftarrow0$\\
    \While{$||\mathbf{g}^k||>\delta$}
    {$\mathbf{T}^k\leftarrow e^{\mathbf{q}^k}$\;
    Calculating the trajectory generated by the temporal parameter $\mathbf{T}^k$ using MINCO\;
    By discretizing the velocity of the obtained trajectory, we derive the penalty functional (\ref{eq:potential})\;
    $\mathbf{g}^k\leftarrow\nabla h(\mathbf{q}^k)$\;
    $\mathbf{d}\leftarrow-\mathbf{B}^k\mathbf{g}^k$\;
    $\mathbf{t}\leftarrow$ Lewis Overton line search\; 
    $\mathbf{q}^{k+1}\leftarrow\mathbf{q}^k+\mathbf{t}\mathbf{d}$\;
    $\mathbf{B}^{k+1}\leftarrow$ Cautious-Limited-Memory-BFGS$(\mathbf{g}^{k+1}-\mathbf{g}^k, \mathbf{q}^{k+1}-\mathbf{q}^k)$\;
    $k\leftarrow k+1$\;}
    Solve angle of attack $\alpha$ by Newton-Raphson method.\\
    Solve states $\mathbf{x}_{full}$ and inputs $\mathbf{u}_{full}$ by differential flatness transformation.
\end{algorithm}

\subsection{Analysis and Discussion}
\label{sec:multi_op6}
Resolving the root-finding problem for $\alpha$ is an integral part of the proposed method (see to Appendix \hyperref[appendixa]{A} for further information). In this subsection, a robust and validated solution for this issue is presented and tested to substantiate the methodology. 

\begin{figure}[h!]
    \centering
    \includegraphics[width=1\linewidth]{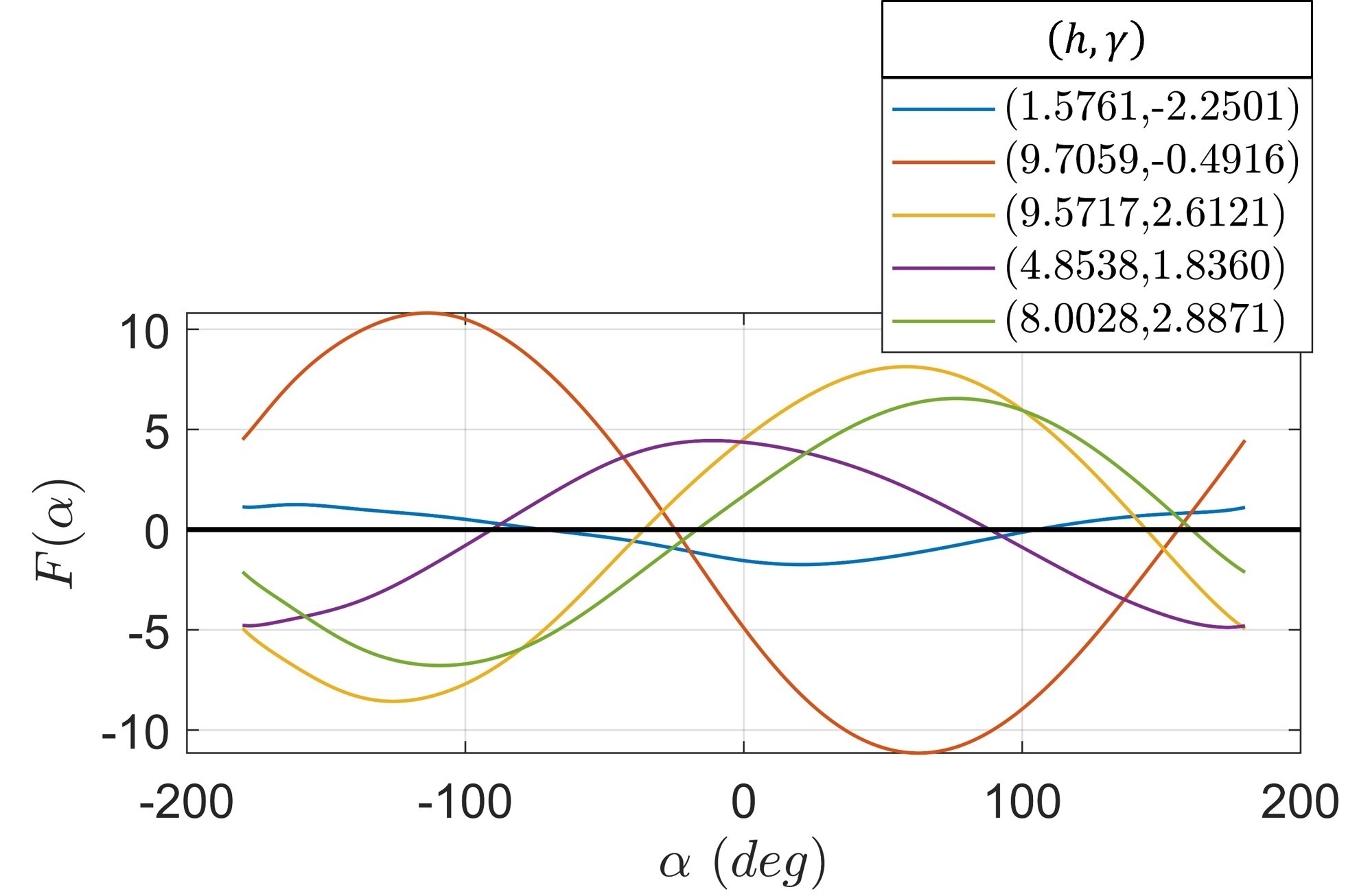}
    \caption{Root-finding problem of $F(\alpha)=0$ for five random pairs of $h$ and $\gamma$.}
    \label{fig:newtonraphson}
\end{figure}

First, in the root-finding problem for $\alpha$, the Newton-Raphson method is used and validated. Specifically, the target is to find the value of $\alpha$ that satisfies (\ref{eq:app5}), where $h$ and $\gamma$ are expressed by (\ref{eq:app4}) and (\ref{eq:app3}), and $\mathbf{C}_z(\alpha,0)$ is a function of $\alpha$. Fig.\ref{fig:newtonraphson} shows the $F(\alpha)$ with respect to $\alpha$ for five random sets of $h$ and $\gamma$. It can be seen from the figure that each $F(\alpha)$ has more than one zero point, but the zero points are far apart. The essence of the Newton-Raphson method is similar to the steepest gradient descent in optimization problem, that is, starting from any point, even if the object is a non-smooth function, it can descend in a certain direction with a certain step size until convergence. This means that when there are multiple zero points, the initial guess for $\alpha$ determines which one it will eventually converge to. 
Therefore, we use the following two approaches to prevent finding inappropriate roots. First, since the calculation of a vertical take-off and landing UAV trajectory generally begins in the hovering state, we can set the initial guess value of $\alpha$ to $90$ at the beginning time, which is an accurate root. Second, since the trajectory is represented by polynomials in time, this means that the next root will be nearly the same as the current root. We can take the current $\alpha$ value as the initial guess for the next root-finding iteration, thereby ensuring the accuracy of root finding and greatly improving the convergence rate. 
By utilizing these two techniques, we can reliably use the Newton-Raphson method to obtain the accurate $\alpha$ of a trajectory every time step and ensure that $\alpha$ will not converge to irrational values.
\begin{figure}[h]
    \centering
    \includegraphics[width=1\columnwidth]{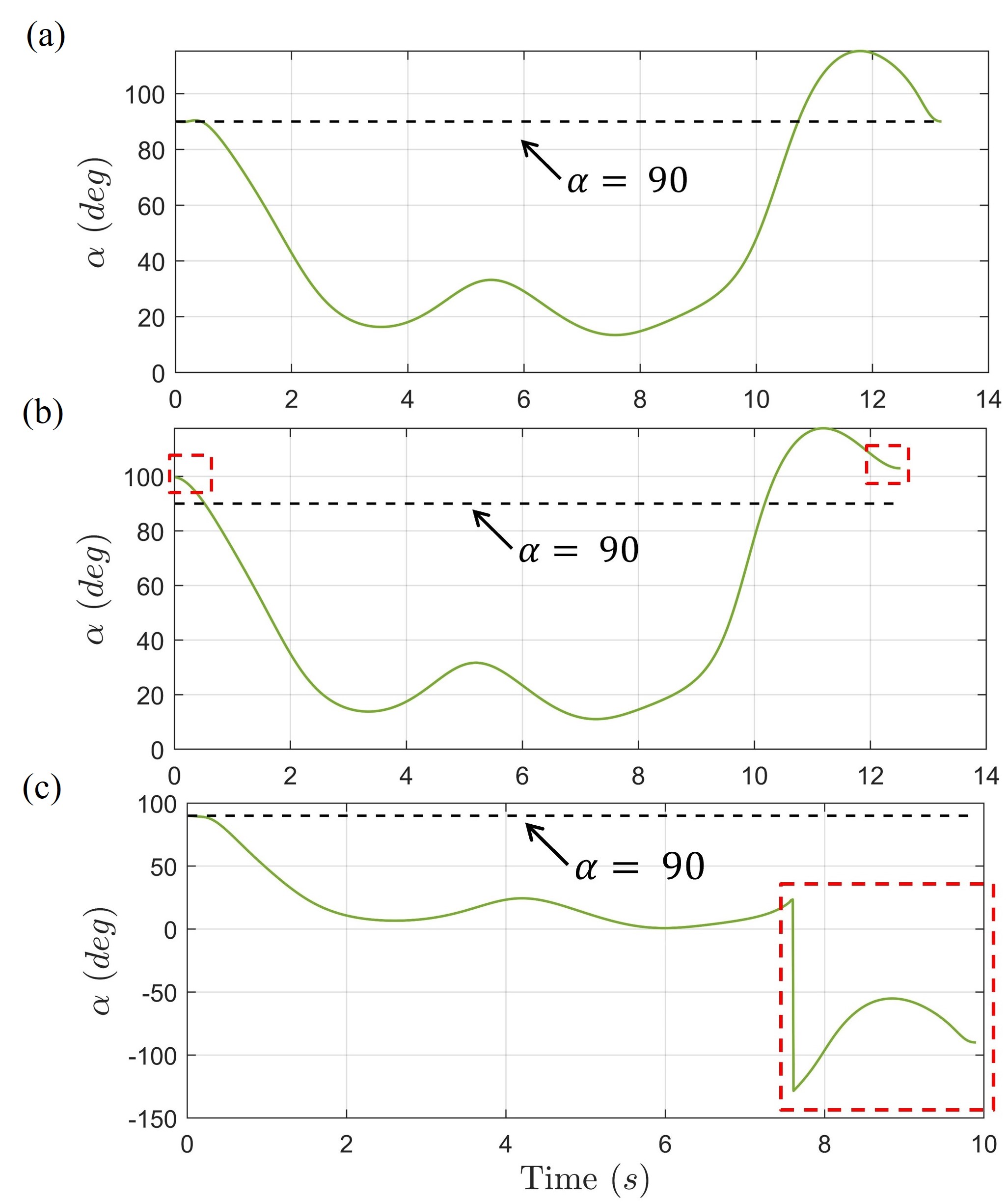}
    \caption{AoA of a trajectory within a vertical plane: (a) normal condition, (b) the trajectory of AoA if the start and final speed was set to 0, (c) the trajectory of AoA if the acceleration $\|\mathbf{a}\|$ is too large.}
    \label{fig:alpha_ab}
\end{figure}

Furthermore, it is found that poor $h$ and $\gamma$ values can still lead to abnormal $\alpha$ value and result in disastrous consequences, since the control inputs of the tail-sitter are all calculated based on its attitude.
Some specific scenarios are presented in Fig.\ref{fig:alpha_ab}. Fig.\hyperref[fig:alpha_ab]{8a} illustrates the angle of attack of a normal trajectory where the tail-sitter is flying forward and ascending in a vertical plane.Fig.\hyperref[fig:alpha_ab]{8b} illustrates the abnormal AoA at the beginning and final phases of the trajectory. Fig.\hyperref[fig:alpha_ab]{8c} shows the impractical AoA found for the tail-sitter during the intermediate phases. 
Through the comparisons with the normal case in Fig.\hyperref[fig:alpha_ab]{8a}, we examine the reasons and propose solutions for other two cases. First, Fig.\hyperref[fig:alpha_ab]{8b} depicts a scenario where the instability in the numerical solution arises from tail-sitter's extremely low speed (according to (\ref{eq:app4}), the value of $h$ thus becomes extremely large) at the beginning and ending phases. Therefore, assigning a low value to the speed at both the initial and final states during the trajectory generation could fix this issue effectively. Viewed differently, the abnormal $\alpha$ value at the start of the trajectory does not impact the root of $F(\alpha)$ once $h$ normalizes, highlighting the robustness of the Newton-Raphson method. Second, Fig.\hyperref[fig:alpha_ab]{8c} indicates a stall scenario, which is due to the tail-sitter requiring a significant acceleration within a specific time interval and it exceeds the capacity of the fuselage and wings to provide the necessary force. Therefore, this issue can be solved by appropriately decreasing the value of the maximum speed constraint.

\section{MODEL PREDICTIVE CONTROL FOR TRAJECTORY TRACKING}
\label{sec:MPC}
In this section, we introduce a trajectory tracking controller that enables the quadrotor tail-sitter to track agile and smooth trajectories accurately in 3-dimensional space. First, Section \hyperref[sec:MPC1]{\Rmnum{4}-A} presents the controller, which is a global controller for the tail-sitter based on MPC. Then, Section \hyperref[sec:MPC2]{\Rmnum{4}-B} demonstrates the ability of our methods to generate and track optimal trajectories in both 3-dimensional Euclidean space and 2-dimensional vertical plane.

\subsection{Global Controller for the Tail-sitter}
\label{sec:MPC1}
Model Predictive Control (MPC) is a common and powerful trajectory tracking method in robotics. Generally, we can design MPC to track the trajectory of a model whose states are all in Euclidean space (such as the unicycle model). This is because all system states are flat vectors, and by calculating their error's norm, we can quantify the tracking error. However, for the dynamic model shown in (\ref{equ:tr_r_tail-sitter}), tail-sitter’s states contain the rotation matrix $\mathbf{R}\in\mathbb{R}^{3\times3}$, which is not in Euclidean space but on a non-vector, curved manifold. This makes it difficult to quantify the rotation matrix error $\delta \mathbf{R}$. Inspired by \cite{lu2022manifold}, the rotation matrix $\mathbf{R}$ can be projected into Euclidean space by mapping:
\begin{align}
    & \boldsymbol{\theta} = Log(\mathbf{R})\in\mathbb{R}^3
\end{align}
where $Log(\cdot)$ is the logarithmic map proposed by \cite{bullo1995proportional}, which is defined as:
\begin{align}
    & Log_{SO(3)}(\mathbf{R}) =  \frac{\phi}{2\sin\phi}\begin{bmatrix}
        R_{32}-R_{23} \\
        R_{13}-R_{31} \\
        R_{21}-R_{12}\end{bmatrix}
\end{align}
where $\phi$ satisfies $\cos(\phi)=\frac{1}{2}(tr(\mathbf{R})-1)$.

Furthermore, $\delta\mathbf{R}$ can be formulated as:
\begin{align}
    \label{eq:Log}
    & \delta\mathbf{R} = \mathbf{R}_r\boxminus\mathbf{R}=Log(\mathbf{R}^T\mathbf{R}_r)
\end{align}
where $\mathbf{R}_r$ is the rotation matrix on the reference trajectory, $\mathbf{R}$ is the actual rotation matrix. The error of the rotation matrix is represented as a 3-dimensional vector, and calculating its norm can track the attitude of the vehicle.

Next, we construct the objective function of MPC to ensure optimal performance in our control strategy. Lu \cite{lu2024trajectory} pointed out that the angular velocity of the tail-sitter can be tracked by three decoupled proportional–integral–derivative (PID) controllers in the autopilot. The Coriolis term $\boldsymbol{\omega}\times \mathbf{J}\boldsymbol{\omega}$ and aerodynamic moment $\mathbf{M}_a$ can be regarded as unknown disturbances and can be compensated in a feed-forward way. Thus, in terms of controller design, the tail-sitter's states and inputs can be simplified to:
\begin{subequations}
    \label{eq:reduced}
    \begin{align}
        &\mathbf{x}=\begin{bmatrix}\mathbf{p}^T\quad\mathbf{v}^T\quad\mathbf{R}^T\end{bmatrix}^T\\
        &\mathbf{u}= \begin{bmatrix}f\quad\boldsymbol{\omega}^T\end{bmatrix}^T
    \end{align}
\end{subequations}

According to (\ref{eq:Log}) and (\ref{eq:reduced}), the error of states and inputs can be represented as:
\begin{subequations}
    \label{eq:error-state}
    \begin{align}
        &\begin{cases}\delta\mathbf{x}=\begin{bmatrix}\delta\mathbf{p}^T\quad\delta\mathbf{v}^T\quad\delta\mathbf{R}^T\end{bmatrix}^T\in\mathbb{R}^9\\ 
        \delta\mathbf{u}= \begin{bmatrix}\delta f\quad\delta\boldsymbol{\omega}^T\end{bmatrix}^T\in\mathbb{R}^4\end{cases}\\
        &\delta\mathbf{p} = \mathbf{p}_r-\mathbf{p} \in \mathbb{R}^3\\
        &\delta\mathbf{v} = \mathbf{v}_r-\mathbf{v} \in \mathbb{R}^3\\
        &\delta\mathbf{R} = \mathbf{R}_r\boxminus\mathbf{R}=Log(\mathbf{R}^T\mathbf{R}_r) \in \mathbb{R}^3\\
        &\delta f = f_r-f \in \mathbb{R}\\
        &\delta\boldsymbol{\omega} = \boldsymbol{\omega}_r-\boldsymbol{\omega} \in \mathbb{R}^3
    \end{align}
\end{subequations}
where the subscript $r$ represents the value on the reference trajectory, and no subscript represents the actual value. Therefore, we can construct an MPC as an optimization problem:
\begin{align}
\label{eq:mpc}
\begin{split}
{\delta \mathbf{u}}_k^{*}=\arg \min _{\delta \mathbf{u}_{k}}& \sum_{k=0}^{N-1}\left(\left\| {\mathbf{x}_r}_{k+1}-\mathbf{x}_{k+1}\right\|_{\mathrm{Q}_{k}}^{2}+\left\|\delta \mathbf{u}_{k}\right\|_{\mathrm{P}_{k}}^{2}\right)\\
\text { s.t. } \quad \delta \mathbf{x}_{k+1}=&\left(\mathbf{I}+\Delta t\mathbf{F}_{\mathbf{x}_k}\right)\delta \mathbf{x}_k+\Delta t \mathbf{F}_{\mathbf{u}_k} \delta \mathbf{u}_k\\
&+\Delta t \mathbf{F}_{\mathbf{w}_k} \delta \mathbf{w}_k  \\
\mathbf{u}_{\text{min}}\leq&{\mathbf{u}_r}_k-\delta\mathbf{u}_k\leq\mathbf{u}_{\text{max}}
\end{split}
\end{align}
where $N$ is receding horizon, $\mathrm{Q}_k$ and $\mathrm{P}_k$ are positive diagonal matrices, $\mathbf{u}_{\text{min}}$ and $\mathbf{u}_{\text{max}}$ denote actual input constraints, $\mathbf{F}_{\mathbf{x}}$, $\mathbf{F}_\mathbf{u}$, and $\mathbf{F}_\mathbf{w}$ are Jacobians of the error-state dynamic model $\dot{\delta\mathbf{x}}$ with respect to $\delta\mathbf{x}$, $\delta\mathbf{u}$ and $\delta\mathbf{w}$, respectively. $\delta\mathbf{w}=\mathbf{w}_r-\mathbf{w}$ where $\mathbf{w}$ is the wind velocity in the environment, and $\mathbf{w}_r$ is the wind velocity used in trajectory optimization. The objective function should be as small as possible so that the states of a trajectory are well tracked and the trajectory inputs are as smooth as possible. Finally, the optimal control of a trajectory is:
\begin{align}
    \mathbf{u}^{*} = \mathbf{u}_r - {\delta\mathbf{u}}^{*}
\end{align}

To obtain $\mathbf{F}_{\mathbf{x}}$, $\mathbf{F}_\mathbf{u}$, and $\mathbf{F}_\mathbf{w}$, we take the derivative of error of states with respect to time to obtain the error dynamic model:
\begin{subequations}
    \label{eq:errordynamic}
    \begin{align}
        \dot{\delta\mathbf{p}} =& \delta\mathbf{v}\\
        \dot{\delta\mathbf{v}} =& \frac{1}{m}\left(f_r\mathbf{R}_r\mathbf{e}_1+\mathbf{R}_r{\mathbf{f}_a}_r-(f\mathbf{R}\mathbf{e}_1+\mathbf{R}{\mathbf{f}_a})\right)\\
        \label{eq:dotdeltaR}
        \dot{\delta\mathbf{R}} =& \mathbf{A}^{-T}(\delta\mathbf{R})\left(-{\mathbf{R}_r}^T \mathbf{R} \boldsymbol{\omega}+\boldsymbol{\omega}_{r}\right)
    \end{align}
\end{subequations}
where ${\mathbf{f}_a}_r$ and $\mathbf{f}_a$ are the aerodynamic forces on the reference trajectory and actual trajectory, respectively. $\mathbf{A}(\cdot):\mathbb{R}^3\rightarrow\mathbb{R}^{3\times3}$ represents a map \cite{bullo1995proportional}:
\begin{align}
\begin{split}
    \mathbf{A}(\delta \mathbf{R}) =&\mathbf{I}+\left(\frac{1-\cos \|\delta \mathbf{R}\|}{\|\delta \mathbf{R}\|}\right) \frac{\lfloor\delta \mathbf{R}\rfloor}{\|\delta \mathbf{R}\|}\\
&+\left(1-\frac{\sin \|\delta \mathbf{R}\|}{\|\delta \mathbf{R}\|}\right) \frac{\lfloor\delta \mathbf{R}\rfloor^{2}}{\|\delta \mathbf{R}\|^{2}}
\end{split}
\end{align}

See Appendix \hyperref[appendixd]{D} for the proof of (\ref{eq:dotdeltaR}).

Linearizing (\ref{eq:errordynamic}):
\begin{align}
    \dot{\delta\mathbf{x}}=\mathbf{F}_{\mathbf{x}} \delta \mathbf{x}+\mathbf{F}_{\mathbf{u}} \delta \mathbf{u}+\mathbf{F}_{\mathbf{w}} \delta \mathbf{w}
\end{align}
where
\begin{subequations}
    \label{eq:FxFu}
    \begin{align}
    \mathbf{F}_{\mathbf{x}}&\approx\begin{bmatrix}
                        \mathbf{0} & \mathbf{I} & \mathbf{0} \\
                        \mathbf{0} & \frac{1}{m} \mathbf{R}_{r} \frac{\partial {\mathbf{f}_a}_r}{\partial {\mathbf{v}_a^B}_r} {\mathbf{R}_r}^{T} &  \frac{\partial\delta\dot{\mathbf{v}}}{\partial\delta\mathbf{R}}\\
                        \mathbf{0} & \mathbf{0} & -\left \lfloor\boldsymbol{\omega} \right \rfloor-\frac{1}{2}\left \lfloor \delta\boldsymbol{\omega} \right \rfloor+\frac{1}{2}\mathbf{K}
                        \end{bmatrix}\\
    \mathbf{F}_{\mathbf{u}}&\approx\begin{bmatrix}
                                    \mathbf{0} & \mathbf{0}\\
                                    \mathbf{R}_r\mathbf{e}_1 &  \mathbf{0}\\
                                    \mathbf{0} & \mathbf{I}+\frac{1}{2}\left \lfloor\delta\mathbf{R} \right \rfloor
                                                \end{bmatrix}\\
    \mathbf{F}_{\mathbf{w}}&\approx\begin{bmatrix}
    \mathbf{0}\\
    -\frac{1}{m} \mathbf{R}_{r} \frac{\partial {\mathbf{f}_a}_r}{\partial {\mathbf{v}_a^B}_r} {\mathbf{R}_r}^{T}\\
    \mathbf{0} 
    \end{bmatrix}\\
    \frac{\partial\delta\dot{\mathbf{v}}}{\partial\delta\mathbf{R}}&\approx\mathbf{R}_r\left(-a_T\left\lfloor\mathbf{e}_{1}\right\rfloor-\frac{1}{m}\left\lfloor\mathbf{f}_a\right\rfloor+\frac{1}{m}\frac{\partial {\mathbf{f}_a}_r}{\partial {\mathbf{v}_a^B}_r}\left\lfloor{\mathbf{R}_r}^T\mathbf{v}_a\right\rfloor\right)
    \end{align}
\end{subequations}
where the matrix $\mathbf{K}$ and the proof of (\ref{eq:FxFu}) are given in Appendix \hyperref[appendixe]{E}.

In conclusion, (\ref{eq:mpc}) can be seen as an optimization problem with an inequality constraint, which can be solved by the Powell-Hestenes-Rockafellar augmented Lagrangian method (PHR-ALM) \cite{rockafellar1973dual}.

\subsection{Validations of both the Trajectory Generation and Tracking in the Vertical Plane and 3-Dimensional Space}
\label{sec:MPC2}
\begin{figure*}[]
    \centering
    \includegraphics[width=1\textwidth]{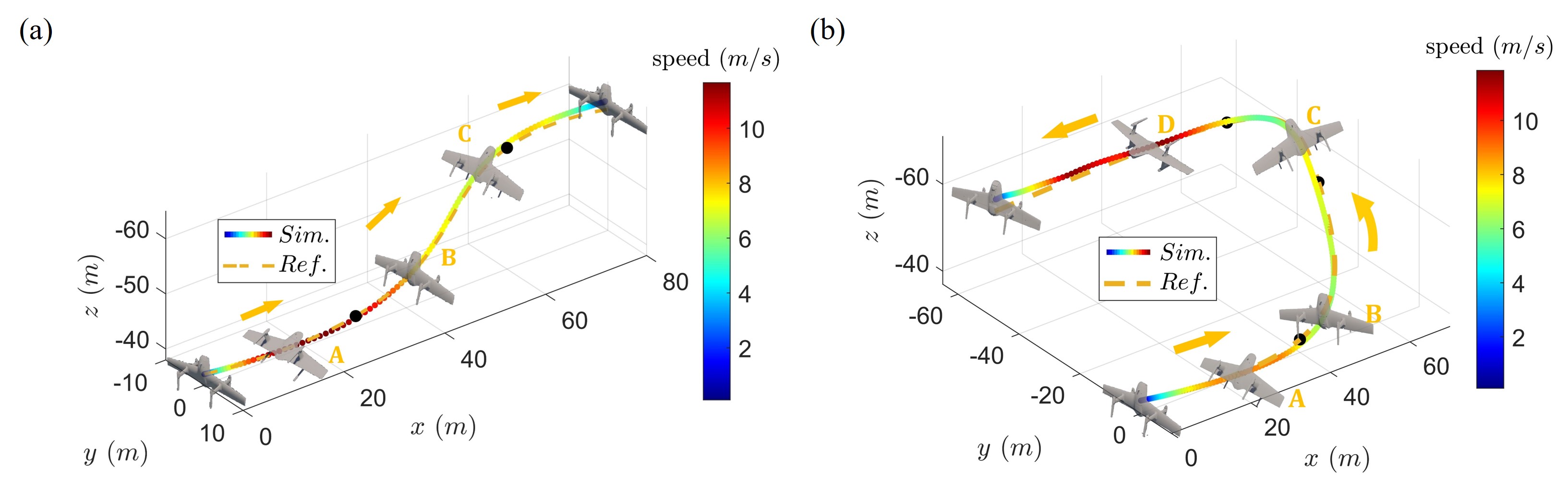}
    \caption{Two instances of trajectory optimization and tracking: (a) Given 2 intermediate waypoints, the tail-sitter ascends within a vertical plane. (b) Given 3 intermediate waypoints, the tail-sitter obtains centripetal force through roll, enabling it to perform three-dimensional flight.}
    \label{fig:result}
\end{figure*}
\begin{figure*}[]
    \centering
    \includegraphics[width=1\textwidth]{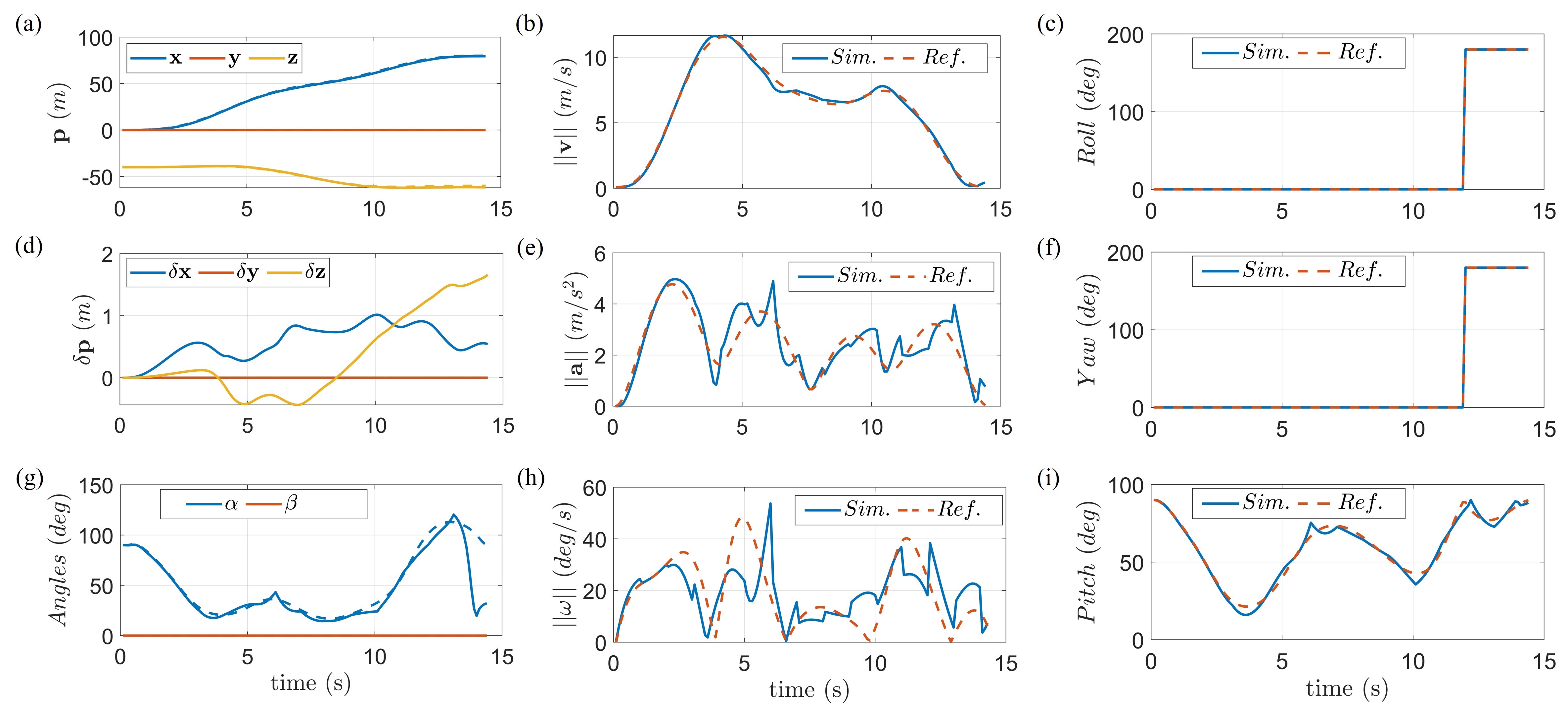}
    \caption{Simulation results of the trajectory as shown in Fig.\hyperref[fig:result]{9a}: (a) position, (b) speed, (c), (f) and (i) attitude Euler angles, (d) position tracking errors, (e) acceleration, (g) angle of attack and sideslip angle, and (h) angular velocity. The dashed lines represent the reference trajectories, while the solid lines denote the simulation results.}
    \label{fig:result1_details}
\end{figure*}

\begin{figure*}[]
    \centering
    \includegraphics[width=1\textwidth]{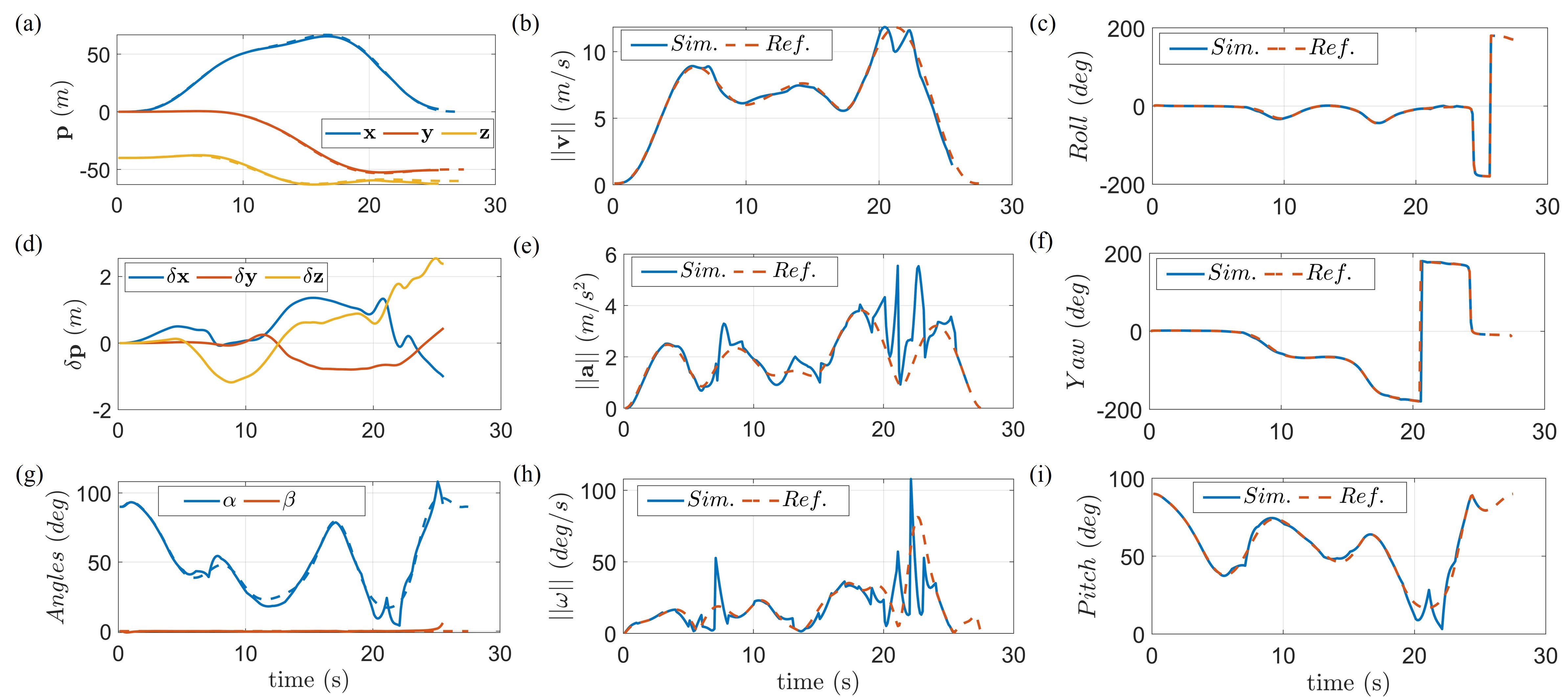}
    \caption{Simulation results of the trajectory as shown in Fig.\hyperref[fig:result]{9b}: (a) position, (b) speed, (c), (f) and (i) attitude Euler angles, (d) position tracking errors, (e) acceleration, (g) angle of attack and sideslip angle, and (h) angular velocity. The dashed lines represent the reference trajectories, while the solid lines denote the simulation results. Considering the vehicle's unstable states during its approach to a hover at the end of the trajectory, we define its arrival at the destination as occurring when its pitch angle closes to 90\degree and the vehicle is in proximity to the endpoint.}
    \label{fig:result2_details}
\end{figure*}

We apply the method from Section \hyperref[sec:multi_op]{\Rmnum{3}} to generate reference trajectories of the tail-sitter. Then, we leverage the dynamics (\ref{equ:tr_r_tail-sitter}) and aerodynamic data obtained from CFD (Section \hyperref[sec:flight_dynamics3]{\Rmnum{2}-C}) for flight simulation. As shown in Fig.\ref{fig:result}, we separate the entire trajectory by multiple pieces by intermediate waypoints (i.e., the black dots) that the trajectory must pass through. At the intermediate waypoints, the vehicle position $\mathbf{p}$ is user-defined, while other states such as speed $\mathbf{v}$, angular velocity $\boldsymbol{\omega}$, and attitude $\mathbf{R}$ do not need to be specified in advance. Both trajectories begin with a forward transition and end with a backward transition to hovering. With these boundary conditions, trajectories with start and final points are optimized by our trajectory optimization framework (\ref{equ:general_problem}). In terms of the MPC controller, horizon $N$ is set to $10$ and step size $\Delta t$ is set to $0.1$.

Fig.\hyperref[fig:result]{9a} shows the time-optimal trajectory of the tail-sitter ascending within a 2-dimensional vertical plane through given waypoints under certain speed constraints. The vehicle performs a trajectory with a width of 80 $m$, a height of 20 $m$, and a time duration of 14.5 $s$. The pitch angle increases from 16.748$\degree$ at position A in level flight to 75.379$\degree$ at position B, then the vehicle begins to lower the pitch angle to 50.428$\degree$ at position C for level flight. The resulting span of AoA is about 105.96$\degree$. The acceleration and angular velocity peak at 4.972 $m^2/s$ and 53.692 $deg./s$, respectively (Fig.\ref{fig:result1_details}).

Fig.\ref{fig:result2_details} presents the detailed flight data of the trajectory in Fig.\hyperref[fig:result]{9b}. The vehicle first transits from hovering to level flight with a speed of 8.84 $m/s$ at position A, then the vehicle begins to pull up the pitch angle for ascending and perform a coupled roll and yaw rotation for smooth turning at position B. The vehicle climbs 20 $m$ to the position C at 15.5 $s$ with yaw angle -82.09$\degree$, and the yaw angle increase to -180$\degree$ to achieve a complete turnaround of the tail-sitter. After that, the vehicle lower the pitch angle to 10.41$\degree$ to perform a 11.835 $m/s$ level flight at position D. Moreover, the acceleration and angular velocity, respectively peak, at 5.546 $m^2/s$ and 107.85 $deg./s$.

\begin{figure*}[p]
    \centering
   \subfloat[3.5s]{
        \includegraphics[width=0.33\textwidth]{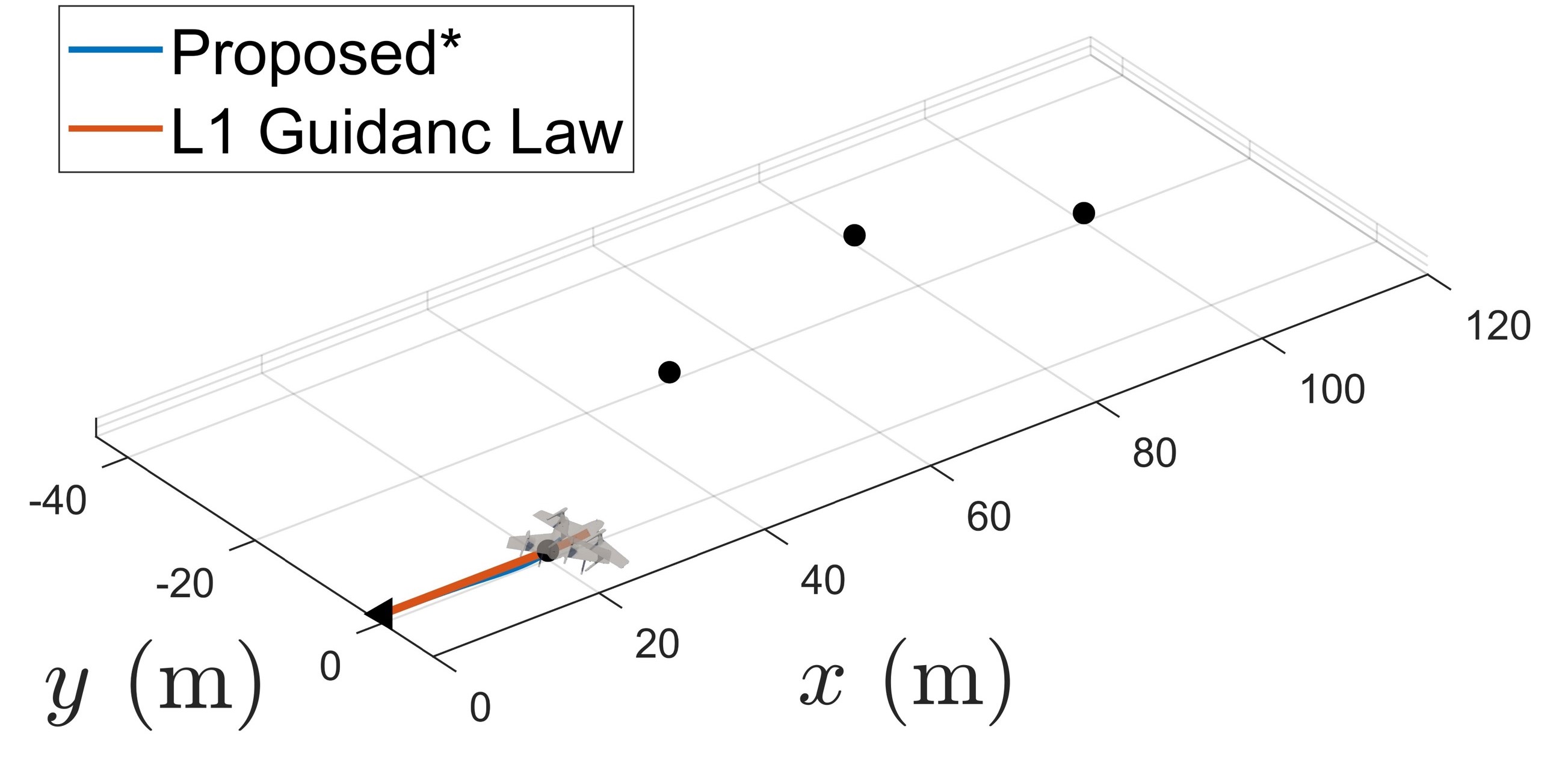}
    }
    \subfloat[7.2s]{
        \includegraphics[width=0.33\textwidth]{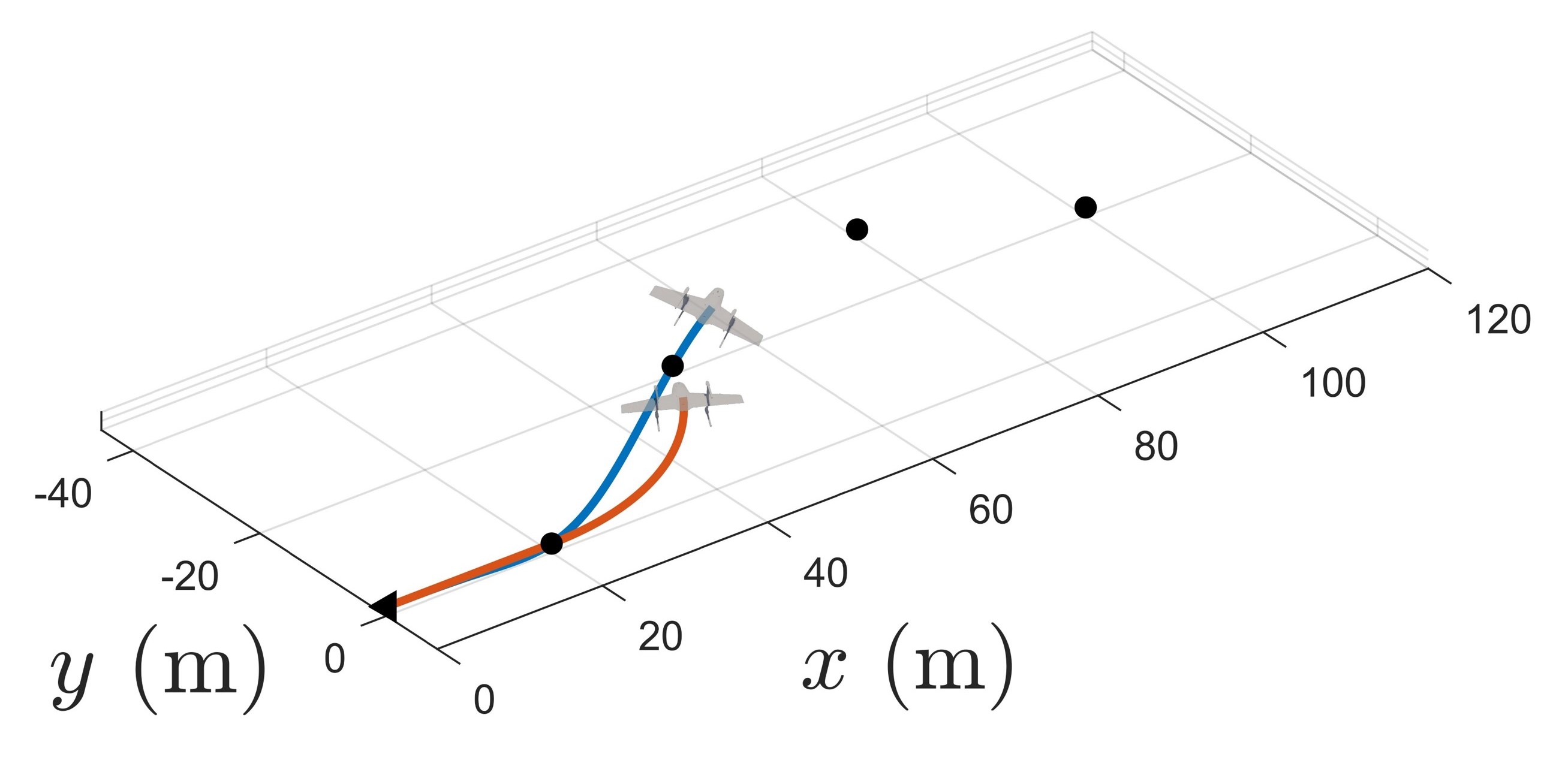}
    }
    \subfloat[10.4s]{
        \includegraphics[width=0.33\textwidth]{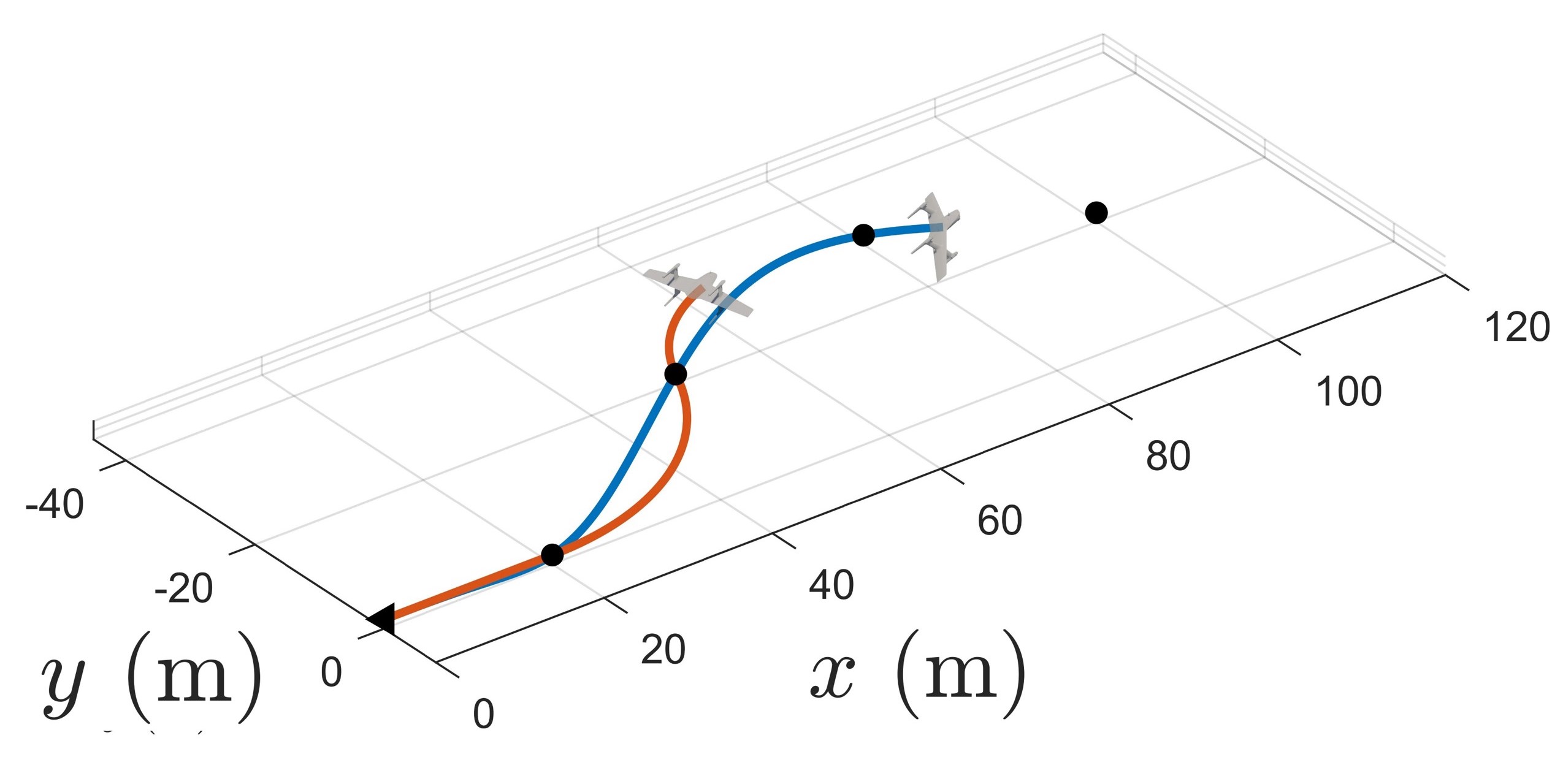}
    }

    \subfloat[13.7s]{
        \includegraphics[width=0.33\textwidth]{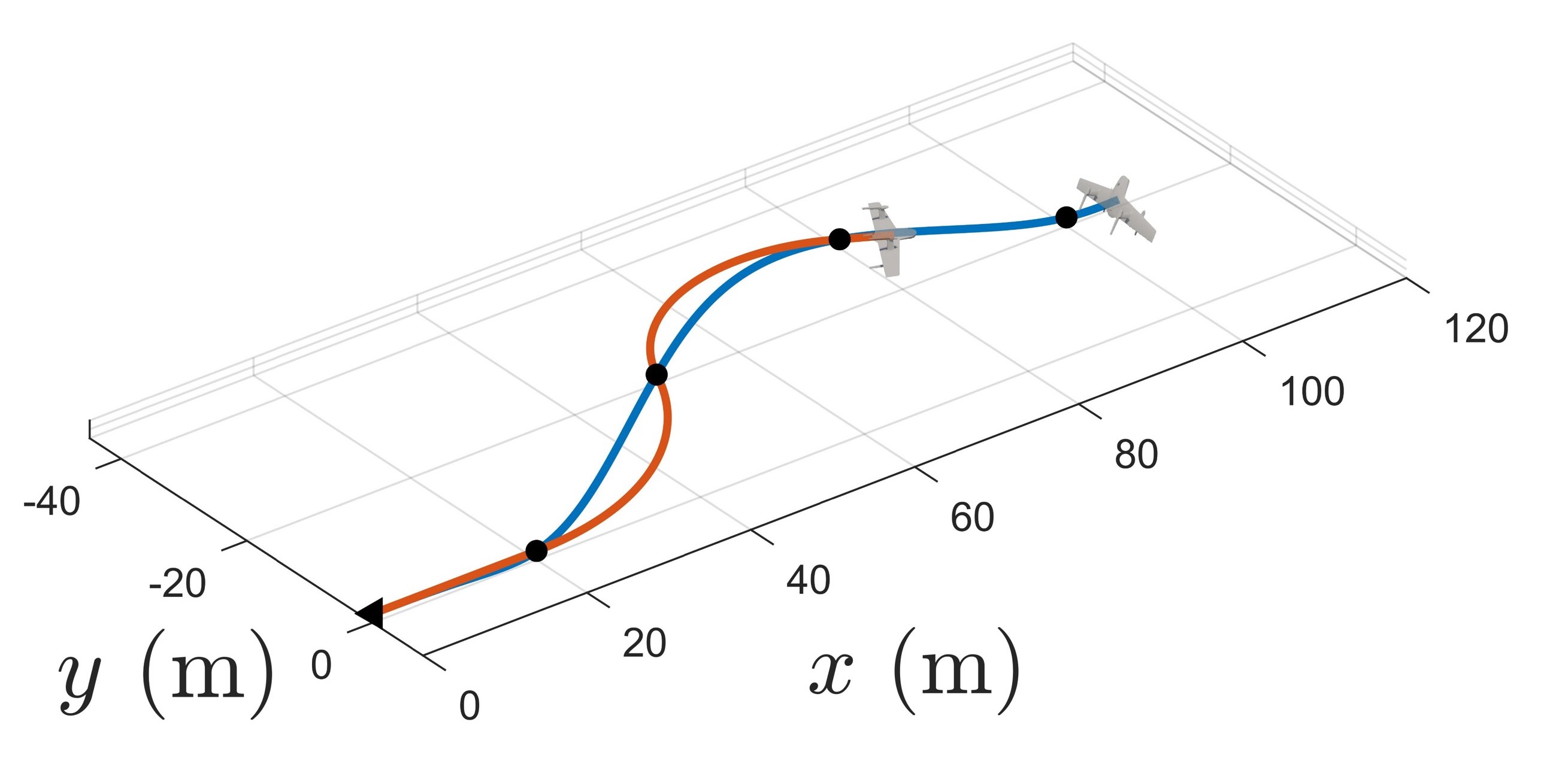}
    }
    \subfloat[15.75s]{
        \includegraphics[width=0.33\textwidth]{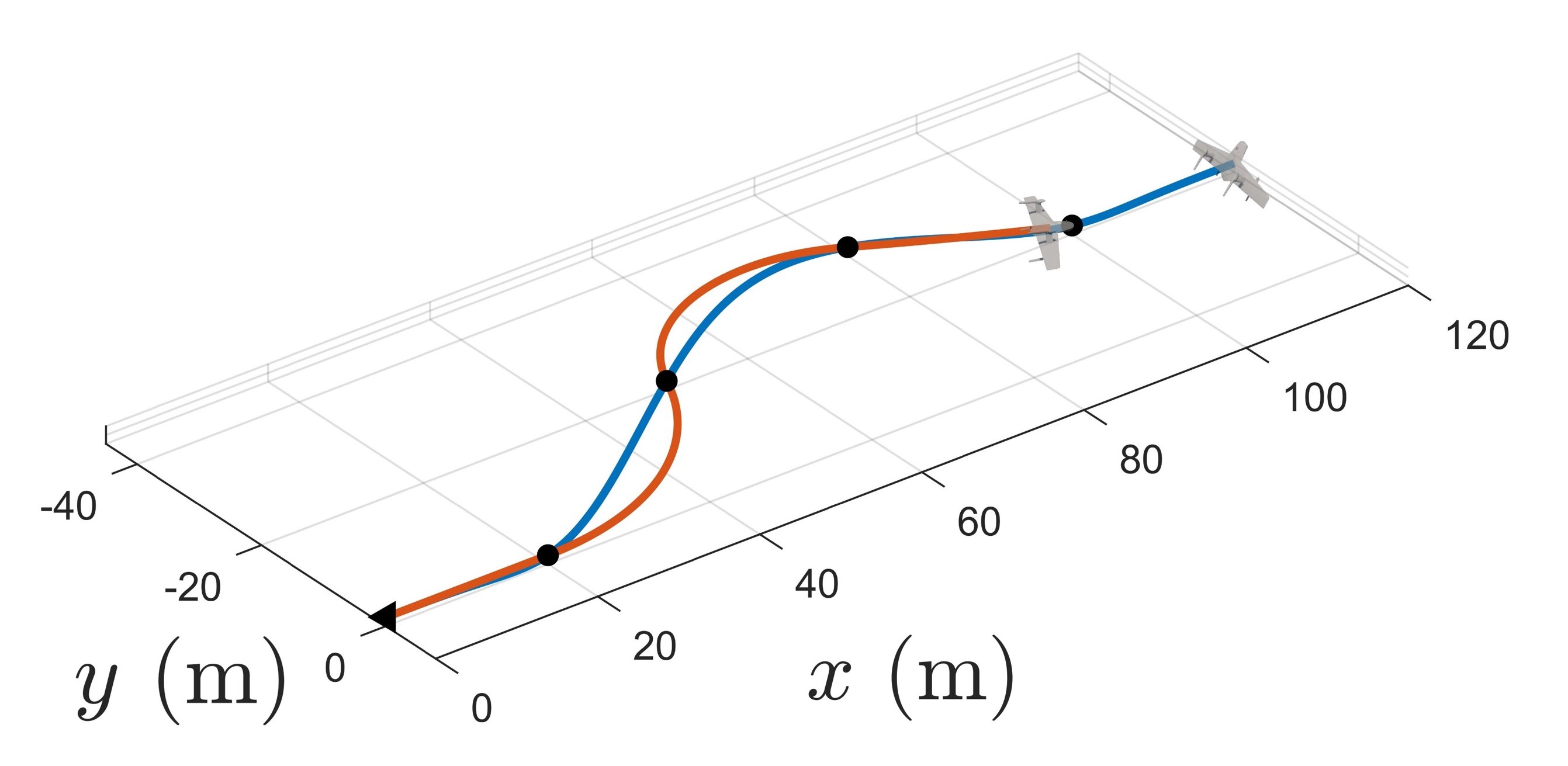}
    }
    \subfloat[18.9s]{
        \includegraphics[width=0.33\textwidth]{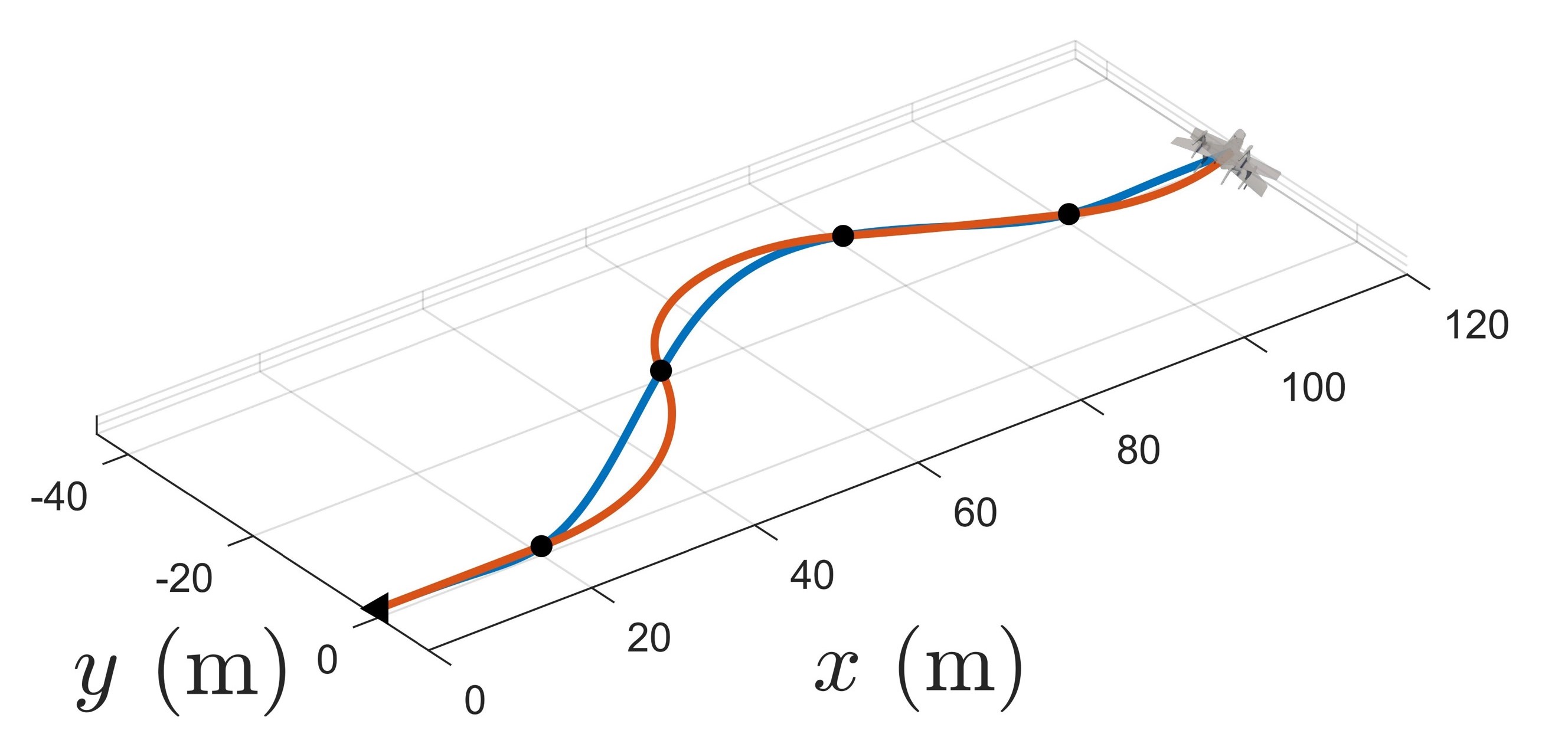}
    }
    \caption{The comparative trajectory of L1 Guidance Law and our method}
    \label{fig:L1}
\end{figure*}

\begin{figure*}[p]
    \centering
    \subfloat[1.1s]{
        \includegraphics[width=0.33\textwidth]{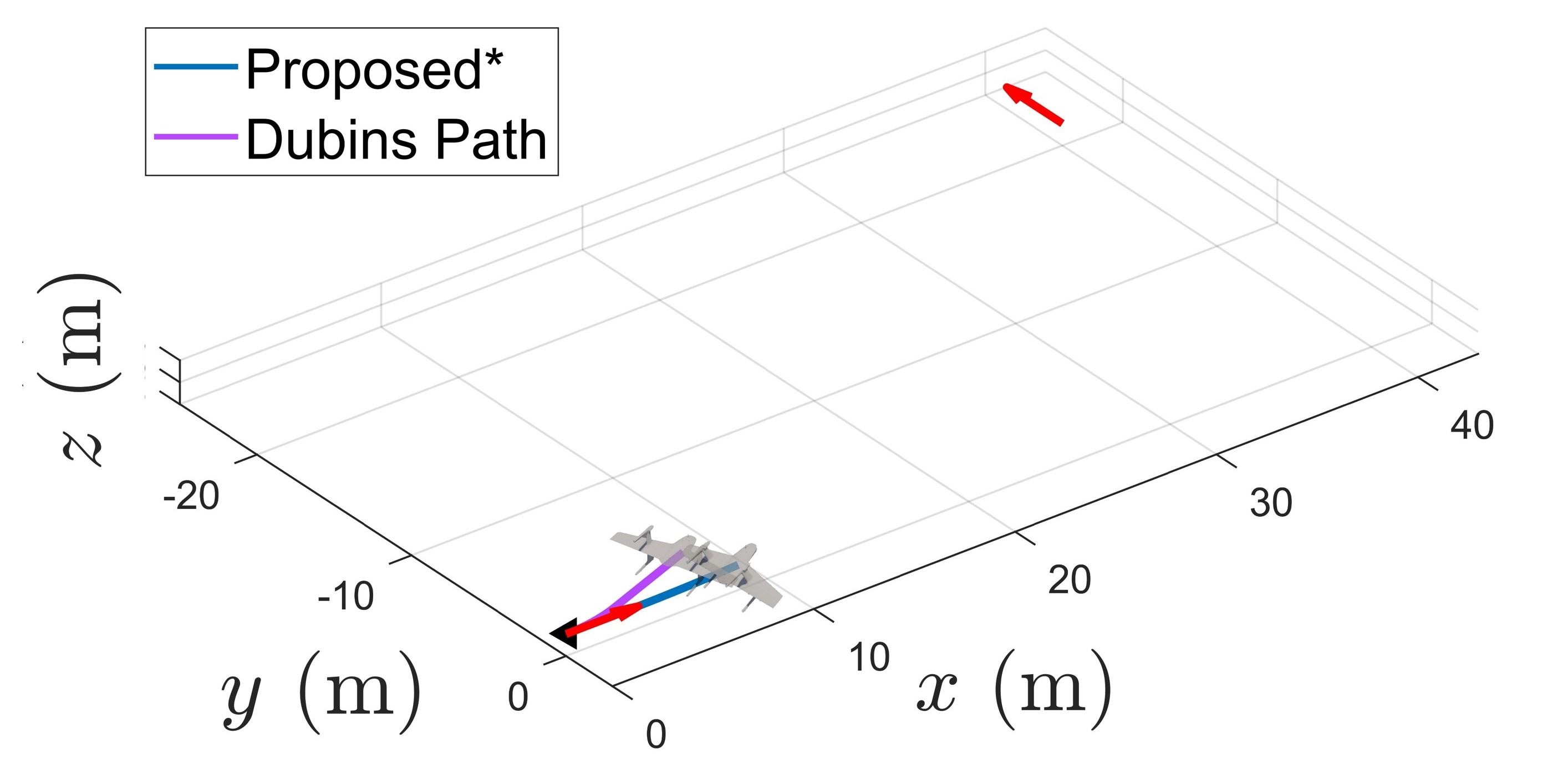}
    }
    \subfloat[1.92s]{
        \includegraphics[width=0.33\textwidth]{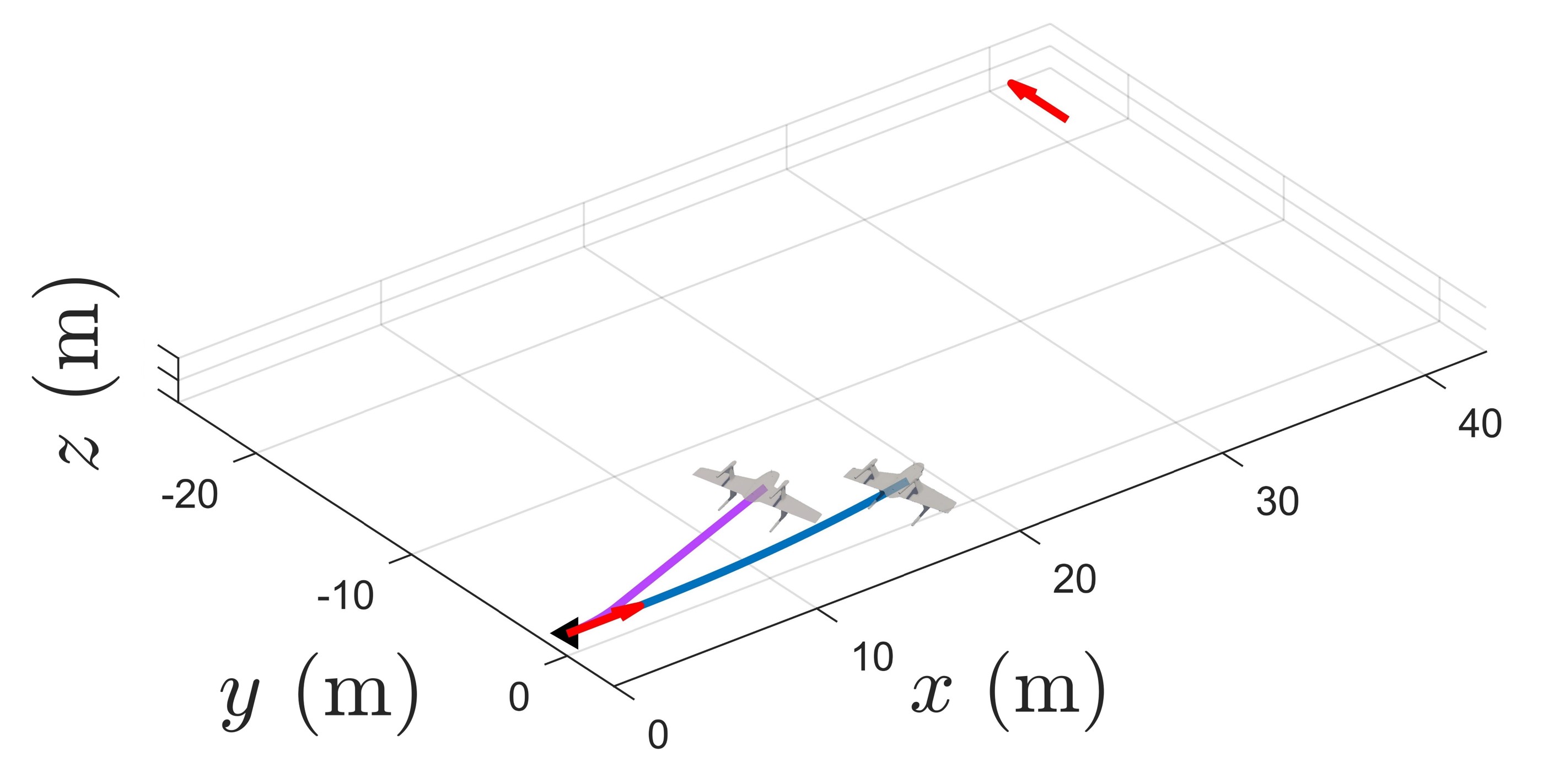}
    }
    \subfloat[2.95s]{
        \includegraphics[width=0.33\textwidth]{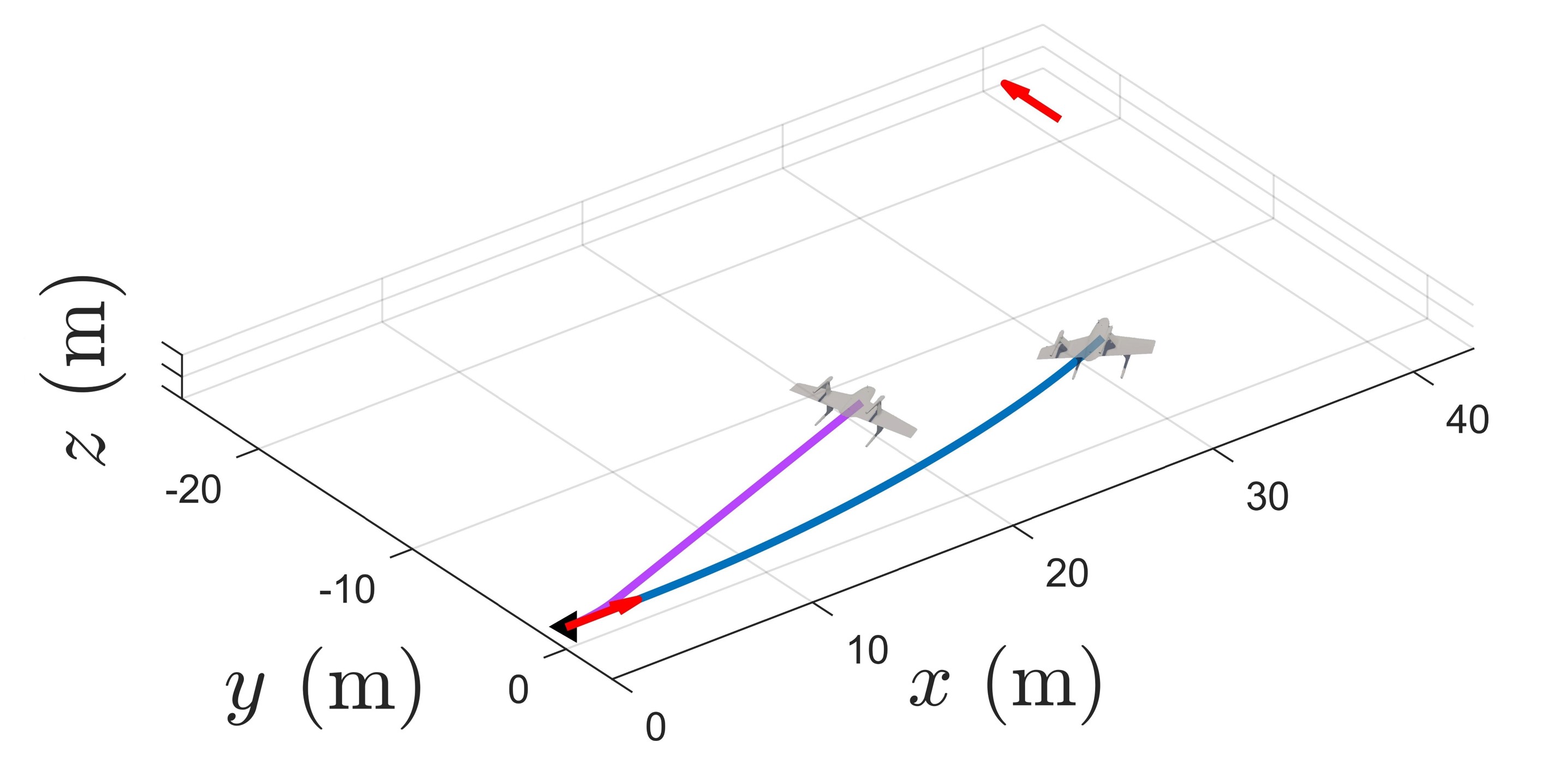}
    }

    \subfloat[4.12s]{
        \includegraphics[width=0.33\textwidth]{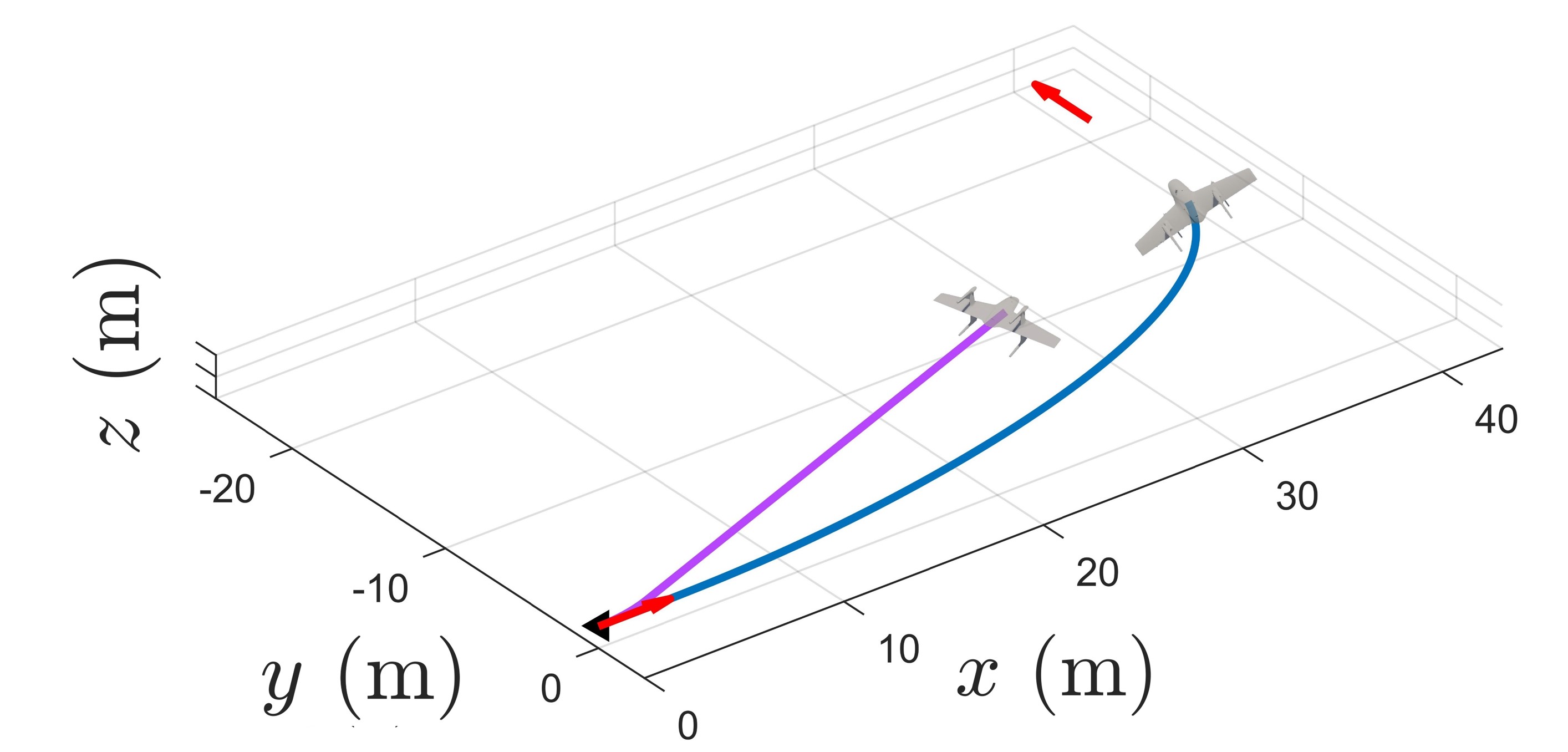}
    }
    \subfloat[5.1662s]{
        \includegraphics[width=0.33\textwidth]{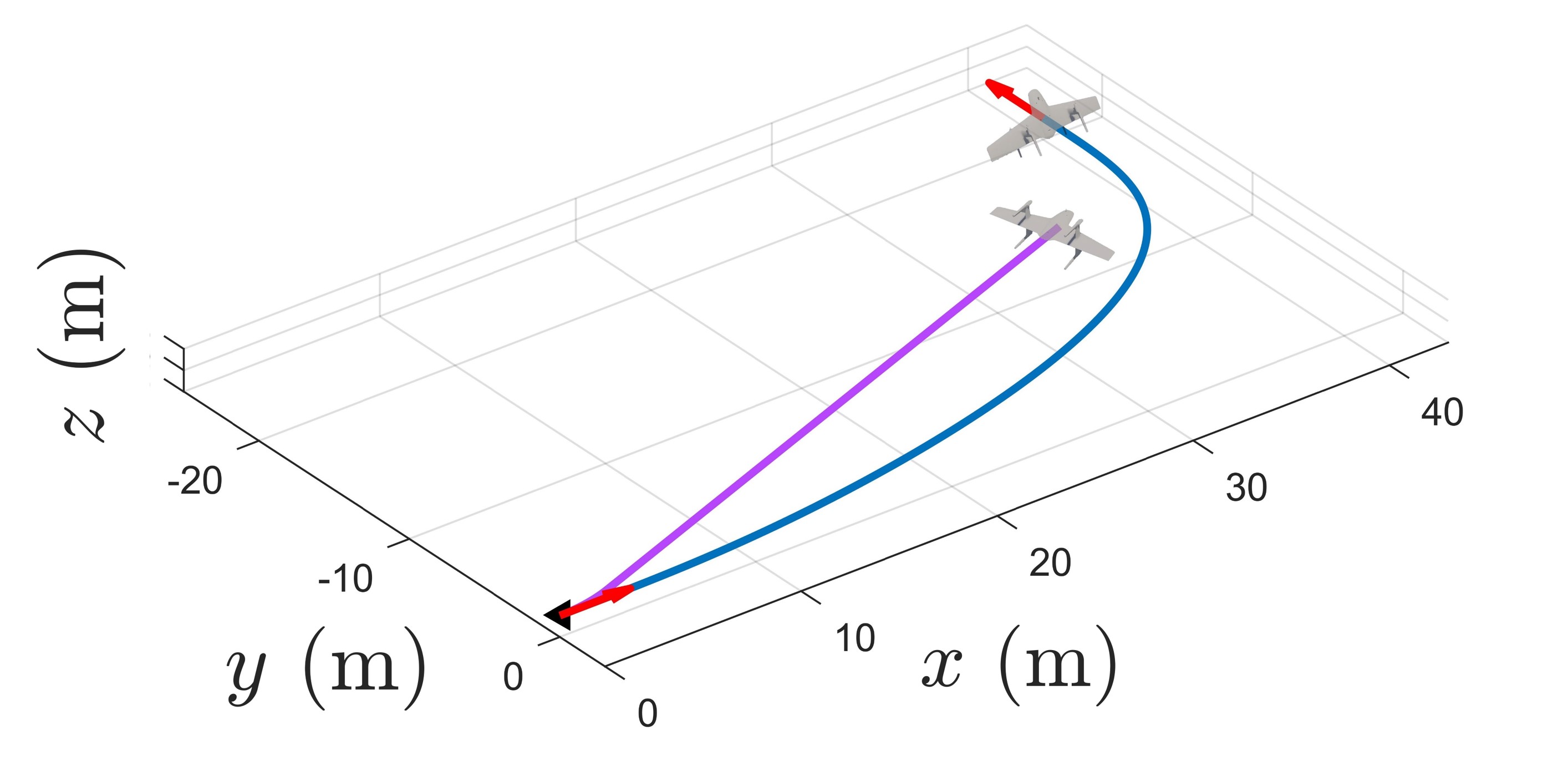}
    }
    \subfloat[6.452s]{
        \includegraphics[width=0.33\textwidth]{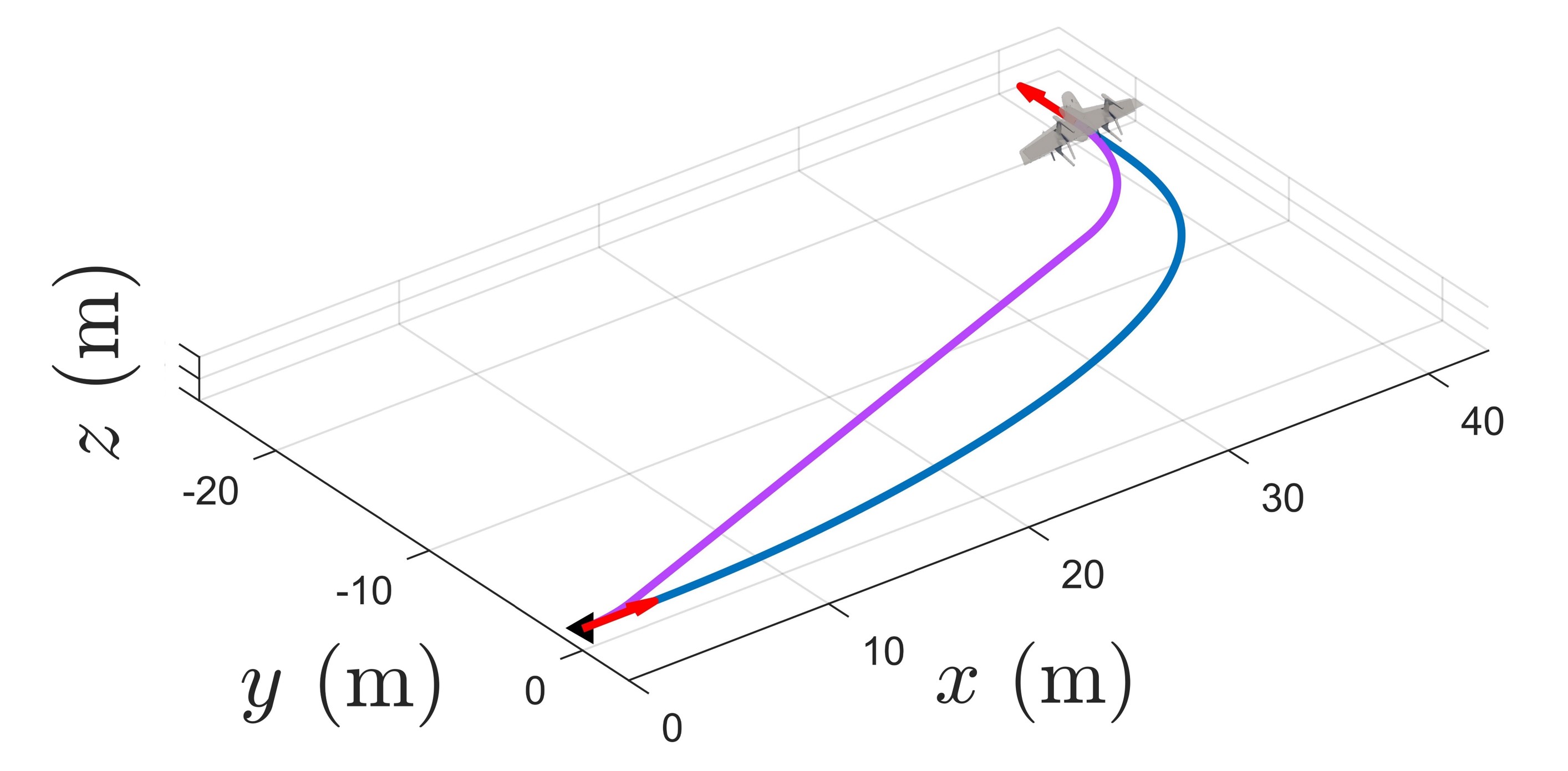}
    }
    \caption{The comparative trajecory of Dubins path and our method}
    \label{fig:dubins}
\end{figure*}
\begin{figure*}[p]
    \centering
    \includegraphics[width=1\linewidth]{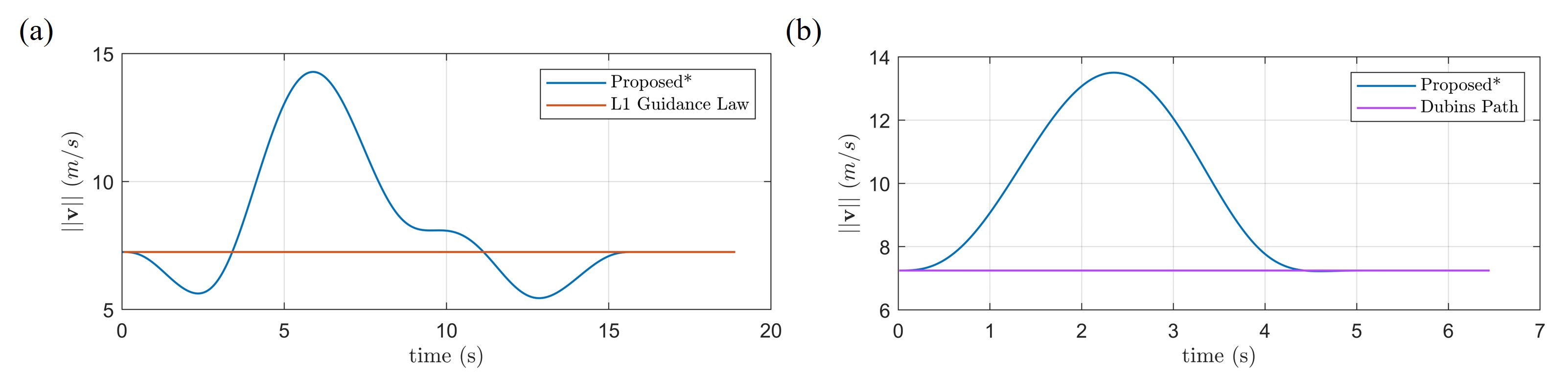}
    \caption{The speed of the tail-sitter in the 2 comparative experiments: (a) proposed* and L1 Guidance Law, (b) proposed* and Dubins path.}
    \label{fig:normv}
\end{figure*}

\begin{table*}[h]
    \centering
    \caption{Comparison of proposed method and L1 Guidance Law and Dubins path}
    \label{results_compare}
    \begin{tabular}{cccccc}
    \hline
        &Three-dimensional  &Multistage trajectory &Time consumed &Continuity of  &High-accuracy     \\ 
        &trajectory generation &generation & &control input &trajecoty tracking \\ \hline
    Proposed*     &\checkmark  &\checkmark  &Less &\checkmark &\checkmark\\
    L1 Guidance Law &$\times$ &\checkmark  &More  &$\times$  &$\times$ \\ 
    Dubins path     &$\times$  &$\times$  &More  &$\times$ &$\times$ \\ \hline
    \end{tabular}
\end{table*}
\begin{figure*}[h]
    \centering
    \subfloat[$\omega_{z_b}$ of L1 Guidance Law]{
        \includegraphics[width=0.5\linewidth]{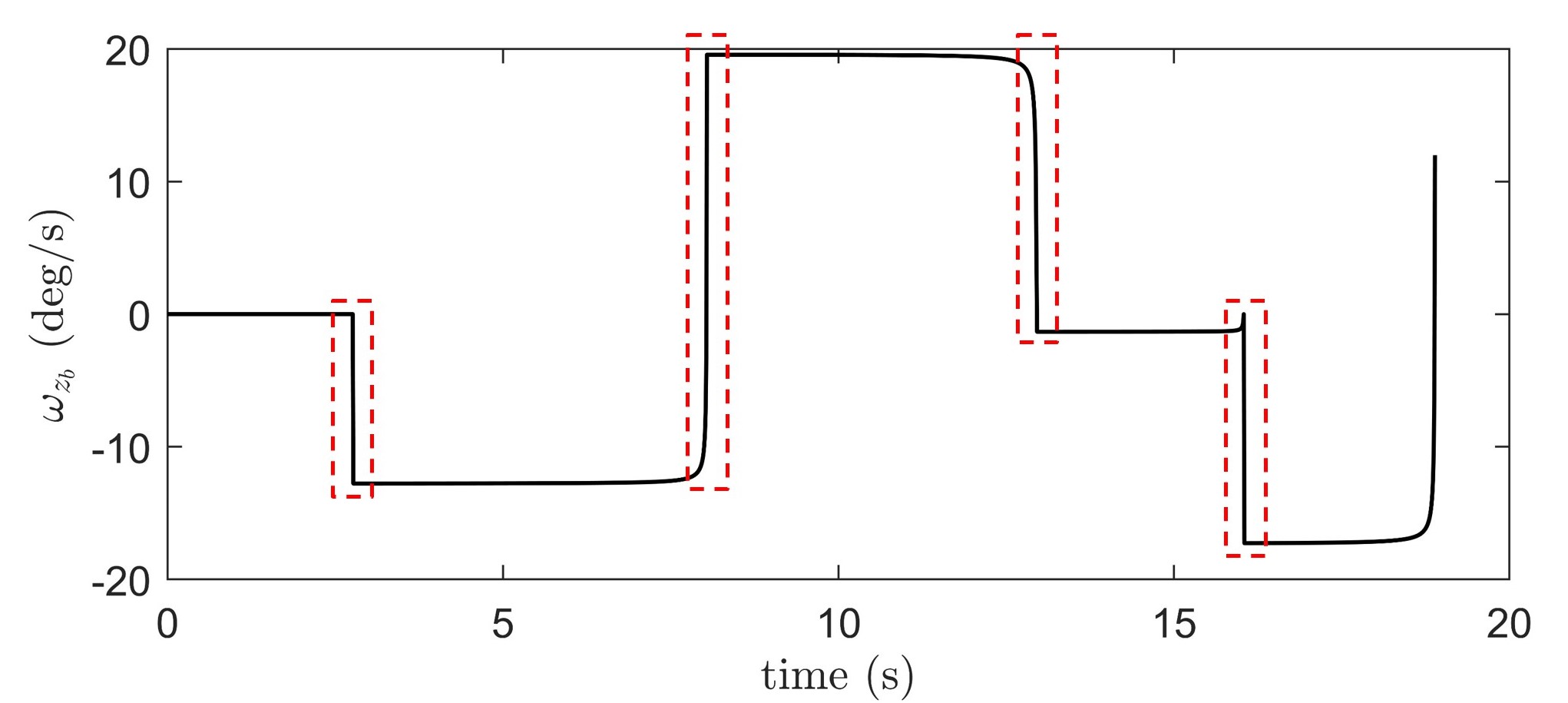}}
    \subfloat[$\omega_{z_b}$ of Dubins path]{
        \includegraphics[width=0.5\linewidth]{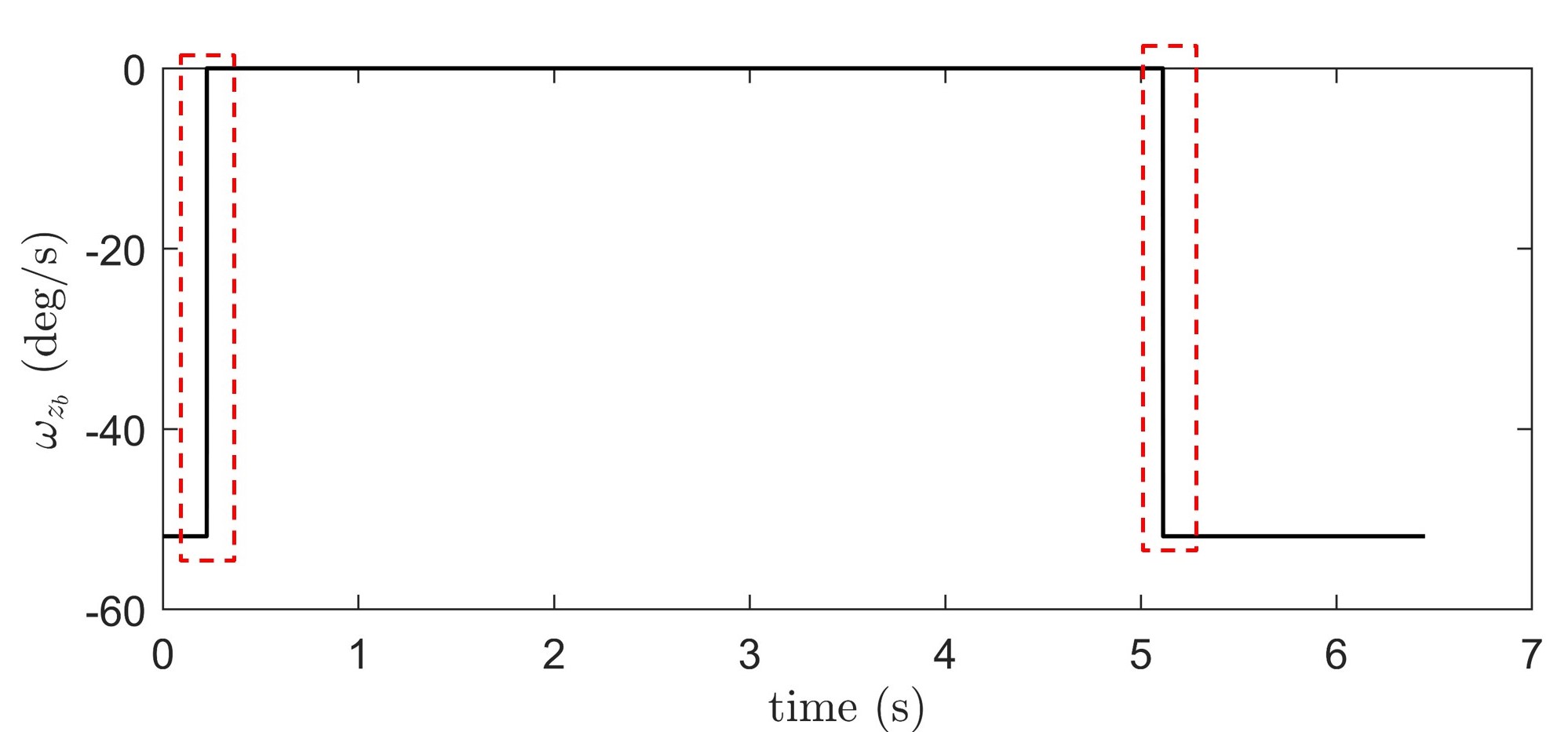}}
    \caption{$\omega_{z_b}$ of L1 Guidance Law and Dubins path: the angular velocity exhibits abrupt changes at specific time (i.e., red dashed boxes).}
    \label{fig:L1_dubins_omega}
\end{figure*}

\section{COMPARISON OF VARIOUS ALGORITHMS}
\label{sec:comparison}
In this section, the efficacy and applicability of proposed trajectory optimization method are verified through a series of comparative experiments in 2-dimensional horizontal plane.

We conducted comparative experiments of our method with well-developed fixed-wing UAV path planning methods, L1 Guidance Law and Dubins path. In the L1 Guidance Law, we predefined intermediate waypoints. Because of the limitations of Dubins path, we are unable to predefine the intermediate waypoints. Instead, we only define the two boundary states and set the turning radius to 8 $m$. For 2-dimensional level flight, the tail-sitter's speed is 7.245 $m/s$ in order to counteract the force of gravity, as stated in (\ref{eq:etkin}).

The advantages and drawbacks of our proposed trajectory generation method, in comparison to the L1 Guidance Law and Dubins path, are presented in Table \hyperref[results_compare]{\Rmnum{1}}. The method we proposed demonstrates its merits in the following aspects:
\begin{enumerate}[leftmargin=0.7cm,align=left,itemsep=1pt,topsep=-5pt,labelsep=0.5em,label=\arabic*)]
    \item \textit{Time consumed}: Fig.\ref{fig:L1} and Fig.\ref{fig:dubins} depict the attitudes and positions of the tail-sitter at various time instants. It is evident that our trajectory, when compared to the L1 Guidance Law and Dubins path, has a shorter duration for the tail-sitter to reach the final point. Due to the aerodynamic forces and moments resulting from the fuselage and wings during flight are strongly influenced by the UAV's speed (as indicated in (\ref{eq:etkin})). Both L1 Guidance Law and Dubins path restrict the UAV within a 2-dimensional plane, causing it to move only at a certain AoA and specific speed. In contrast, the differential flatness property enables the tail-sitter to decrease its speed during flight by raising the AoA during turns, thereby achieving high-speed flight along straight lines. This allows the tail-sitter to execute the trajectory within the entire flight envelope. The speed of the tail-sitter are shown in Fig.\ref{fig:normv}.
    \item \textit{Multistage trajectory generation}: In practical applications, it is typically necessary for the UAV to reach the specified locations in order to carry out a certain mission. Both our proposed method and the L1 Guidance Law enable the tail-sitter to navigate through intermediate waypoints without the need to predefine the tail-sitter's attitudes, based on the given positions of the waypoints. However, in the case of the Dubins Path, it is necessary to specify the initial and final yaw angles of the tail-sitter in order to select the shortest path from the six available types: RSR (Right-Straight-Right), RSL (Right-Straight-Left), LSR (Left-Straight-Right), LSL (Left-Straight-Left), RLR (Right-Left-Right), and LRL (Left-Right-Left). Meanwhile, the Dubins Path can only find the shortest path between two given points. Therefore, if intermediate waypoints are predefined, it is necessary to provide the UAV's yaw at these intermediate waypoints. However, it should be noted that the total path obtained may not necessarily be the shortest, as it is only the sum of the individual sub-paths. In order to determine the shortest path that includes intermediate waypoints, more sophisticated algorithms like A* or Dijkstra are usually necessary for Dubins. Therefore, this leads to additional computational burden. In addition, the L1 Guidance Law must verify in each iteration whether the L1 distance is smaller than a specified value $\delta$. If the value of $\delta$ is too tiny in practical applications, it can lead to the generated path deviating from the waypoint in certain instances, which is unacceptable.
    \item \textit{Continuity of control input}: Methods like L1 Guidance Law and Dubins Path explore the potential within the kinematics, without explicitly considering the aircraft dynamics. Steps in derivatives exist at the connecting points of geometrical components or at boundary points \cite{hong2021hierarchical}, as illustrated in Fig.\ref{fig:L1_dubins_omega}. This has a negative effect on the ability to accurately follow a desired path. In contrast, we use polynomials to express the positional information of the trajectory, which ensures the continuity and feasibility of reference control inputs, hence avoiding abrupt changes on the states.
    \item \textit{High-accuracy trajectory tracking}: While MPC is model-based controllers, our method enables the controller to exploit a high-fidelity aerodynamic model of the vehicle and its derivative(see (\ref{equ:tr_r_tail-sitter})), rather than relying on a unicycle model, in order to achieve high-accuracy trajectory tracking. Moreover, L1 Guidance Law and Dubins path both have discontinuity states that make tracking difficult.
    
\end{enumerate}
\vspace{5pt}

To sum up, our method is capable of generating consistent and reliable outcomes in both theoretical analysis and practical implementations. This significantly improves the safety of the tail-sitter during flight while also reducing the duration of the trip.

\section{CONCLUSION}
\label{sec:conclusion}
In this paper, in order to address the problem that tail-sitters typically can only cruise in 2-dimensional space, we proposed a 3-dimensional trajectory optimization method which is powered by several core features such as differential flatness and the MINCO trajectory, and conducted validations and analyses.

To demonstrate the top-level solution quality of our trajectory optimization method, we compared our algorithm with widely recognized fixed-wing UAV methods, Dubins path and L1 guidance law. Our method utilizes high-fidelity dynamics, which not only reduces the duration by an average of 18.298 $\%$, but also prevents abrupt changes in trajectory states, leading to more reliable outcomes. In addition, we introduced a global controller for trajectory tracking, which enables the 3-dimensional trajectories of the tail-sitter to be tracked. In a space of 80 $m\times$60 $m\times$30 $m$, the maximum error in 3-dimensional trajectory position does not exceed 2.688 $m$.

In conclusion, our method enables the generation of minimum control effort trajectories that traverse through any given waypoints in 3-dimensional space. We have implemented a temporal regularization term to balance the control effort and the overall trajectory duration. This ensures that the motion of the tail-sitter is adequately smooth and suitable for meeting the time constraints of the task. The tail-sitter performs agile flights within the entire envelope. The proposed method is universally suitable for all tail-sitter UAVs that utilize quadrotor to control their attitude.

\phantomsection
\section*{APPENDIX A: The differential flatness transformation of the tail-sitter during coordinated flight}
\label{appendixa}
\renewcommand{\theequation}{A.\arabic{equation}}
\renewcommand{\thefigure}{A.\arabic{figure}}
\setcounter{equation}{0}
\setcounter{figure}{0}

\begin{figure}[h]
    \centering
    \includegraphics[width=1\linewidth]{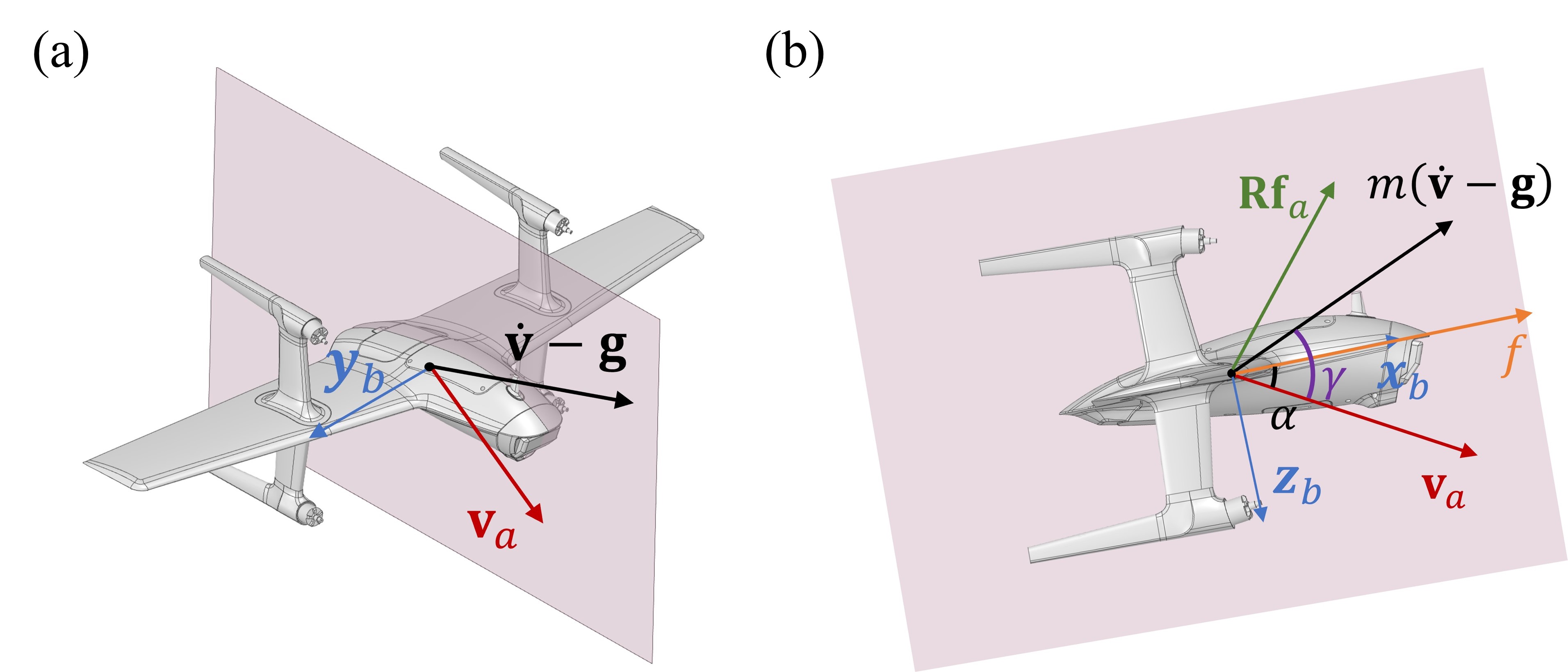}
    \caption{The tail-sitter in coordinated flight (the color pink represents the $\mathbf{x}_b\mathbf{O}_b\mathbf{z}_b$ plane): (a) $\mathbf{y}_b$ is perpendicular to both $\mathbf{v}_a$ and $\dot{\mathbf{v}}-\mathbf{g}$, (b) states of the tail-sitter and forces acting on the tail-sitter in coordinated flight.}
    \label{fig:coflight}
\end{figure}

The position $\mathbf{p}$ and velocity $\mathbf{v}$ of tail-sitter are flat output $\mathbf{p}$ itself and its first-order derivatives, respectively. As shown in Fig.\ref{fig:coflight}, in the coordinated flight, (1) $\mathbf{y}_b$ is perpendicular to the airspeed $\mathbf{v}_a$, (2) given that $\beta=0$, the direction of thrust is parallel to the axis $\mathbf{O}_b\mathbf{x}_b$, and the airframe is symmetric to the $\mathbf{x}_b\mathbf{O}_b\mathbf{z}_b$ plane, sideslip force $\mathcal{Y}$ is zero. That is being said, there is no force except the component of gravity along the axis $\mathbf{O}_b\mathbf{y}_b$ (i.e., $\mathbf{y}_b$ is perpendicular to $\dot{\mathbf{v}}-\mathbf{g}$). Therefore, the direction of $\mathbf{y}_b$ must coincide with or be opposite to the result of the cross product of $\dot{\mathbf{v}}-\mathbf{g}$ and $\mathbf{v}_a$, depending on the directions of $\dot{\mathbf{v}}-\mathbf{g}$ and $\mathbf{v}_a$. We denote the $\mathbf{y}_b$ at the previous time stamp as $\mathbf{y}_b'$:
\begin{align}
    & r=\operatorname{sign}\left(\left(\mathbf{v}_{a} \times(\dot{\mathbf{v}}-\mathbf{g})\right) \cdot \mathbf{y}_b'\right)
\end{align}
where $\operatorname{sign}(a)$ denotes the sign of $a\in\mathbb{R}$ and the $r$ denotes the direction of $\mathbf{y}_b$, ensuring that $\mathbf{y}_b\cdot\mathbf{y}_b'>0$ (i.e., the angle between $\mathbf{y}_b$ and $\mathbf{y}_b'$ is always less than $90\degree$, which ensures the direction of $\mathbf{y}_b$ solved at the current time stamp). To sum up, $\mathbf{y}_b$ can be determined as:
\begin{align}
    & \mathbf{y}_{b}=r \frac{\mathbf{v}_{a} \times(\dot{\mathbf{v}}-\mathbf{g})}{\left\|\mathbf{v}_{a} \times(\dot{\mathbf{v}}-\mathbf{g})\right\|}
\end{align}

Due to coordinated flight and the assumption of no crosswind, the sideslip force $\mathcal{Y}$ is zero. Hence, the aerodynamic forces $\mathbf{f}_a=\left[\mathbf{f}_{a_x} \quad 0 \quad \mathbf{f}_{a_z}\right]^{T}$ and $\mathbf{R} \mathbf{f}_{a}=\mathbf{x}_{b} \mathbf{f}_{a_x}+\mathbf{z}_b \mathbf{f}_{a_z}$. (\ref{co_R}) can be rewritten as:
\begin{align}
    \label{eq:app1}
    & \dot{\mathbf{v}}=\frac{f}{m} \mathbf{x}_{b}+\frac{\mathbf{f}_{a_{x}}}{m} \mathbf{x}_{b}+\frac{\mathbf{f}_{a_{z}}}{m} \mathbf{z}_{b}+\mathbf{g}
\end{align}

As shown in Fig.\ref{fig:coflight}, the forces acting on the tail-sitter are all within the $\mathbf{x}_b\mathbf{O}_b\mathbf{z}_b$ plane, therefore, all the forces can be decomposed into the axis $\mathbf{O}_b\mathbf{x}_b$ and axis $\mathbf{O}_b\mathbf{z}_b$, and (\ref{eq:app1}) can be written as:
\begin{subequations}
    \begin{align}
        & a_{T}=\mathbf{x}_{b}^{T}(\dot{\mathbf{v}}-\mathbf{g})-\mathbf{f}_{a_{x}} / m\\
        & \mathbf{z}_{b}^{T}(\dot{\mathbf{v}}-\mathbf{g})=\mathbf{f}_{a_{z}} / m
    \end{align}
\end{subequations}
where $a_T=f/m$ denotes thrust acceleration scalar. $\mathbf{x}_{b}^{T}(\dot{\mathbf{v}}-\mathbf{g})$ and $\mathbf{z}_{b}^{T}(\dot{\mathbf{v}}-\mathbf{g})$ are the components of $\dot{\mathbf{v}}-\mathbf{g}$ along the axis $\mathbf{O}_b\mathbf{x}_b$ and axis $\mathbf{O}_b\mathbf{z}_b$, respectively. Hence, we have:
\begin{subequations}
    \begin{align}
        &a_{T}=\|\dot{\mathbf{v}}-\mathbf{g}\| \cos (\gamma-\alpha)-\mathbf{f}_{a_{x}} / m \\
        \label{eq:app2}
        &-\|\dot{\mathbf{v}}-\mathbf{g}\| \sin (\gamma-\alpha)=\mathbf{f}_{a_{z}} / m
    \end{align}
\end{subequations}
where
\begin{align}
    & \gamma=r\cdot\operatorname{atan} 2\left(\left\|(\dot{\mathbf{v}}-\mathbf{g}) \times \mathbf{v}_{a}\right\|,(\dot{\mathbf{v}}-\mathbf{g}) \cdot \mathbf{v}_{a}\right)  
    \label{eq:app3}
\end{align}
and $\gamma$ denotes the angle that rotating $\mathbf{v}_a$ along $\mathbf{y}_b$ will reach $\dot{\mathbf{v}}-\mathbf{g}$, hence, as same as $\mathbf{y}_b$, $\gamma$ requires $r$ to determines its sign.

We observe that (\ref{eq:app2}) only involves the second-order derivatives of flat output $\mathbf{p}$ and the AoA $\alpha$. So (\ref{eq:app2}) can be rewritten as a nonlinear root-finding problem in terms of $\alpha$:
\begin{align}
    & F(\alpha)=h \sin (\gamma-\alpha)+\mathbf{C}_z(\alpha,0)=0
    \label{eq:app5}
\end{align}
where 
\begin{align}
    & h=\frac{2 m\|\dot{\mathbf{v}}-\mathbf{g}\|}{\rho V^{2} S}
    \label{eq:app4}
\end{align}

\begin{figure}[h]
    \centering
    \includegraphics[width=1\linewidth]{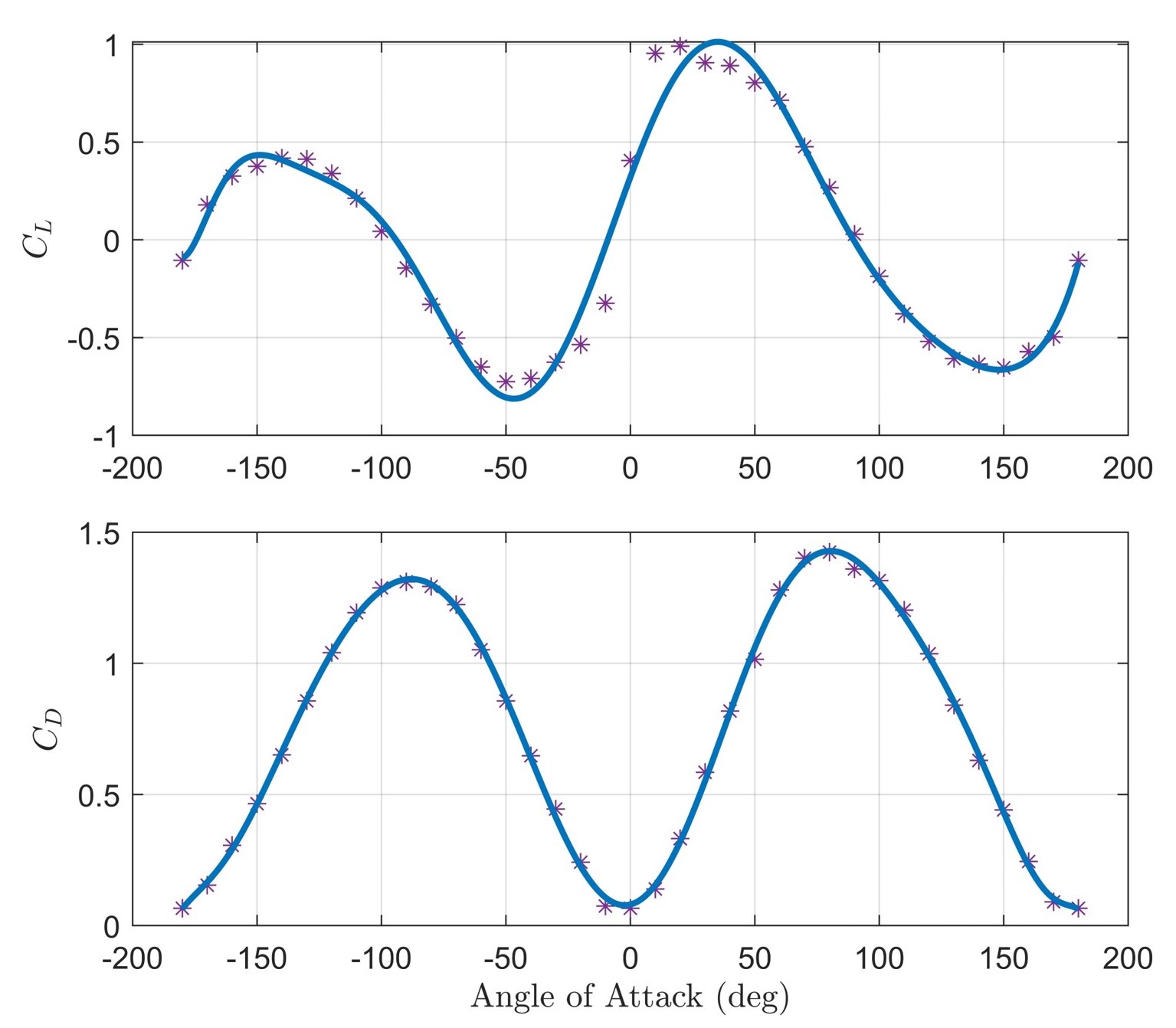}
    \caption{Longitudinal aerodynamic coefficients $C_L$, $C_D$ of SWAN K1 PRO tail-sitter prototype, identified by CFD simulation.}
    \label{fig:CLCD}
\end{figure}

According to (\ref{eq:app3}) and (\ref{eq:app4}), in the function $F(\alpha)$, $h$ and $\gamma$ are only related to second-order derivaties of flat output $\mathbf{p}$ and wind velocity $\mathbf{v}_a$. $\mathbf{C}_z$ is the component of the aerodynamic coefficient $\mathbf{C}$ on the axis $\mathbf{O}_b\mathbf{z}_b$ (as shown in (\ref{eq:app_fa})), which is solely related to the aerodynamic shape of the vehicle. As can be seen, the (\ref{eq:app5}) is highly nonlinear, hence no closed-form solution can be found in general. In practice, we fit the $C_L(\alpha,0)$ and $C_D(\alpha,0)$ using 2 carefully chosen polynomial functions, as shown in Fig.\ref{fig:CLCD}. So the (\ref{eq:app5}) can be solved numerically, such as Newton-Raphson method using $\mathbf{C}_z(\alpha,0)$ and $\partial \mathbf{C}_{z}(\alpha, 0) / \partial \alpha$.

With the solved AoA $\alpha$, $\mathbf{x}_b$ and $\mathbf{z}_b$ can be defined as:
\begin{subequations}
    \begin{align}
        &\mathbf{x}_{b}=\frac{\mathbf{v}_{a}}{\left\|\mathbf{v}_{a}\right\|} \cos(\alpha)+(\mathbf{y}_{b}\times\frac{\mathbf{v}_{a}}{\left\|\mathbf{v}_{a}\right\|})\sin(\alpha)\\
        &\mathbf{z}_b= \mathbf{x}_b \times \mathbf{y}_b
    \end{align}
\end{subequations}

Then, the rotation matrix $\mathbf{R}$ is:
\begin{align}
    & \mathbf{R}=\left[\mathbf{x}_{b} \quad \mathbf{y}_{b} \quad \mathbf{z}_{b}\right]
\end{align}

Thus, we have obtained the position $\mathbf{p}$, velocity $\mathbf{v}$, and rotation matrix $\mathbf{R}$ of the tail-sitter solely through flat output $\mathbf{p}$. According to the (\ref{co_R}) and (\ref{eq:faLDY}), we are able to compute one of the control input $f$ and its first-order derivative $a_T$ ($a_T=f/m$).

Next, we demonstrate how to represent state $\boldsymbol{\omega}$ and the control input $\boldsymbol{\tau}$ using flat output $\mathbf{p}$. Differentiate both sides of (\ref{co_R}):
\begin{align}
    \label{eq:app6}
    \begin{split}
        \ddot{\mathbf{v}}=&\left(\dot{a_T} \mathbf{R}+a_{T} \mathbf{R}\lfloor\boldsymbol{\omega}\rfloor\right) \mathbf{e}_{1} \\
        &+\frac{1}{m} \mathbf{R}\left(\lfloor\boldsymbol{\omega}\rfloor \mathbf{f}_{a}+\frac{\partial \mathbf{f}_{a}}{\partial\left(\mathbf{R}^{T} \mathbf{v}_{a}\right)} \frac{d}{d t}\left(\mathbf{R}^{T} \mathbf{v}_{a}\right)\right)\\
         =&\frac{1}{m} \mathbf{R} \frac{\partial \mathbf{f}_{a}}{\partial \mathbf{v}_{a}^{B}} \mathbf{R}^{T} \dot{\mathbf{v}_{a}}+\dot{a_T} \mathbf{R} \mathbf{e}_{1}\\
        &+\mathbf{R}\left(-\left\lfloor\left(a_T \mathbf{e}_{1}+\frac{\mathbf{f}_{a}}{m}\right)\right\rfloor+\frac{1}{m} \frac{\partial \mathbf{f}_{a}}{\partial \mathbf{v}_{a}^{B}}\left\lfloor\mathbf{v}_{a}^{B}\right\rfloor\right) \boldsymbol{\omega}
    \end{split}
\end{align}
where 
\begin{align}
    \label{eq:appa12}
    &\frac{\partial \mathbf{f}_{a}}{\partial \mathbf{v}_{a}^{B}}=\frac{\rho S}{2}\left(2 \mathbf{C} {\mathbf{v}_{a}^{B}}^{T}+\frac{\partial \mathbf{C}}{\partial \alpha} {\mathbf{v}_{a}^{B}}^{T}\left\lfloor\mathbf{e}_{2}\right\rfloor+V \frac{\partial \mathbf{C}}{\partial \beta} {\mathbf{e}_{2}}^{T}\right)
\end{align}

$\partial \mathbf{f}_{a}/\partial \mathbf{v}_{a}^{B}$ is calculated at $\beta=0$, the proof is given in Appendix \hyperref[appendixb]{B}. Moreover, we can decouple the lateral and longitudinal dynamics due to the model choices and the assumption of no sideslip, hence $\forall \alpha$:
\begin{subequations}
    \label{eq:partialc}
    \begin{align}
        C_{Y}(\alpha, 0)=&0\\
        \left.\frac{\partial C_{L}(\alpha, \beta)}{\partial \beta}\right|_{\beta=0}=&\left.\frac{\partial C_{D}(\alpha, \beta)}{\partial \beta}\right|_{\beta=0}=0 \\
        \left.\frac{\partial \mathbf{C}(\alpha, \beta)}{\partial \beta}\right|_{\beta=0}=&\left[
                                        0 \quad \left.\frac{\partial \mathbf{C}_{y}(\alpha, \beta)}{\partial \beta}\right|_{\beta=0} \quad 0
                                            \right]^{T} \\
        \left.\frac{\partial \mathbf{C}(\alpha, \beta)}{\partial \alpha}\right|_{\beta=0}=&\left[
\frac{\partial \mathbf{C}_{x}(\alpha, 0)}{\partial \alpha} \quad 0 \quad \frac{\partial \mathbf{C}_{z}(\alpha, 0)}{\partial \alpha}\right]^{T} 
    \end{align}
\end{subequations}

Recall that in coordinated flight the tail-sitter has no lateral airspeed:
\begin{align}
    \label{eq:app7}
    &\mathbf{v}_{a_{y}}^{B}={\mathbf{e}_{2}}^T \mathbf{R}^{T} \mathbf{v}_{a} \equiv 0
\end{align}

Differentiate both sides of (\ref{eq:app7}): 
\begin{align}
    \label{eq:app7_1}
    &-{\mathbf{v}_{a}}^T \mathbf{R}\left\lfloor\mathbf{e}_{2}\right\rfloor \boldsymbol{\omega}+{\mathbf{y}_{b}}^T \dot{\mathbf{v}_{a}}=0
\end{align}

Also, in (\ref{eq:app6}), $\ddot{\mathbf{v}}$, $\mathbf{R}$, $a_T$, $\dot{{\mathbf{v}_a}}$, $\mathbf{f}_a$ and $\partial \mathbf{f}_{a}/\partial \mathbf{v}_{a}^{B}$ are known. By combining (\ref{eq:app6}) and (\ref{eq:app7_1}), we obtain four linear equations in terms of $\dot{a_T}$ and $\boldsymbol{\omega}$, which can be written as:
\begin{align}
    \label{eq:app8}
    \begin{split}
        &\begin{bmatrix}
        \dot{{a}_T} \\
        \boldsymbol{\omega}
        \end{bmatrix}=\mathbf{N}^{-1} \mathbf{h}=\begin{bmatrix}
        \mathbf{N}_{1} \\
        \mathbf{N}_{2}
        \end{bmatrix}^{-1}\begin{bmatrix}
        \mathbf{h}_{1} \\
        \mathbf{h}_{2}
        \end{bmatrix}
    \end{split}
\end{align}
where 
\begin{subequations}
    \begin{align}
        \mathbf{h}_{1}=&{\mathbf{y}_{b}}^{T} \dot{\mathbf{v}_a}\\
        \mathbf{h}_{2}=&\ddot{\mathbf{v}}-\frac{1}{m} \mathbf{R} \frac{\partial \mathbf{f}_{a}}{\partial \mathbf{v}_{a}^{B}} \mathbf{R}^{T} \dot{\mathbf{v}_a}\\
        \mathbf{N}_{1}=&\left[0 \quad {\mathbf{v}_{a}}^{T} \mathbf{R}\left\lfloor\mathbf{e}_{2}\right\rfloor\right]\\
        \mathbf{N}_{2}=&\left[\mathbf{R}\mathbf{e}_1\quad\mathbf{R}\left(-\left\lfloor\left(a_{T} \mathbf{e}_{1}+\frac{\mathbf{f}_{a}}{m}\right)\right\rfloor+\frac{1}{m} \frac{\partial \mathbf{f}_{a}}{\partial \mathbf{v}_{a}^{B}}\left\lfloor\mathbf{v}_{a}^{B}\right\rfloor\right)\right]
    \end{align}
\end{subequations}

Solving the linear system (\ref{eq:app8}), we obtain the state of tail-sitter $\boldsymbol{\omega}$. To compute the control input $\boldsymbol{\tau}$, we need $\dot{\boldsymbol{\omega}}$ according to (\ref{eq:app_omega}). Hence, we differentiate both sides of (\ref{eq:app8}):
\begin{align}
    &\begin{bmatrix}
\ddot{a_T} \\
\dot{\boldsymbol{\omega}}
\end{bmatrix}=\frac{d}{d t}\left(\mathbf{N}^{-1} \mathbf{h}\right)=-\mathbf{N}^{-1} \dot{\mathbf{N}} \mathbf{N}^{-1} \mathbf{h}+\mathbf{N}^{-1} \dot{\mathbf{h}}
\end{align}
where the derivation of $\mathbf{N}$ and $\mathbf{h}$ is given in Appendix \hyperref[appendixc]{C}. 

With available $\dot{\boldsymbol{\omega}}$ and $\mathbf{M}_a$ ((\ref{eq:Ma})), control input $\boldsymbol{\tau}$ can be solved from (\ref{eq:app_omega}).

In conclusion, through the flat output $\mathbf{p}$, we have obtained all the state and control element of the tail-sitter.

\phantomsection
\section*{APPENDIX B: Proof of $\partial \mathbf{f}_{a}/\partial \mathbf{v}_{a}^{B}$}
\label{appendixb}
\renewcommand{\theequation}{B.\arabic{equation}}
\renewcommand{\thefigure}{B.\arabic{figure}}
\setcounter{equation}{0}
\setcounter{figure}{0}

Because of coordinated flight, there is no lateral airspeed (i.e., ${\mathbf{e}_{2}}^{T} \mathbf{v}_{a}^{B}=0$). Take the partial derivative of (\ref{eq:fa}) with respect to  $\mathbf{v}_{a}^{B}$:
\begin{align}
    \label{eq:appb_1}
    &\frac{\partial \mathbf{f}_{a}}{\partial \mathbf{v}_{a}^{B}}=\frac{\rho S}{2}\left(\mathbf{C} \frac{\partial V^{2}}{\partial \mathbf{v}_{a}^{B}}+V^{2} \frac{\partial \mathbf{C}}{\partial \alpha} \frac{\partial \alpha}{\partial \mathbf{v}_{a}^{B}}+V^{2} \frac{\partial \mathbf{C}}{\partial \beta} \frac{\partial \beta}{\partial \mathbf{v}_{a}^{B}}\right)
\end{align}
where
\begin{subequations}
    \label{eq:appb_2}
        \begin{align}
        \frac{\partial V^{2}}{\partial \mathbf{v}_{a}^{B}}=&\frac{\partial\left\|\mathbf{v}_{a}^{B}\right\|^{2}}{\partial \mathbf{v}_{a}^{B}}=2 {\mathbf{v}_{a}^{B}}^T\\
        \begin{split}
            \frac{\partial \alpha}{\partial \mathbf{v}_{a}^{B}}=&\frac{\partial}{\partial \mathbf{v}_{a}^{B}}\arctan\frac{{\mathbf{e}_{3}}^{T} \mathbf{v}_{a}^{B}}{{\mathbf{e}_{1}}^{T} \mathbf{v}_{a}^{B}}\\
             =& \frac{1}{1+\left(\frac{{\mathbf{e}_{3}}^{T} \mathbf{v}_{a}^{B}}{{\mathbf{e}_{1}}^{T} \mathbf{v}_{a}^{B}}\right)^{2}} \frac{{\mathbf{e}_{1}}^{T} \mathbf{v}_{a}^{B} {\mathbf{e}_{3}}^{T}-{\mathbf{e}_{3}}^{T} \mathbf{v}_{a}^{B} {\mathbf{e}_{1}}^{T}}{\left({\mathbf{e}_{1}}^{T} \mathbf{v}_{a}^{B}\right)^{2}} \\
            =&\frac{\left[-\mathbf{v}_{a_{z}}^{B}\quad0\quad \mathbf{v}_{a_{x}}^{B}\right]}{{\mathbf{v}_{a_{x}}^{B}}^{2}+{\mathbf{v}_{a_{z}}^{B}}^{2}} \\
            =&\frac{{\mathbf{v}_{a}^{B}}^{T}\left\lfloor\mathbf{e}_{2}\right\rfloor}{V^{2}}
        \end{split}
        \end{align}
        \vspace{-0.5cm}
        \begin{align}
        \begin{split}
            \frac{\partial \beta}{\partial \mathbf{v}_{a}^{B}}=&\frac{\partial}{\partial \mathbf{v}_{a}^{B}}\arcsin\frac{{\mathbf{e}_{2}}^{T} \mathbf{v}_{a}^{B}}{\left\|\mathbf{v}_{a}^{B}\right\|}\\
            =&\frac{1}{\sqrt{1-\left(\frac{{\mathbf{e}_{2}}^{T} \mathbf{v}_{a}^{B}}{\left\|\mathbf{v}_{a}^{B}\right\|}\right)^{2}}} \frac{\left\|\mathbf{v}_{a}^{B}\right\| {\mathbf{e}_{2}}^{T}-{\mathbf{e}_{2}}^{T} \mathbf{v}_{a}^{B} \frac{{\mathbf{v}_{a}^{B}}^{T}}{\left\|\mathbf{v}_{a}^{B}\right\|}}{\left\|\mathbf{v}_{a}^{B}\right\|^{2}} \\
            =&\frac{{\mathbf{e}_{2}}^{T}}{V} \\
        \end{split}
        \end{align}
\end{subequations}

Then, substituting (\ref{eq:appb_2}) into (\ref{eq:appb_1}), we have:
\begin{align}
    &\frac{\partial \mathbf{f}_{a}}{\partial \mathbf{v}_{a}^{B}}=\frac{\rho S}{2}\left(2 \mathbf{C} {\mathbf{v}_{a}^{B}}^{T}+\frac{\partial \mathbf{C}}{\partial \alpha} {\mathbf{v}_{a}^{B}}^{T}\left\lfloor\mathbf{e}_{2}\right\rfloor+V \frac{\partial \mathbf{C}}{\partial \beta} {\mathbf{e}_{2}}^{T}\right)
\end{align}

\begin{figure}[h]
    \centering
    \includegraphics[width=1\linewidth]{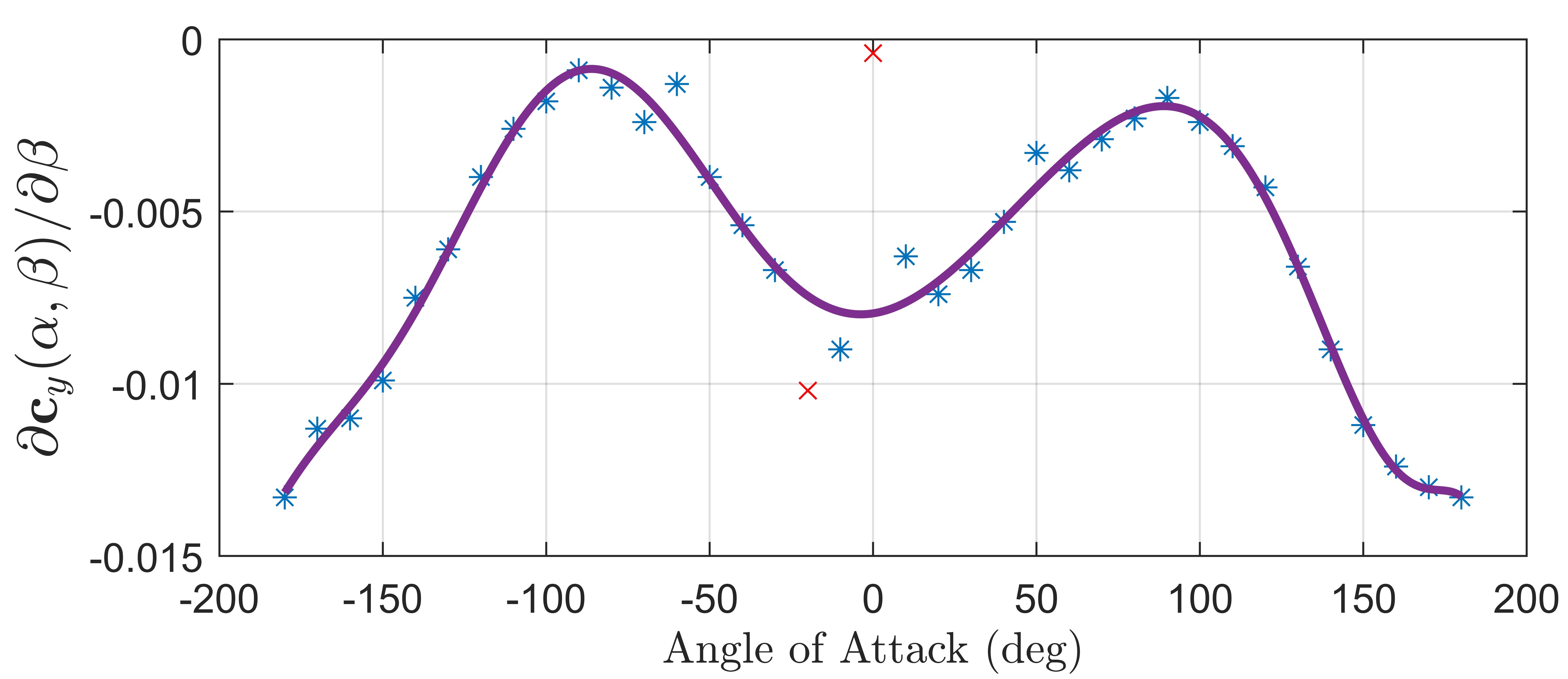}
    \caption{The aerodynamic coefficient gradient $\partial \mathbf{C}_y(\alpha,\beta)/\partial \beta$ when $\beta=0$. As same as $C_L$, $C_D$ and $C_m$, we fit the results with a polynomial function. Specifically, we ignore the abnormal results.}
    \label{fig:Cy_beta}
\end{figure}
The aerodynamic coefficient gradients $\partial \mathbf{C}/\partial \alpha$ and $\partial \mathbf{C}/\partial \beta$ can be obtained from (\ref{eq:partialc}). Specifically, $\partial \mathbf{C}_y(\alpha,\beta)/\partial \beta$ which is the second ond the only elment of $\partial \mathbf{C}/\partial \beta$ when $\beta=0$ can be obtained from the CFD simulation, as shown in Fig.\ref{fig:Cy_beta}.

\phantomsection
\section*{APPENDIX C: Derivation of $\mathbf{N}$ and $\mathbf{h}$}
\label{appendixc}
\renewcommand{\theequation}{C.\arabic{equation}}
\renewcommand{\thefigure}{C.\arabic{figure}}
\setcounter{equation}{0}
\setcounter{figure}{0}

According to (\ref{eq:app8}), we have:
    \begin{align}
        \dot{\mathbf{h}}=\begin{bmatrix}
        \dot{\mathbf{h}_1} \\
        \dot{\mathbf{h}_2}\end{bmatrix}, \dot{\mathbf{N}}=\begin{bmatrix}
        \dot{\mathbf{N}_1} \\
        \dot{\mathbf{N}_2}
        \end{bmatrix}
    \end{align}
where
\begin{subequations}
    \begin{align}
    \begin{split}
       \dot{\mathbf{h}_1}  =&\frac{d}{d t}\left({\mathbf{y}_{b}}^{T} \dot{\mathbf{v}_{a}}\right)=\frac{d}{d t}\left({\mathbf{e}_{2}}^{T} \mathbf{R}^{T} \dot{\mathbf{v}_a}\right) \\
=&{\mathbf{e}_{2}}^{T}\left(-\lfloor\boldsymbol{\omega}\rfloor \mathbf{R}^{T} \dot{\mathbf{v}_a}+\mathbf{R}^{T} \ddot{\mathbf{v}_a}\right)
    \end{split}\\
    \begin{split}
        \dot{\mathbf{h}_2}=& \frac{d}{d t}\left(\ddot{\mathbf{v}_a}-\frac{1}{m} \mathbf{R} \frac{\partial \mathbf{f}_{a}}{\partial \mathbf{v}_{a}^{B}} \mathbf{R}^{T} \dot{\mathbf{v}_a}\right) \\
=& \dddot{\mathbf{v}}-\frac{1}{m} \mathbf{R}\Biggl(\lfloor\boldsymbol{\omega}\rfloor \frac{\partial \mathbf{f}_{a}}{\partial \mathbf{v}_{a}^{B}} \mathbf{R}^{T} \dot{\mathbf{v}_a}+\frac{d}{d t}\left(\frac{\partial \mathbf{f}_{a}}{\partial \mathbf{v}_{a}^{B}}\right) \mathbf{R}^{T} \dot{\mathbf{v}_a} \\
&-\frac{\partial \mathbf{f}_{a}}{\partial \mathbf{v}_{a}^{B}}\lfloor\boldsymbol{\omega}\rfloor \mathbf{R}^{T} \dot{\mathbf{v}_a}+\frac{\partial \mathbf{f}_{a}}{\partial \mathbf{v}_{a}^{B}} \mathbf{R}^{T} \ddot{\mathbf{v}_a}\Biggr)
    \end{split}\\
    \dot{\mathbf{N}_1}=&\left[0 \quad \dot{\mathbf{N}_1}_2\right]\\
    \dot{\mathbf{N}_1}_2=&\frac{d}{d t}\left({\mathbf{v}_{a}}^T \mathbf{R}\left\lfloor\mathbf{e}_{2}\right\rfloor\right)=\left(\dot{\mathbf{v}_a}^{T} \mathbf{R}+{\mathbf{v}_{a}}^{T} \mathbf{R}\lfloor\boldsymbol{\omega}\rfloor\right)\left\lfloor\mathbf{e}_{2}\right\rfloor\\
    \dot{\mathbf{N}_2}=&\left[\dot{\mathbf{N}_2}_1 \quad \dot{\mathbf{N}_2}_2\right]\\
    \dot{\mathbf{N}_2}_1 =&\frac{d}{d t}\left(\mathbf{R} \mathbf{e}_{1}\right)=\mathbf{R}\lfloor\boldsymbol{\omega}\rfloor \mathbf{e}_{1}\\
        \begin{split}
        \dot{\mathbf{N}_2}_2= & \frac{d}{d t}\left(\mathbf{R}\left(-\left\lfloor\left(a_{T} \mathbf{e}_{1}+\frac{\mathbf{f}_{a}}{m}\right)\right\rfloor+\frac{1}{m} \frac{\partial \mathbf{f}_{a}}{\partial \mathbf{v}_{a}^{B}}\left\lfloor\mathbf{v}_{a}^{B}\right\rfloor\right)\right) \\
        = & \mathbf{R}\lfloor\boldsymbol{\omega}\rfloor\left(-\left\lfloor\left(a_{T} \mathbf{e}_{1}+\frac{\mathbf{f}_{a}}{m}\right)\right\rfloor+\frac{1}{m} \frac{\partial \mathbf{f}_{a}}{\partial \mathbf{v}_{a}^{B}}\left\lfloor\mathbf{v}_{a}^{B}\right\rfloor\right) \\
        & +\mathbf{R}\left(-\left\lfloor\left(\dot{a_T} \mathbf{e}_{1}+\frac{1}{m}\left(\frac{\partial \mathbf{f}_{a}}{\partial V} \dot{V}+\frac{\partial \mathbf{f}_{a}}{\partial \alpha} \dot{\alpha}\right)\right)\right\rfloor\right. \\
        & \left.+\frac{1}{m}\left(\left(\frac{d}{d t}\left(\frac{\partial \mathbf{f}_{a}}{\partial \mathbf{v}_{a}^{B}}\right)\right)\left\lfloor\mathbf{v}_{a}^{B}\right\rfloor+\frac{\partial \mathbf{f}_{a}}{\partial \mathbf{v}_{a}^{B}}\left\lfloor\dot{\mathbf{v}_{a}^{B}}\right\rfloor\right)\right)
        \end{split}
        \end{align}
        \end{subequations}
        
Moreover, we have:
        \begin{subequations}
        \begin{align}
        \frac{\partial \mathbf{f}_{a}}{\partial V}=&\rho V S \mathbf{C}\\
        \frac{\partial \mathbf{f}_{a}}{\partial \alpha}=&\frac{1}{2} \rho V^{2} S \frac{\partial \mathbf{C}}{\partial \alpha}\\
        \dot{\mathbf{v}_{a}^{B}}=&\frac{d}{d t}\left(\mathbf{R}^{T} \mathbf{v}_{a}\right)=-\lfloor\boldsymbol{\omega}\rfloor \mathbf{R}^{T} \mathbf{v}_{a}+\mathbf{R}^{T} \dot{\mathbf{v}_{a}}\\
        \begin{split}
            \dot{\alpha} =&\frac{1}{1+\left(\frac{\mathbf{v}_{a_{z}}^{B}}{\mathbf{v}_{a_{x}}^{B}}\right)^{2}} \frac{\dot{\mathbf{v}_{a_{z}}^{B}} \mathbf{v}_{a_{x}}^{B}-\mathbf{v}_{a_{z}}^{B} \dot{\mathbf{v}_{a_{x}}^{B}}}{{\mathbf{v}_{a_{x}}^{B}}^2} \\
            =&\frac{\dot{\mathbf{v}_{a_{z}}^{B}} \mathbf{v}_{a_{x}}^{B}-\mathbf{v}_{a_{z}}^{B} \dot{\mathbf{v}_{a_{x}}^{B}}}{V^{2}}
        \end{split}\\
        \dot{\mathbf{v}_a}=&\dot{\mathbf{v}}-\dot{\mathbf{w}}\\
        \ddot{\mathbf{v}_a}=&\ddot{\mathbf{v}}-\ddot{\mathbf{w}}\\
        \dot{V}=&{\mathbf{v}_{a}}^{T} \dot{\mathbf{v}_a} / V
    \end{align}
\end{subequations}

Differentiate both sides of (\ref{eq:appa12}), we have:
\begin{align}
\begin{split}
\frac{d}{d t}\left(\frac{\partial \mathbf{f}_{a}}{\partial \mathbf{v}_{a}^{B}}\right)= & \frac{\rho S}{2}\Bigg(2\left(\frac{\partial \mathbf{C}}{\partial \alpha} \dot{\alpha} {\mathbf{v}_{a}^B}^{T}+\mathbf{C} \dot{\mathbf{v}_{a}^{B}}^{T}\right)\\
&+\left(\frac{\partial^{2} \mathbf{C}}{\partial \alpha^{2}} \dot{\alpha} {\mathbf{v}_{a}^{B}}^{T}+\frac{\partial \mathbf{C}}{\partial \alpha}{\dot{\mathbf{v}_{a}^{B}}}^{T}\right)\left\lfloor\mathbf{e}_{2}\right\rfloor\\
&+\left(\dot{V} \frac{\partial \mathbf{C}}{\partial \beta}+V \frac{\partial^{2} \mathbf{C}}{\partial \beta \partial \alpha} \dot{\alpha}\right) \mathbf{e}_{2}^{T} \Bigg)
\end{split}
\end{align}

\phantomsection
\section*{APPENDIX D: Proof of (\ref{eq:dotdeltaR})}
\label{appendixd}
\renewcommand{\theequation}{D.\arabic{equation}}
\renewcommand{\thefigure}{D.\arabic{figure}}
\setcounter{equation}{0}
\setcounter{figure}{0}

According to \cite{bullo1995proportional}, we have:
\begin{align}
    \dot{\boldsymbol{\theta}}=\mathbf{A}^{-T}(\boldsymbol{\theta})\boldsymbol{\omega}=\mathbf{A}^{-T}(\boldsymbol{\theta})\left(\mathbf{R}^{T} \dot{\mathbf{R}}\right)^{\vee}
\end{align}
where $(\cdot)^{\vee}$ means inverse of $\left \lfloor \cdot \right \rfloor $. At the same time, we note that $\boldsymbol{\theta} = Log(\mathbf{R})$ and $\delta\mathbf{R} =Log(\mathbf{R}^T\mathbf{R}_r)$, thus $\dot{\delta\mathbf{R}}$ is:
\begin{align}
\begin{split}
    \dot{\delta\mathbf{R}}=&\mathbf{A}^{-T}(\delta\mathbf{R})\left({\mathbf{R}_r}^T\mathbf{R}\frac{d(\mathbf{R}^T\mathbf{R}_r)}{dt}\right)^{\vee}\\
    =&\mathbf{A}^{-T}(\delta\mathbf{R})\left(-{\mathbf{R}_r}^T\mathbf{R}\left \lfloor \boldsymbol{\omega} \right \rfloor\mathbf{R}^T\mathbf{R}_r+\left \lfloor \boldsymbol{\omega}_r \right \rfloor \right)^{\vee}\\
    =& \mathbf{A}^{-T}(\delta\mathbf{R})\left(-{\mathbf{R}_r}^T \mathbf{R} \boldsymbol{\omega}+\boldsymbol{\omega}_{r}\right)
\end{split}
\end{align}

\phantomsection
\section*{APPENDIX E: Proof of (\ref{eq:FxFu})}
\label{appendixe}
\renewcommand{\theequation}{E.\arabic{equation}}
\renewcommand{\thefigure}{E.\arabic{figure}}
\setcounter{equation}{0}
\setcounter{figure}{0}

According to \cite{bullo1995proportional}, we have:
\begin{align}
    \mathbf{R}^T\mathbf{R}_r=Exp(\delta\mathbf{R})
\end{align}
where $Exp(\cdot)$ is exponential map and also the inverse of the logarithmic map $Log(\cdot)$, which is defined as:
\begin{align}
\label{eq:exp}
\begin{split}
    Exp_{SO(3)}(\delta\mathbf{R})=&\mathbf{I}+\frac{\sin \|\delta\mathbf{R}\|}{\|\delta\mathbf{R}\|}\left \lfloor \delta\mathbf{R} \right \rfloor \\
    &+\frac{1-\cos \|\delta\mathbf{R}\|}{\|\delta\mathbf{R}\|^{2}}\left \lfloor \delta\mathbf{R} \right \rfloor^2
\end{split}
\end{align}

Note that when $\delta\mathbf{R}\to0$, $\mathbf{R}^T\mathbf{R}_r\approx\mathbf{I}+\left \lfloor \delta\mathbf{R} \right \rfloor$, implies the following:
\begin{align}
    \label{eq:deltav}
    \begin{split}
        \dot{\delta\mathbf{v}} =& \frac{1}{m}\left(f_r\mathbf{R}_r\mathbf{e}_1+\mathbf{R}_r{\mathbf{f}_a}_r-(f\mathbf{R}\mathbf{e}_1+\mathbf{R}{\mathbf{f}_a})\right)\\
        \approx&\left({a_T}_r\mathbf{R}_r-({a_T}_r-\delta a_T)\mathbf{R}_r(\mathbf{I}+\left \lfloor \delta\mathbf{R} \right \rfloor )^T\right)\mathbf{e}_1\\
        &+\frac{1}{m}\mathbf{R}_r({\mathbf{f}_a}_r-(\mathbf{I}+\left \lfloor \delta\mathbf{R} \right \rfloor )^T((\mathbf{f}_a)_r-\delta\mathbf{f}_a))\\
        =&\delta a_T\mathbf{R}_r\mathbf{e}_1+a_T\mathbf{R}_r\left \lfloor \delta\mathbf{R} \right \rfloor\mathbf{e}_1+\frac{1}{m}\mathbf{R}_r(\delta\mathbf{f}_a+\left \lfloor \delta\mathbf{R} \right \rfloor\mathbf{f}_a)\\
        =&\delta a_T\mathbf{R}_r\mathbf{e}_1-a_T\mathbf{R}_r\left \lfloor \mathbf{e}_1 \right \rfloor\delta\mathbf{R}\\
        &+\frac{1}{m}\mathbf{R}_r\delta\mathbf{f}_a-\frac{1}{m}\mathbf{R}_r\left \lfloor \mathbf{f}_a \right \rfloor\delta\mathbf{R}
    \end{split}
\end{align}
where
\begin{subequations}
\label{eq:deltafa}
\begin{align}
    \delta\mathbf{f}_a=&{\mathbf{f}_a}_r-\mathbf{f}_a\approx\frac{\partial {\mathbf{f}_a}_r}{\partial {\mathbf{v}_a^B}_r}\delta\mathbf{v}_a^B\\
    \begin{split}
        \delta\mathbf{v}_a^B =&{\mathbf{v}_a^B}_r-\mathbf{v}_a^B={\mathbf{R}_r}^T\mathbf{v}_r-\mathbf{R}^T\mathbf{v}\\
        \approx&{\mathbf{R}_r}^T\mathbf{v}_r-(\mathbf{I}+\left \lfloor \delta\mathbf{R} \right \rfloor){\mathbf{R}_r}^T({\mathbf{v}_a}_r-\delta\mathbf{v}_a)\\
        =&{\mathbf{R}_r}^T\delta\mathbf{v}_a+\left \lfloor {\mathbf{R}_r}^T\mathbf{v}_a \right \rfloor\delta\mathbf{R}
    \end{split}\\
    \delta\mathbf{v}_a =& {\mathbf{v}_a}_r-\mathbf{v}_a=\delta\mathbf{v}-\delta\mathbf{w}
\end{align}
\end{subequations}

Substituting (\ref{eq:deltafa}) to (\ref{eq:deltav}), we can obtain error dynamics $\dot{\delta\mathbf{v}}$ with respect to $\delta a_T$, $\delta\mathbf{v}$, $\delta\mathbf{R}$, and $\delta\mathbf{w}$.

According to \cite{bullo1995proportional}, we have:
\begin{align}
    \mathbf{A}^{-T}(\delta\mathbf{R}) =& \mathbf{I}+\frac{1}{2}\left \lfloor \delta\mathbf{R} \right \rfloor+(1-\alpha(\|\delta\mathbf{R}\|))\frac{\left \lfloor \delta\mathbf{R} \right \rfloor^2}{\|\delta\mathbf{R}\|^2}
\end{align}
where $\alpha(\|\delta\mathbf{R}\|) = \frac{\|\delta\mathbf{R}\|}{2}\cot\left(\frac{\|\delta\mathbf{R}\|}{2}\right)$.

Note that $\delta\mathbf{R}\to0$, $\mathbf{A}^{-T}(\delta\mathbf{R})\approx\mathbf{I}+\frac{1}{2}\left \lfloor \delta\mathbf{R} \right \rfloor$. Thus we have:
\begin{align}
\begin{split}
    \dot{\delta\mathbf{R}}\approx&\left(\mathbf{I}+\frac{1}{2}\left \lfloor \delta\mathbf{R} \right \rfloor\right)\left(-{\mathbf{R}_r}^T \mathbf{R} \boldsymbol{\omega}+\boldsymbol{\omega}_{r}\right)\\
    \approx&\left(\mathbf{I}+\frac{1}{2}\left \lfloor \delta\mathbf{R} \right \rfloor\right)\left(-(\mathbf{I}-\left \lfloor \delta\mathbf{R} \right \rfloor)\boldsymbol{\omega}+\boldsymbol{\omega}_r\right)\\
    =&\left(\mathbf{I}+\frac{1}{2}\left \lfloor \delta\mathbf{R} \right \rfloor\right)(\delta\boldsymbol{\omega}-\left \lfloor\boldsymbol{\omega} \right \rfloor\delta\mathbf{R})\\
    =&\left(\mathbf{I}+\frac{1}{2}\left \lfloor \delta\mathbf{R} \right \rfloor\right)\delta\boldsymbol{\omega}-\left \lfloor\boldsymbol{\omega} \right \rfloor \delta\mathbf{R}+\frac{1}{2}\left \lfloor \delta\mathbf{R} \right \rfloor^2\boldsymbol{\omega}
\end{split}
\end{align}

Moreover, we have:
\begin{subequations}
\begin{align}
\frac{\partial\left \lfloor \delta\mathbf{R} \right \rfloor^2\boldsymbol{\omega}}{\partial\delta\mathbf{R}}=&\mathbf{K}=\begin{bmatrix}\mathbf{K}_1 \\
                       \mathbf{K}_2 \\
                        \mathbf{K}_3
                        \end{bmatrix}\\
       \mathbf{K}_1=&\begin{bmatrix}\delta\mathbf{R}_2\boldsymbol{\omega}_y+\delta\mathbf{R}_3\boldsymbol{\omega}_z\\ -2\delta\mathbf{R}_2\boldsymbol{\omega}_x+\delta\mathbf{R}_1\boldsymbol{\omega}_y\\ -2\delta\mathbf{R}_3\boldsymbol{\omega}_x+\delta\mathbf{R}_1\boldsymbol{\omega}_z\end{bmatrix}^T\\
       \mathbf{K}_2=&\begin{bmatrix}\delta\mathbf{R}_2\boldsymbol{\omega}_x-2\delta\mathbf{R}_1\boldsymbol{\omega}_y\\ \delta\mathbf{R}_1\boldsymbol{\omega}_x+\delta\mathbf{R}_3\boldsymbol{\omega}_z\\ -2\delta\mathbf{R}_3\boldsymbol{\omega}_y+\delta\mathbf{R}_2\boldsymbol{\omega}_z\end{bmatrix}^T\\
       \mathbf{K}_3=&\begin{bmatrix}\delta\mathbf{R}_3\boldsymbol{\omega}_x-2\delta\mathbf{R}_1\boldsymbol{\omega}_z\\ -2\delta\mathbf{R}_2\boldsymbol{\omega}_z+\delta\mathbf{R}_3\boldsymbol{\omega}_y\\ \delta\mathbf{R}_1\boldsymbol{\omega}_x+\delta\mathbf{R}_2\boldsymbol{\omega}_y\end{bmatrix}^T\
\end{align}
\end{subequations}

\section*{ACKNOWLEDGMENT}
The authors would like to thank Xinyu Lv and Shihao Wu for their support in CFD and Yifeng Yu for his contribution in flight visualization. We also would like to thank anonymous reviewers for enhancing the article's quality.

\bibliographystyle{IEEEtaes}
\bibliography{myrefs}

\begin{IEEEbiography}[{\includegraphics[width=1in,height=1.25in,clip,keepaspectratio]{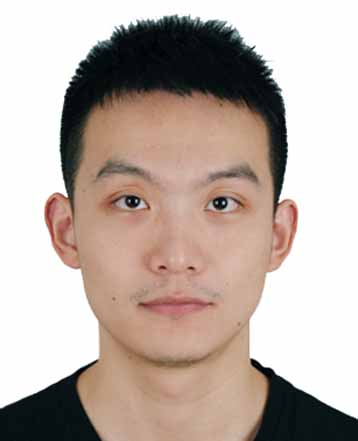}}]{Mingyue Fan}received B.Eng. and M.Eng. degrees in 2018 and 2021, respectively. He is currently working toward the Ph.D. degree in aerospace science and technology with Zhejiang University, Hangzhou, China.

His research interests include vertical take-off and landing aircraft, trajectory planning for atonomous systems, and numerical optimization.
\end{IEEEbiography}%

\begin{IEEEbiography}[{\includegraphics[width=1in,height=1.25in,clip,keepaspectratio]{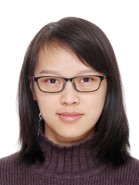}}]{Fangfang Xie}received the Ph.D. degree in fluid mechanics from Zhejiang University, Hangzhou, China, in 2013.

She is currently an Associate Professor with the School of Aeronautics and Astronautics, Zhejiang University, Hangzhou, China. Her research interests include intelligent bio-inspired unmanned aerial vehicles, physics-informed and data-driven flow modeling and control.
\end{IEEEbiography}

\begin{IEEEbiography}[{\includegraphics[width=1in,height=1.25in,clip,keepaspectratio]{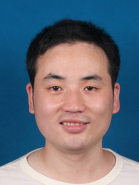}}]{Tingwei Ji} received the Ph.D. degree in materials processing engineering from Shandong University, Jinan, China, in 2008.

He is currently an Associate Research Fellow with the School of Aeronautics and Astronautics, Zhejiang University, Hangzhou, China. His research interests include bio-inspired unmanned aerial vehicles and intelligent design of aircraft.
\end{IEEEbiography}
\begin{IEEEbiography}[{\includegraphics[width=1in,height=1.25in,clip,keepaspectratio]{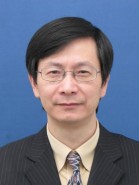}}]{Yao Zheng} is a Cheung Kong chair professor with Zhejiang University, appointed by the Ministry of Education of China since 2001. He is the founding deputy dean of School of Aeronautics and Astronautics, Zhejiang University, China. He had been a Senior Research Scientist for NASA Glenn Research Center, Cleveland, Ohio, USA, from 1998 to 2002. His current research interests include aerospace information technology, flight vehicle design, and fluid mechanics.
\end{IEEEbiography}

\end{document}